%% file: gauge_iclr.tex
\documentclass{article}
\usepackage{iclr2027_conference,times}

\usepackage[utf8]{inputenc}
\usepackage[T1]{fontenc}
\usepackage{microtype}
\usepackage{hyperref}
\usepackage{url}
\usepackage{amsmath,amssymb}
\usepackage{booktabs}
\usepackage{graphicx}
\usepackage{caption}
\usepackage{multicol}

\usepackage{tikz}
\usetikzlibrary{positioning, arrows.meta}
\definecolor{gnavy}{RGB}{8,48,107}
\definecolor{gmid}{RGB}{43,108,176}
\definecolor{glight}{RGB}{123,175,212}
\definecolor{gpale}{RGB}{214,228,240}
\definecolor{gfill}{RGB}{238,244,250}

\usepackage{colortbl}
\usepackage{tabularx}
\usepackage{tcolorbox}
\tcbuselibrary{listings,skins,breakable}
\newtcblisting{promptbox}[2][]{%
  listing only, breakable, enhanced jigsaw,
  colback=gfill, colframe=gnavy, boxrule=0.6pt, arc=1.2pt,
  left=4pt, right=4pt, top=3pt, bottom=3pt,
  title={#2}, fonttitle=\bfseries\footnotesize,
  coltitle=white, colbacktitle=gnavy,
  listing options={basicstyle=\ttfamily\scriptsize, breaklines=true,
    breakindent=0.9em, columns=fullflexible, keepspaces=true,
    postbreak=\mbox{\textcolor{gmid}{$\hookrightarrow$}\,}},
  #1}
\newtcolorbox{quotebox}[2][]{%
  breakable, enhanced jigsaw,
  colback=gfill, colframe=gnavy, boxrule=0.6pt, arc=1.2pt,
  left=4pt, right=4pt, top=3pt, bottom=3pt,
  title={#2}, fonttitle=\bfseries\footnotesize,
  coltitle=white, colbacktitle=gnavy, fontupper=\footnotesize,
  #1}
\newcommand{\subcaprow}[1]{\rowcolor{gpale}\multicolumn{5}{l}{\textbf{#1}}\\}
\newcommand{\graderD}{\textbf{[D]}}
\newcommand{\graderJ}{\textbf{[J]}}
\newcommand{\graderR}{\textbf{[R]}}
\newcommand{\yes}{\textcolor{gnavy}{\checkmark}}
\newcommand{\pmark}{\textcolor{gmid}{(\checkmark)}}
\newcommand{\nomark}{\textcolor{black!40}{$\times$}}

\iclrfinalcopy  

\title{One Analyst Is Not Ground Truth:\\
Grading Agent-Built Financial Models\\
Against Observed Professional Practice}

\author{Jiacheng Lu$^{1}$ \quad Yiming Li$^{2}$ \quad Sinuo Wang$^{3}$ \quad Wentao Zhao$^{4}$\\
\bfseries Rui Sun$^{5,\dagger}$ \quad Cheng Hua$^{1}$ \quad Tao Song$^{1,*}$ \quad Hui Cai$^{5}$ \quad Beidi Luan$^{5}$\\
\bfseries Tianyi Ma$^{1}$ \quad Qixuan Yang$^{1}$ \quad Zhengze Wu$^{1}$ \quad Jing Li$^{5}$\\
\bfseries Daxin Jiang$^{5}$ \quad Zuo Bai$^{5,*}$ \quad Haibing Guan$^{1}$\\
{\normalfont\small $^{1}$Shanghai Jiao Tong University \quad $^{2}$Nanyang Technological University}\\
{\normalfont\small $^{3}$University of Adelaide \quad $^{4}$Tsinghua University}\\
{\normalfont\small $^{5}$StepFun (Shanghai Jieyue Xingchen Intelligent Technology Co., Ltd)}\\
{\normalfont\small $^{*}$Co-corresponding authors. \quad $^{\dagger}$Project Leader.}\\
{\normalfont\small Tao Song: \href{mailto:songt333@sjtu.edu.cn}{\texttt{songt333@sjtu.edu.cn}}; \quad Zuo Bai: \href{mailto:baizuo@stepfun.com}{\texttt{baizuo@stepfun.com}}}}

\hypersetup{
  pdftitle={One Analyst Is Not Ground Truth: Grading Agent-Built Financial Models Against Observed Professional Practice},
  pdfauthor={Jiacheng Lu, Yiming Li, Sinuo Wang, Wentao Zhao, Rui Sun, Cheng Hua, Tao Song, Hui Cai, Beidi Luan, Tianyi Ma, Qixuan Yang, Zhengze Wu, Jing Li, Daxin Jiang, Zuo Bai, Haibing Guan}
}

\begin{document}

\maketitle
\lhead{}  

\begin{abstract}
    Language model agents can now construct complete financial models, but it remains unclear whether they have learned the financial judgment that gives those models meaning. Existing benchmarks often anchor correctness to an expert-authored solution (e.g., numerical targets or detailed rubrics) which introduces a implicit assumption: \emph{one expert solution can serve as ground truth.} This is appropriate when finance provides a unique answer, but not when judgment is required. Evidence from professional practice challenges this assumption: When financial models built by different analysts for the same company are graded against one another, the median score is only 0.33 under standard tolerances, \emph{revealing that single-reference grading confounds professional disagreement with error.} We therefore introduce GAUGE, a benchmark that decomposes financial-model evaluation into deterministic checks, rubric judgments and numerical rules. Built from 1,001 professional valuation  spanning 922 companies and all 25 GICS industry groups, GAUGE checks mechanical properties deterministically and evaluates judgment-bearing quantities against ranges in professional practice. We then validate GAUGE as a measurement instrument by testing expertise ordering, held-out professional values, and robustness to LM judgements. Our findings show that current LM agents are far better at constructing financial models than at deriving the company-specific assumptions that drive valuation, a gap that even persists after fine-tuning.
\end{abstract}

\section{Introduction}
\label{sec:intro}

A financial model is a core tool for company valuation. It reconstructs a company's three financial statements from reported history, projects them several years forward under explicit assumptions, and converts those projections into a value per share, most commonly through discounted-cash-flow analysis (DCF, i.e., forecasting the company's future cash flows and discounting them at a rate that reflects their risk). Building such models requires substantial expert time and effort: recent benchmarks report 8--10 hours for a DCF model and 13--16 hours for a three-statement model completed by human experts~\citep{ff}. Analysts in equity research and investment banking build these models routinely~\citep{demirakos2004valuation,imam2008use}. LM agents can now build them end to end, raising the prospect of automating a substantial part of this labor-intensive workflow~\citep{btb,ff,bfb,bluefin}.

This progress makes reliable evaluation increasingly important. Producing a
complete workbook is not enough. Some parts of a financial model are
\emph{mechanically} verifiable, i.e., historical figures
must match the source data, the balance sheet must balance, and
forecast cells must contain live formulas rather than pasted
constants. Other parts depend on \emph{financial judgment}. Revenue
growth, operating margins, discount rates, and target prices can
reasonably differ across competent analysts, and prior work has long
documented such disagreement in forecasts and valuation
practice~\citep{diether2002differences,imam2008use}. For these
judgment-bearing quantities, disagreement with one analyst is not
necessarily an error. This leads to a key question: \emph{Can we evaluate agent-built financial models by checking what has a clear answer while allowing for reasonable differences in financial judgment?} 

Existing financial-agent benchmarks, however, typically use an
expert-authored reference to grade outputs that depend on financial
judgment~\citep{btb,ff,bfb,bluefin}. Such grading makes an implicit assumption: \emph{when an agent differs from the reference
analyst, the agent is wrong.} Yet professional valuation gives us
little reason to make this assumption.
We test it directly by grading independently built professional
models against one another. Across 108 same-company pairs from
65 companies, the median score is only 0.33, and 92.6\% of pairs fall
below 0.70 under standard tolerances. Even at four times these
tolerances, a third remain below 0.70. Among models built from the
same fiscal-year data, no pair agrees on the implied share price
within 10\% (Section~\ref{sec:ava}). In other words, a grading rule
that treats one analyst as ground truth rejects much of the real,
meaningful variation that already exists among professionals.

This finding suggests a simple principle: use exact answers where
finance provides them, and professional practice where judgment can
reasonably differ. We build \textbf{GAUGE} on this principle which draws on 1{,}001 valuation workbooks created by practicing analysts
for real commercial engagements, spanning 922 companies and all 25
GICS industry groups. To our best knowledge, this is the largest
professional valuation corpus used in a financial-agent benchmark.
From it, we construct 196 full-model tasks that ask agents to build a
complete model from three years of historical financials. Scoring follows the same principle. Mechanical properties are checked
directly with deterministic code, while major structural errors
trigger hard \emph{gates}. Quantities that depend on financial
judgment are compared with an \emph{envelope}, i.e., ranges observed
in professional practice. Qualitative aspects are scored by a frozen
LLM judge under explicit rubrics~\citep{zheng2023judging,
guerdan2025validating}.

We evaluate GAUGE on 55 human participants under the same task setup and scoring rules used for agents. Senior analysts junior analysts, and finance students obtain mean scores of 88.3, 66.0, and 43.2, respectively. This ordering remains unchanged across our robustness checks (Section~\ref{sec:validation}). This wide spread provides a human
scale from students to experienced professionals for interpreting
agent performance. The practice envelopes include held-out
professional values while rejecting deliberately shifted ones, and
replacing Claude Opus 4.8 with GPT-5.6-sol as the judge produces
nearly the same agent ranking. Against this human baseline, we
evaluate 24 leading proprietary and open-weight LM agents
(Section~\ref{sec:results}). Every agent performs better on mechanical
construction than on financial judgment, with a median gap of
26 points. Claude Fable 5 leads the panel with a score of 53.4, above
the finance-student mean of 43.2 but below all 12 senior analysts and
15 of the 18 junior analysts. More interestingly, senior analysts show
only a 3.6-point gap on the same facets, suggesting that current
agents have learned to build financial models much faster than they
have learned to exercise judgment through them. Even trajectory fine-tuning on rollouts generated from the GAUGE training split improves valuation performance without closing this gap, suggesting that benchmark-aligned supervision can teach agents where plausible values lie more readily than the judgment behind them.

In summary, we make three contributions. (1) We show that
single-reference grading breaks down for financial judgment: when
independently built professional models of the same company are graded
against one another, most fail under standard benchmark tolerances,
with disagreement concentrated on the quantities where judgment
matters. (2) We introduce GAUGE, a benchmark of 196 full-model
tasks built from 1{,}001 licensed professional workbooks spanning
922 companies and all 25 GICS industry groups. GAUGE checks mechanical
correctness directly and evaluates judgment against
professional practice, with a 55-participant human baseline and
audits on held-out professional values and an independent judge.
(3) GAUGE reveals a gap across 24 LM agents between building financial
models and making the judgments that drive them. Every agent is
stronger at mechanical construction than financial judgment, with the
best agent still below every senior analyst. Further analysis
localizes this deficit to company-specific derivation rather than
broad risk ranking, and trajectory fine-tuning on GAUGE rollouts
improves valuation performance without fully closing the underlying gap.

\begin{figure}
    \centering
    \includegraphics[width=\linewidth]{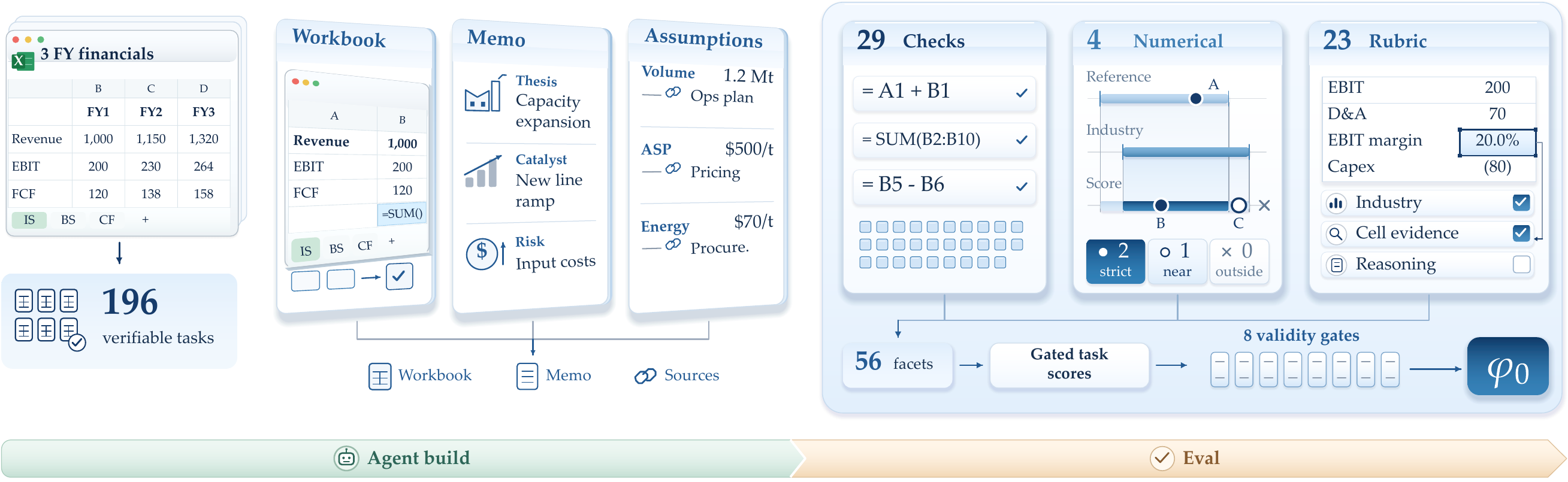}
    \caption{\textbf{GAUGE overview.}
Across 196 verifiable tasks, agents use three years of financial statements to produce a financial workbook, an investment memo, and source-linked assumptions. GAUGE evaluates these outputs through deterministic checks, numerical rules, and rubric judgments, followed by task-level score aggregation and validity gating to obtain the final benchmark score.
    }
    \label{fig:pipeline}
\end{figure}
\section{One Analyst Is Not Ground Truth}
\label{sec:ava}
\subsection{Revisiting the Single-Reference Assumption}
\label{sec:single-reference}

Existing financial-agent benchmarks commonly evaluate outputs against an
expert-authored reference. For quantities with a uniquely determined answer,
this is natural. For quantities that depend on financial judgment, however,
the same design embeds a stronger assumption: \emph{disagreement with the
reference analyst can be treated as error}. We test this assumption directly
on professionals themselves. Concretely, if one professional analyst's model
were graded using another analyst's independently built model of the same
company as the answer key, how well would it score?

To answer this, we use 137 workbooks from Section~\ref{sec:corpus} covering
65 companies, with two or three independently built models each. For each
ordered same-company pair, one workbook serves as the reference and the other
is graded against it. We compare nine quantities: revenue forecasts for the
next two fiscal years, implied share price, and six discount-rate quantities,
namely WACC, the risk-free rate, equity risk premium, beta, tax rate, and
terminal growth. Base tolerances are $\pm5\%$ for revenue, $\pm50$\,bp for
WACC, and $\pm10\%$ for implied share price; the rest use criterion-specific
tolerances. A pair's score is the fraction of criteria within tolerance. Of
158 directed pairs, 108 report at least three criteria and enter the analysis;
we repeat it with tolerances widened to $2\times$, $3\times$, and
$4\times$.

\textbf{Professionals score poorly when graded against one another.}
At standard tolerances, the median score is only 0.33, and 92.6\% of pairs
score below 0.70 (Figure~\ref{fig:ava}). The median rises to 0.50, 0.67, and
0.80 as tolerances widen to $2\times$, $3\times$, and $4\times$, respectively,
yet even at $4\times$ one third of professional pairs remain below 0.70.
The pattern persists among same-vintage models: across the 14 price pairs built
from the same fiscal-year data, none agrees on implied share price within 10\%,
and restricting the audit to same-vintage pairs leaves the median score at
0.33. The differences are economically substantial as well: the median
same-company WACC difference is 147\,bp and the median implied-price difference
is 25\%, with 90th percentiles of 374\,bp and 107\%
(Appendix~\ref{app:ava}).

These results expose a basic limitation of single-reference grading: \emph{any
difference from one analyst's estimate is counted as an error, even when
professional practice itself contains a range of defensible values.} An
expert-authored estimate can therefore serve as a reference without serving as
ground truth. More importantly, Figure~\ref{fig:ava} shows that disagreement is
not evenly distributed across quantities; it is concentrated precisely
where professional judgment matters most, which we examine next.
\subsection{Disagreement Concentrates Where Judgment Matters}
\label{sec:judgment-disagreement}

\begin{figure}[t]
\begin{minipage}[c]{0.45\textwidth}
\centering
\includegraphics[width=\linewidth]{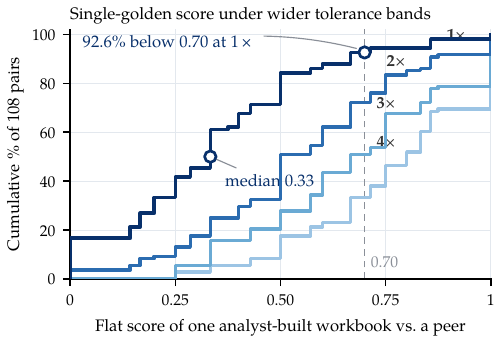}
\end{minipage}\hfill
\begin{minipage}[c]{0.455\textwidth}
\centering
\footnotesize
\setlength{\tabcolsep}{4pt}
\begin{tabular}{@{}lrcc@{}}
\toprule
Criterion & $n$ & $1\times$ & $3\times$\\
\midrule
Risk-free rate      & 82  & 66\% & 90\%\\
Forecast revenue FY+1 & 43  & 47\% & 60\%\\
Equity risk premium & 86  & 42\% & 81\%\\
Tax rate            & 100 & 32\% & 66\%\\
Beta                & 96  & 27\% & 67\%\\
Terminal growth     & 30  & 27\% & 47\%\\
Implied share price & 34  & 24\% & 53\%\\
WACC                & 122 & 13\% & 56\%\\
\bottomrule
\end{tabular}
\end{minipage}
\caption{\textbf{Professionals graded against professionals under a
single-reference rule.} \emph{Left:} across 108 directed same-company
pairs, the median score is 0.33 and 92.6\% score below 0.70 at
standard ($1\times$) tolerances; even at $4\times$, a third remain
below 0.70. \emph{Right:} pass rates by criterion ($n$ = directed
pairs stating it), with agreement declining for more judgment-intensive
quantities (Appendix~\ref{app:ava}).}
\label{fig:ava}
\label{tab:avacrit}
\end{figure}

Agreement varies systematically across quantities. It is highest for
quantities closely tied to observable inputs and lower for quantities that
require more analyst judgment (Figure~\ref{fig:ava}). The risk-free rate, which is largely observable from the market, falls within
tolerance in 66\% of pairs. Agreement declines across increasingly
judgment-based quantities: forecast revenue and the equity risk premium pass in
47\% and 42\%, while tax rate and beta pass in 32\% and 27\%. For quantities
deeper in valuation process, terminal growth passes in 27\% of pairs,
implied share price in 24\%, and WACC in only 13\%.

This pattern reveals a systematic \emph{observability gradient}
in professional practice: as a quantity moves from externally anchored
information toward company-specific judgment, agreement across professionals
becomes less concentrated around a single point estimate. Single-reference
grading is therefore least informative precisely where the task depends most
on financial judgment. What appears as error against one analyst can instead
fall within the variation exhibited by professionals.
\subsection{The Grader Should Follow What Is Answerable}
\label{sec:answerability}

The problem is not simply that the tolerances are too narrow. Even after every
tolerance is widened to $4\times$ its standard value, one third of professional
pairs still score below 0.70. Widening them further also weakens the test
itself: for example, a $\pm40\%$ implied-price band would admit substantially
different valuations while providing little discrimination. The more useful distinction is between quantities that have a uniquely
verifiable answer and those that admit professional judgment. Historical
figures, accounting identities, and formula relationships can be checked
directly. Forecast assumptions and valuation outputs instead require a
reference that reflects the range of values observed in professional practice.
This leads to the principle behind GAUGE: \emph{verify what is mechanically
answerable, and evaluate judgment against professional practice}. Section~\ref{sec:design}
turns it into benchmark's scoring design.

\section{GAUGE}
\label{sec:design}

\subsection{Corpus: Professional Practice as the Benchmark Foundation}
\label{sec:corpus}

GAUGE uses Finance Artifacts both to construct evaluation tasks and to
characterize professional practice for judgment-dependent scoring. The
corpus contains 1,001 valuation workbooks created by practicing financial
analysts for commercial deal and coverage work, licensed through a
commercial data provider. It spans 922 distinct tickers and all 25 GICS industry
groups across four complexity tiers, with workbooks ranging from roughly
2,000 to 35,000 spreadsheet cells and from 3 to 30 stated analyst-hours.
This gives GAUGE, to our knowledge, the first large-scale basis for
evaluating from-scratch valuation models against observed professional
practice rather than an expert-authored answer key. Crucially, these workbooks also contain the range-bearing artifacts needed
to define professional-practice scoring bands: 35\% use multiple valuation
methods, 31\% contain machine-readable sensitivity grids, and 54\% contain
at least one such artifact. Section~\ref{sec:envelope} uses these signals
to construct the corresponding professional-practice bands. To keep task
construction, calibration, and future evaluation cleanly separated, we
partition the corpus by company identity into 196 evaluation tasks, a
137-workbook subset covering 65 multi-covered companies for the audit in
Section~\ref{sec:ava} and envelope calibration, a 200-company training
split, and roughly 600 withheld workbooks for future tasks. Additional corpus curation and quality-control procedures are described in Appendix~\ref{app:corpus}.

\subsection{Task: A Complete Valuation Work Product}
\label{sec:task}

Each of the 196 tasks defines \emph{a fixed input--output contract}. The
input pack contains three fiscal years of audited historical financials,
segment and KPI skeletons, management guidance where disclosed, and an
as-of date that defines the information available to the agent. Consensus
and broker forecasts are withheld because they constitute the evaluation
targets. For each task, the agent must return \emph{three artifacts}: an eleven-sheet, formula-driven workbook covering the three statements, a driver-based revenue build, and a debt schedule, alongside WACC, DCF, sensitivity analysis, and self-check tab; a short investment memo with three to five
falsifiable quantified claims; and a supporting machine-readable assumptions file
in which every key assumption has a resolvable source pointer.

To ensure that task inputs are valid and auditable, a
provenance-tracked extractor records the source sheet, row, and label
for every input number and verifies the accounting identity
(e.g., total assets $=$ total liabilities $+$ equity) before release. Of 200
candidate tasks, 196 (98\%) pass one of five archetype schemas: 175
three-statement models, 13 income-statement/valuation or REIT/DDM
models, 5 bank payout/DDM models, and 3 pure-DCF models. The four
unresolved candidates are recorded as abstentions, and no reference
value is imputed. All agents then run in the same scaffold with access
to only two tools, a Python interpreter and file writes, without network
access, for at most 50 tool turns and 1{,}800 seconds. Complete prompts
and scaffold constants are provided in Appendix~\ref{app:harness}.

\subsection{Scoring: From Capability Taxonomy to a Failure-Aware Score}
\label{sec:scoring}

\textbf{Facet taxonomy and grading.}
GAUGE scores each run on 56 atomic and individually auditable
facets across \emph{five pillars}: input comprehension, model
construction, forecast reasoning, valuation, and communication \&
auditability. Each facet receives one of three scores: 0 for fail, 1
for pass, or 2 for excellent; facets whose preconditions are absent are
marked N/A. Complete rubrics are provided in Appendix~\ref{app:rubrics}.
Grading follows the type of facet being assessed: 29 deterministic facets use
binary code checks; 23 judged facets use the frozen LLM judge
described in Section~\ref{sec:judge}; and 4 rule-based facets use
workbook values under the rules in Section~\ref{sec:envelope}. The
rule-based facets include two envelope-band checks, one hidden-reference
accuracy check, and one self-referenced consistency check. Code accounts
for 54.0\% of scored facet outcomes and determines every gate decision.

\textbf{Task Dependent Activation.} Not every criterion applies to every company.
Banks, for example, are not valued by a WACC-discounted DCF, so WACC
facets deactivate and a dividend-based build is required instead; a
per-industry overlay plus instance-level N/A
determine the active set: 
\begin{equation}
\label{eq:active}
\mathcal{A}(x)=\bigl\{f :\;
\omega_{\mathrm{ind}(x)}(f)\neq\mathrm{N}
\;\wedge\; s_f(x)\neq\text{N/A}\bigr\},
\end{equation}
Inactive facets are omitted from aggregation rather than scored as zero, as reported in Appendix~\ref{app:overlays}.

\textbf{Score construction and validity gates.} Active facet scores map to a 0--100 scale via
$\phi(0)=0$, $\phi(1)=60$, and $\phi(2)=100$. The pre-gate score is
the bottom-up mean over active facets:
\begin{equation}
\label{eq:pregate}
\widetilde{\Phi}(x)
=
\operatorname{mean}_{\mathrm{pillars}}
\operatorname{mean}_{\mathrm{sub\text{-}caps}}
\operatorname{mean}_{f\in\mathcal{A}(x)}
\phi\!\left(s_f(x)\right).
\end{equation}

The scale has three implications: (1) a score of 60 means that every
active requirement reaches the professional-pass level; (2) scores above 60
require excellent performance on judged or envelope facets, since
deterministic checks have no excellent level; (3) and the fail-to-pass jump
is larger than the pass-to-excellent jump. Thus, a score below 60
indicates missing required substance.

We reveal that \emph{facet averages cannot capture categorical failures}. A workbook with an
unbalanced balance sheet is not simply a weaker model; it is not a
usable model. Eight deterministic gates detect such failures:
unbalanced balance sheets, untied cash flows, segments that do not sum
to the total, hardcoded forecasts, look-ahead beyond the as-of date,
unresolvable citations, unresolved circular references, and broken
enterprise-value-to-equity bridges. Each triggered gate $g$ imposes a
score ceiling $\kappa_g$, with the lowest triggered ceiling determining
the final score:
\begin{equation}
\label{eq:gates}
\Phi(x)
=
\min\left(
\widetilde{\Phi}(x),
\min_{g\in\mathcal{G}(x)}\kappa_g
\right).
\end{equation}
This prevents high facet coverage from offsetting a structural failure.
For example, one workbook passed 96\% of its facets but failed its
balance-sheet identity in all five forecast years; additive scoring
would assign it 60, whereas the gate caps it at 40. Gate definitions
and caps are provided in Appendix~\ref{app:gatedefs}, and the worked
example appears in Appendix~\ref{app2:example2}.

\textbf{Failure-aware headline score.} Averaging scores only over completed runs can overstate an agent’s overall performance. The headline score
$\phi_0$ therefore averages $\Phi$ over a fixed task denominator and
scores \emph{capability failures}, meaning runs that produce no valid
workbook within budget, as zero; the completed-only $\phi$ is
reported as a conditional-quality diagnostic. We retain capability failures without retries, so the reported scores reflect \emph{overall agent} performance under the fixed task budget.

\subsection{Practice Aware Grading.}
\label{sec:envelope}

Judgment-bearing quantities are scored against an \emph{envelope}: an
empirical range of values that professional practice actually uses,
derived from the corpus. For envelope facet $f$ on task $x$, let
$v_f(x)$ be the value extracted from the agent's workbook and
$E_f(x)$ the strict band; the near band $\widetilde{E}_f(x)$ widens
the strict band by half of its own width on each side. Membership sets
the facet score:
\begin{equation}
\label{eq:envstate}
s_f(x) \;=\; 2\cdot\mathbf{1}\!\left[v_f(x)\in E_f(x)\right]
\;+\; 1\cdot\mathbf{1}\!\left[v_f(x)\in \widetilde{E}_f(x)\setminus E_f(x)\right],
\end{equation}
with unmeasurable facets recorded as N/A, never zero. The strict bands
are built from the corpus, layer by layer. \textbf{E-method} comes
from the task's own reference workbook: for the WACC facet, the
author's stated WACC
widened by the 374\,bp cross-analyst dispersion measured in
Section~\ref{sec:ava}; for the implied-multiples facet, the valuation
ratios that the reference workbook itself implies (its P/E and
EV/EBITDA, i.e., price and enterprise value relative to earnings
measures) with 20\% flex, the two ratios' scores averaged.
\textbf{E-industry} is the 10th-to-90th-percentile range of the
assumption across all corpus workbooks in the same GICS industry
group, extractable for WACC from 86\% of workbooks; where both layers exist, the WACC band is their
intersection, so the specific layer tightens the general one. Two of
the four envelope facets use direct rules rather than bands: the
forecast-margin facet grades point accuracy against a hidden reference
margin derived from the reference workbook (full credit within 1.5
percentage points or 8\% relative, partial credit within 4 points or
20\%), and the segment-margin facet checks the model against itself,
crediting segment margins that are genuinely differentiated (a spread
of at least 3 points) over margins that are copies of one another.
The hidden margin target is withheld from the input pack, so matching
it cannot be done by reading the agent's own inputs. The third layer,
\textbf{E-company} (the 65
multi-covered companies), never supplies a scoring band: it
calibrates the dispersion widths and
audits the other two layers out of sample
(Section~\ref{sec:envvalid}). Envelope facets cover exactly the
quantities professionals disagree on most (Section~\ref{sec:ava}):
WACC, margin trajectories, and the valuation multiples implied by the
model's own output.

\paragraph{The judge.}\label{sec:judge}
Some criteria are qualitative craft that neither code nor bands can
grade, such as whether the revenue build follows real economic
drivers or whether the memo's risks map to model drivers. These 23
facets are
scored by a single frozen judge model (Claude Opus 4.8) that sees the
formula-visible workbook, the memo, and the assumptions file, and must
return a score with cited cell evidence against that facet's written
0/1/2 rubric rungs. Each facet receives $k{=}5$ independent votes,
reduced by majority with ties resolved \emph{toward the lower rung};
N/A requires a strict majority. Because the judge is part of the
instrument, Section~\ref{sec:validation} audits its stability, expert
agreement, and provider bias; verbatim prompts and the evidence
protocol are in Appendix~\ref{app:judge}.

\section{Validating the Instrument}
\label{sec:validation}

A benchmark’s scores should reliably reflect the ability it is designed to measure. Before evaluating agents, we test whether GAUGE meets this requirement using three sources of evidence: human performance, held-out professional data, and analyses of its scoring components.

\subsection{Known-groups validity: 55 humans under agent conditions}
\label{sec:human}

If GAUGE measures valuation ability, people with more of it should
score higher. We recruited 55 participants (12 senior analysts, 18
junior analysts, and 25 finance students, classified by the
professional network through which they were engaged) and had each
complete three GAUGE tasks under the agent's exact conditions: the
same input packs, the same three-artifact contract, and no network
access, scored by the identical frozen stack (165 attempts across 47
tasks; humans worked without the agents' wall clock, at a
median of 200 to 273 minutes per attempt).

\begin{figure}[t]
\begin{minipage}[c]{0.4\textwidth}
\centering
\includegraphics[width=\linewidth]{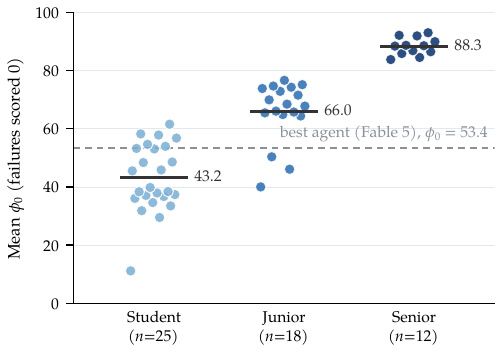}
\end{minipage}\hfill
\begin{minipage}[c]{0.55\textwidth}
\centering
\scriptsize
\renewcommand{\arraystretch}{1.5}
\setlength{\tabcolsep}{3pt}
\begin{tabular}{lrrrrrrrr}
\toprule
Group & $n$ & Compl. & $\phi$ & $\phi_0$ & Mech. & Judg. & Gate & Time\\
\midrule
Senior  & 12 & 100\% & \textbf{88.3} & \textbf{88.3} & 92.5 & 88.9 & 17\% & 200m\\
Junior  & 18 & 94\%  & 69.9 & 66.0 & 85.8 & 59.1 & 55\% & 243m\\
Student & 25 & 81\%  & 53.1 & 43.2 & 65.7 & 45.2 & 74\% & 273m\\
\midrule
Fable 5 & --- & 100\% & 53.4 & 53.4
& \multicolumn{2}{c}{93\%\,/\,78\%} & 10\% & \~20m\\
\bottomrule
\end{tabular}
\end{minipage}
\caption{\textbf{Known-groups validity: 55 humans under agent
conditions.} \emph{Left:} each dot is one participant's mean
failure-aware $\phi_0$ over three assigned tasks (bars: group means
43.2\,/\,66.0\,/\,88.3); the dashed line is the best agent
($\phi_0=53.4$), outscored by 33 of 55 humans including all 12
seniors. \emph{Right:} group summary; $\phi_0$ scores non-completions
as zero, Gate is the share of completed attempts triggering a validity
gate, and Time is the median minutes per attempt.}
\label{fig:known_groups}\label{tab:human}
\end{figure}

The three groups are cleanly ordered (Figure~\ref{fig:known_groups}):
$\phi_0$ of 88.3 (senior), 66.0 (junior), and
43.2 (student), with the same ordering on every secondary metric
(completion, gate rate, contract delivery).
Separations are individual-level, not just mean-level: Cliff's
$\delta$ is 1.00 (senior vs.\ junior) and 0.996 (junior vs.\
student), both Mann--Whitney $p<10^{-5}$, and the
weakest senior (83.8) outscores the strongest junior (76.6). Nor is
the ordering an artifact of who got which tasks: it holds in all
10{,}000 task-cluster bootstrap resamples, and task fixed effects move
the group gaps by under two points.

The human study also rules out one deflationary reading of
Section~\ref{sec:results}, namely that judgment facets are simply
\emph{harder} than mechanical ones for everyone. Senior analysts pass
mechanical and judgment facets almost equally (92.5 versus 88.9, a
3.6-point gap), against gaps of 26.7 points for juniors and 20.5 for
students. The gap is smaller among senior analysts but remains evident in agent performance.

\subsection{Envelope validity: held-out coverage and counterfeit rejection}
\label{sec:envvalid}

Because dispersion estimates and band widths are calibrated on the same
65-company multi-covered corpus that could otherwise validate them, we
evaluate the envelope layers with company-grouped cross-fitting (20
repetitions of five-fold splits keeping all of a company's workbooks in
one fold; dispersion tails and industry bands re-estimated with each
held-out company removed from its calibration pools). In this audit,
E-method is the minimum-to-maximum range of the
reference workbook's own multi-method valuations and sensitivity grids,
tested on the implied price, with a near rule that widens the band by
the held-out p90 cross-analyst dispersion. On held-out
companies, E-method covers 53.8\% of peer implied prices strictly
(95\% CI 38.5--68.6\%) and 91.2\% under the near rule
(82.6--97.2\%); strict held-out E-industry coverage is 82.4\% for
WACC, 90.2\% for terminal growth, and 75--85\% for the remaining
assumption facets (Appendix~\ref{app2:additional}).

Coverage alone cannot distinguish a valid band from a wide one, so we
add a counterfeit-rejection control: displace every held-out
professional value by $\pm k$ times its facet's typical same-company
professional disagreement ($k\in\{1,2,3\}$; e.g., 147\,bp for WACC)
and re-test membership under the unchanged rule. The E-industry layer
admits 80.8\% of real held-out values while rejecting 66.1\% of
$\pm2$-unit and 85.4\% of $\pm3$-unit counterfeits, monotonically in
$k$ for every facet. A p75-to-p90 percentile sweep traces
the frontier between coverage and selectivity; the released p90
setting is an
interior point on it. One component still earns
only partial trust: the near-price band is asymmetric and
rejects only upward counterfeits, so membership in it earns at most
partial credit, while the strict price band that alone earns full
credit rejects 65.4\% of the same counterfeits
(Appendix~\ref{app2:additional}).

\subsection{Judge validity: stability, experts, and a rival judge}
\label{sec:judgeval}

\textbf{Stability.} With $k{=}5$ votes, two independent re-judgings of
the same cells agree at Kendall $\tau=0.944$ with a 2.2\% facet flip
rate, and larger $k$ improves this only gradually (full curve in
Appendix~\ref{app:kablation}).

\textbf{Experts.} In a 460-case expert audit of the judged facets
(three expert annotators; provenance and scope in
Appendix~\ref{app:judge}), the judge matches
the expert consensus on 86.7\% of cases (weighted $\kappa=0.81$),
while the experts agree with each other on 89.4\% ($\kappa=0.85$): the
judge is nearly as close to the expert consensus as the experts
are to each other. Agreement falls from mechanical-adjacent facets
(91.2\%) to assumptions (85.0\%) to valuation (82.9\%), which is the
same in Figure~\ref{fig:ava}.

\textbf{A rival judge.} The frozen judge shares a provider with the
two top-scoring agents, so we re-judged a stratified 96-cell subset (12
tasks $\times$ 8 agents spanning five providers) with a judge from a
competing provider, GPT-5.6-sol, under the identical protocol. The
cross-family judge agrees exactly on 73.5\% of facet scores and within
one rung on 92.2\% (quadratic-weighted $\kappa=0.675$). It is
uniformly stricter ($-5.0\,\phi$ on judged facets), and the shift is
family-neutral: cells generated by the judge's rival family move
$-5.2$ versus $-4.8$ for all others (permutation $p=0.71$). Agent
ordering is preserved (Kendall $\tau=0.857$), and the top agent is
unchanged: the OpenAI judge also ranks Claude Fable 5 first, above
its own maker's GPT-5.6-sol (Appendix~\ref{app2:xfam}).

\textbf{Which grader class carries the ranking?} Removing one grader
class at a time and re-ranking (5{,}000 paired bootstraps), the
ranking correlates at Kendall $\tau=0.888$ without deterministic
facets, $0.965$ without rule-based facets, and $0.683$ without judged
facets: the judged facets carry the most ranking information while
the deterministic layer and its gates anchor validity. Additive credit
with gates removed flips 11 of 276 model-pair orderings, and in 10 of
the 11 it prefers the model that triggers \emph{more}
structural-failure gates (Appendix~\ref{app2:additional}).

\section{Measuring 24 Agents}
\label{sec:results}

\subsection{Setup}
We evaluate 24 agents from twelve providers: Claude Fable 5, Claude
Opus 4.8, Claude Sonnet 5, GPT-5.6 (sol, terra, luna), Gemini 3.1 Pro
and 3.5 Flash, Grok 4.5, DeepSeek v4 (pro, flash), Kimi k2.6,
k2.7-code, and k3, Qwen3 235B, Qwen3 Coder, and Qwen3.7-max, Hunyuan
hy3, GLM 5.2, MiniMax M3, Doubao 2.1-pro and Doubao-seed-evolving,
Step 3.7 Flash, and
GPT-OSS-120B~\citep{claudefable,claudeopus,claudesonnet,gpt56,
gemini31pro,gemini35flash,grok,deepseekv4,kimik2,kimik3,qwen3,
qwen3coder,qwen37max,hunyuan,glm5,minimaxm3,seed18,step37flash,
gptoss}. All run the identical scaffold(Section~\ref{sec:task}),
differing only in the provider-specific wire adapter
(Appendix~\ref{app2:wire}). Each agent attempts a fixed 48-task panel
stratified by tier and industry, once per task
1{,}152 cells.

\begin{table}[t]
\caption{\textbf{Failure-aware results on the 48-task panel}. $\phi_0$ scores capability
failures as zero on the fixed 48-task denominator and orders the
table. Compl.\ is the share of tasks with a valid workbook; Mech./Judg.\
are completed-cell facet pass rates (score $\geq1$); Gate is the share
of completed cells triggering at least one validity gate; Op.\ marks
open-weight releases.}
\label{tab:leaderboard}
\centering
\scriptsize
\setlength{\tabcolsep}{2.8pt}
\renewcommand{\arraystretch}{1.02}
\begin{minipage}[t]{0.503\textwidth}
\centering
\begin{tabular}{@{}lrrrrrc@{}}
\toprule
Agent & Compl. & $\phi_0$ & Mech. & Judg. & Gate & Op.\\
\midrule
Claude Fable 5         & 100\% & \textbf{53.4} & \textbf{93\%} & \textbf{78\%} & \textbf{10\%} & \nomark\\
Claude Opus 4.8        & 100\% & 49.6 & 92\% & 68\% & 17\% & \nomark\\
GPT-5.6-sol            & 100\% & 46.5 & 86\% & 73\% & 46\% & \nomark\\
Kimi k2.7-code         & 98\%  & 41.3 & 86\% & 59\% & 47\% & \yes\\
GLM 5.2                & 100\% & 41.0 & 80\% & 58\% & 54\% & \yes\\
Grok 4.5               & 88\%  & 40.7 & 85\% & 68\% & 43\% & \nomark\\
GPT-5.6-terra          & 100\% & 39.8 & 79\% & 61\% & 71\% & \nomark\\
Claude Sonnet 5        & 98\%  & 39.7 & 81\% & 60\% & 55\% & \nomark\\
DeepSeek v4-pro        & 100\% & 39.0 & 81\% & 57\% & 62\% & \yes\\
Doubao-seed-evolving   & 92\%  & 38.8 & 85\% & 60\% & 39\% & \nomark\\
DeepSeek v4-flash      & 100\% & 38.7 & 84\% & 54\% & 77\% & \yes\\
GPT-5.6-luna           & 100\% & 37.6 & 80\% & 54\% & 75\% & \nomark\\
\bottomrule
\end{tabular}
\end{minipage}\hfill
\begin{minipage}[t]{0.478\textwidth}
\centering
\begin{tabular}{@{}lrrrrrc@{}}
\toprule
Agent & Compl. & $\phi_0$ & Mech. & Judg. & Gate & Op.\\
\midrule
Hunyuan hy3            & 100\% & 37.5 & 81\% & 47\% & 50\% & \yes\\
Gemini 3.1 Pro         & 100\% & 36.8 & 80\% & 53\% & 54\% & \nomark\\
Gemini 3.5 Flash       & 100\% & 35.7 & 80\% & 48\% & 79\% & \nomark\\
Kimi k3                & 90\%  & 34.7 & 83\% & 54\% & 58\% & \nomark\\
Step 3.7 Flash         & 96\%  & 32.9 & 74\% & 46\% & 85\% & \yes\\
Kimi k2.6              & 100\% & 32.5 & 77\% & 44\% & 88\% & \yes\\
MiniMax M3             & 79\%  & 31.0 & 77\% & 56\% & 71\% & \yes\\
Qwen3 Coder            & 100\% & 28.1 & 68\% & 34\% & 88\% & \yes\\
Qwen3.7-max            & 65\%  & 24.9 & 83\% & 54\% & 55\% & \nomark\\
Doubao 2.1-pro         & 25\%  & 10.2 & 81\% & 60\% & 58\% & \nomark\\
Qwen3 235B             & 33\%  & 9.6  & 66\% & 40\% & 88\% & \yes\\
GPT-OSS-120B           & 44\%  & 9.3  & 61\% & 21\% & 76\% & \yes\\
\bottomrule
\end{tabular}
\end{minipage}
\end{table}

\subsection{Every agent constructs better than it judges}
\label{sec:gap}

\begin{figure}[t]
\centering
\includegraphics[width=0.7\textwidth]{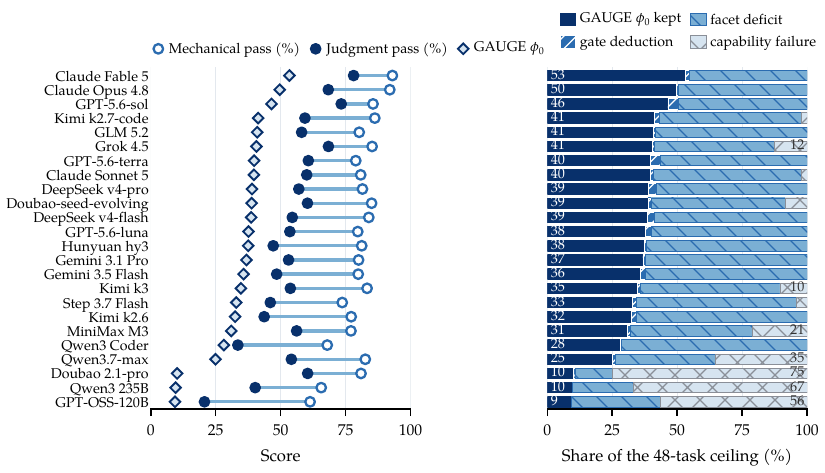}
\caption{\textbf{Mechanical pass rates exceed judgment pass rates
for all 24 agents.}
\textbf{(a)}~Open and filled circles show mechanical and judgment
pass rates, respectively, with a median gap of 26 percentage points;
diamonds show the failure-aware score $\phi_0$.
\textbf{(b)}~Each bar partitions 100 points on the 48-task panel into
the reported $\phi_0$, shortfalls in facet scores, gate deductions,
and capability failures. Facet shortfalls are the largest source
of score loss for 21 of 24 agents, while gate deductions are
4.2 points.}
\label{fig:scatter}\label{fig:loss}
\end{figure}

Table~\ref{tab:leaderboard} and Figure~\ref{fig:scatter} report the
results. The facet taxonomy assigns every facet to
mechanical construction or judgment before any scoring. All 24 agents pass a
larger share of mechanical facets than judgment facets: mechanical
pass rates span 61--93\% across the fleet, judgment rates 21--78\%,
and the median within-agent gap is 26 points. The best
agent, Claude Fable 5, passes 93\% of mechanical but 78\% of judgment
facets while triggering gates on only 10\% of tasks. GPT-5.6-sol passes 73\% of judgment facets, but 46\% of its
completed tasks trigger at least one validity gate.

Placed on the human scale, the best agent's 53.4 sits above the
student mean (43.2) but below every one of the 12 senior analysts
(minimum 83.8) and 15 of the 18 juniors; 33 of the 55 participants
outscore it, and the junior and senior group means beat it in 100\% of
bootstrap resamples. The one reversal is the gate rate (best agent
10\% against the seniors' 17\%): gates measure schema compliance,
where generated workbooks can be byte-precise and hand-built ones are
not. The experience ordering is carried by judgment rather than
formatting discipline, which is what a valuation benchmark reward.

\subsection{Score composition and ranking stability}
\label{sec:anatomy}

\textbf{Sources of score loss.}
Facet shortfalls dominate losses for 21 of 24 agents,
while gate deductions are at most 4.2 points
(Figure~\ref{fig:scatter}).
The free cash flow definition passes in 98\% of applicable cases versus 2\% for separating maintenance from growth capital
expenditure and 9\% for sensitivity analysis on operating drivers.
This suggests difficulty adapting standard calculations
to a company's operations.

\textbf{WACC estimates.}
Among 532 completed runs reporting WACC, 40\% of estimates are
multiples of 50\,bp and 27\% of 100\,bp, compared with 12\% and
7\% in 120 workbooks.
Agent--ref rank correlations across companies
($\rho=0.36$--$0.42$) are nevertheless close to cross-analyst
agreement on the same companies ($\rho=0.38$).
Similar rank agreement thus coexists with coarser numerical
assumptions.

\textbf{Ranking stability.}
Across 50{,}000 paired bootstraps, the leader and
top-three set are preserved in 99.998\% and 99.848\% of replicates,
respectively, while the top-five set is preserved in 22.4\%.
The leading group is stable, with uncertainty around the top-five
cutoff.
Completion affects ranking: at 25\% completion, Doubao 2.1-pro
falls from rank 8 under completed-only scoring to rank 22 under
$\phi_0$.
The construction--judgment gap persists across corpus tiers and
all 11 sectors (Appendix~\ref{app2:slices}).

\subsection{Can the gap be trained away?}
\label{sec:train}

GAUGE's training split (200 companies, disjoint from evaluation and
calibration) lets us test whether the gap yields to corpus
supervision. In-context, adding E-industry assumption tables (with no
figures from the test company) improves the valuation-judgment
subscore by $+4.0\,\phi$ (95\% CI $[+1.3,+6.7]$) while mechanical
scores are unchanged; worked-example cards yield a non-significant
$+1.0$. Fine-tuning on 146 rejection-sampled trajectories moves
valuation by $+8.4\,\phi$ but assumptions by $-4.9$, leaving the
overall score unchanged ($-0.8$, CI $[-3.6,+2.0]$). In both
experiments the gains concentrate on envelope-scored facets, the
exact quantities the corpus statistics describe (Appendix~\ref{app2:training}): corpus data teaches an agent where
professional numbers live, not the judgment that surrounds them.

\section{Conclusion}
\label{sec:conclusion}

GAUGE addresses the failure of single-reference grading through deterministic checks, validity gates, and judgment scoring grounded in professional practice. Validation with 55 humans, held-out professional values, and an independent judge supports this design. All 24 agents score higher on construction than judgment, and the best remains below every senior analyst in our study. Fine-tuning improves valuation scores without closing this gap. GAUGE thus provides a testbed for improving agents' ability to derive and defend company-specific valuation assumptions.

\subsubsection*{Reproducibility Statement}
The methodology tier is public: the complete 56-facet rubric set,
gate definitions, industry overlays, scorer, judge prompts, harness
and extraction code, envelope statistics as raw per-workbook value
lists, the 632 analyst-vs-analyst pair records behind
Section~\ref{sec:ava}, and the frozen run ledgers including all five
raw votes for every judged facet (inventory in
Appendix~\ref{app2:release}). De-identified evaluation and training
workbooks are released as a gated tier under a research-use
agreement, as the underlying artifacts are commercially licensed and
cannot be redistributed raw; roughly 600 workbooks are withheld to
supply future refresh versions. Repository and dataset links will be
provided in a future version.

\subsubsection*{Ethics Statement}
The corpus is licensed for research use and contains public market
and accounting information only; released workbooks pass an XML-level
de-identification pipeline (author metadata, paths, and link targets
scrubbed with formula counts verified unchanged), and an independent
re-scan reports zero residual findings
(Appendix~\ref{app2:release}). Human participants consented to
research use of their de-identified outputs, no identifying
information was collected, and compensation was at prevailing
professional rates (Section~\ref{sec:human}). GAUGE evaluates
model-building competence; its scores are not investment advice, and
the released data are gated to research use to respect the underlying
commercial licenses.

\bibliography{gauge}
\bibliographystyle{iclr2027_conference}

\clearpage
\appendix
\input{appendix}

\end{document}

%% file: appendix.tex
\newcommand{\appcontentsitem}[3]{%
  \noindent\textbf{#1.}\enspace
  \hyperref[#3]{#2}\dotfill\enspace
  \hyperref[#3]{\pageref{#3}}\par\vspace{2pt}}

\begin{center}
{\Large\bfseries Appendix}\\[4pt]
\begin{minipage}{0.86\textwidth}\small
The appendix is organized in dependency order: audit details and
reproducibility first, then the instrument itself (harness, rubrics,
gates, overlays, judge protocol, provenance), then full results,
worked examples, and the release protocol.
\end{minipage}
\end{center}
\medskip
\begin{multicols}{2}\small
\appcontentsitem{A}{Additional Peer-Workbook Audit Details}{app:comparison}
\appcontentsitem{B}{Reproducibility Details}{app:repro}
\appcontentsitem{C}{Corpus Organization and Quality Control}{app:corpus}
\appcontentsitem{D}{Task Construction and Agent Harness}{app:harness}
\appcontentsitem{E}{Wire Adapters and Recovery Shims}{app2:wire}
\appcontentsitem{F}{Tool Use Under One Scaffold}{app2:tooluse}
\appcontentsitem{G}{The 56-Facet Taxonomy: Complete Rubrics}{app:rubrics}
\appcontentsitem{H}{Validity Gates: Detection and Caps}{app:gatedefs}
\appcontentsitem{I}{Industry-Conditional Activation}{app:overlays}
\appcontentsitem{J}{Instrument Provenance and Calibration History}{app2:provenance}
\appcontentsitem{K}{Judge Protocol and Expert Audit}{app:judge}
\appcontentsitem{L}{Per-Facet Results: The Full Matrix}{app:facetmatrix}
\appcontentsitem{M}{Results by Slice}{app2:slices}
\appcontentsitem{N}{Additional Main-Result Diagnostics}{app2:additional}
\appcontentsitem{O}{Training-Signal Experiments}{app2:training}
\appcontentsitem{P}{Cost, Latency, and Compute}{app2:cost}
\appcontentsitem{Q}{Worked Example: One Task End to End}{app:example}
\appcontentsitem{R}{Worked Example II: A Gated Cell}{app2:example2}
\appcontentsitem{S}{Three Failure Trajectories, Verbatim}{app2:failures}
\appcontentsitem{T}{Gate Trigger Profiles}{app:gates}
\appcontentsitem{U}{Judge Vote-Sampling Ablation}{app:kablation}
\appcontentsitem{V}{Release Protocol and Checklist}{app2:release}
\appcontentsitem{W}{Related Work}{app:related}
\appcontentsitem{X}{Limitations}{app:limitations}
\end{multicols}
\clearpage

\section{Additional Peer-Workbook Audit Details}
\label{app:comparison}
\label{app:ava}

\begin{figure*}[t]
\centering
\includegraphics[width=\textwidth]{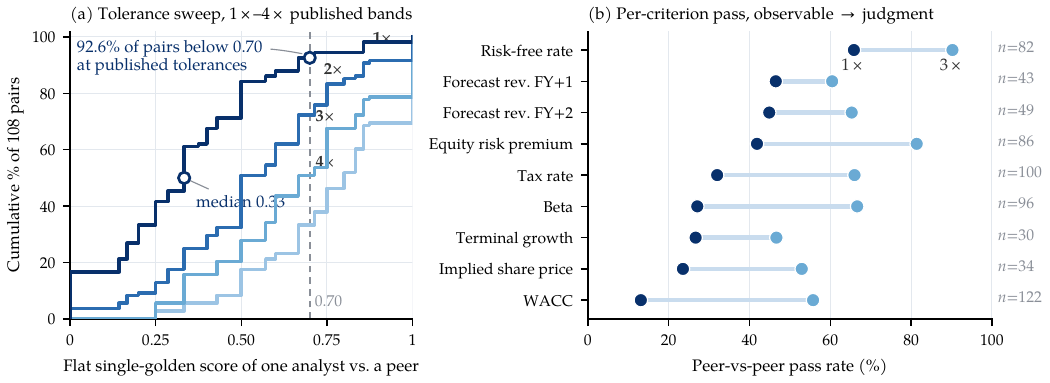}
\caption{\textbf{Full peer-workbook audit.} (a) Cumulative single-golden
scores for 108 directed pairs as tolerance bands widen from $1\times$ to
$4\times$. (b) Criterion pass rates at $1\times$ and $3\times$.}
\label{fig:ava-full}
\end{figure*}

\begin{figure}[t]
\centering
\includegraphics[width=0.58\textwidth]{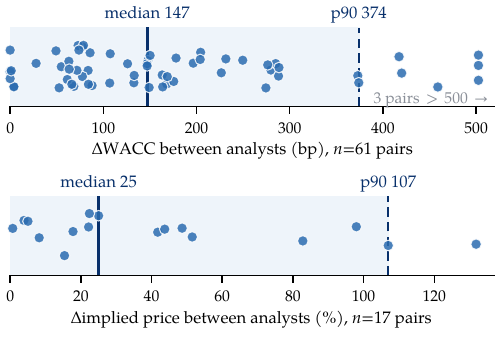}
\caption{\textbf{Observed same-company disagreement.} Median/p90
absolute differences are 147/374\,bp for WACC and 25\%/107\% for
implied price. Tail estimates use 61 and 17 undirected pairs,
respectively.}
\label{fig:disagree}
\end{figure}

Each workbook is used as the reference for every same-ticker peer. Of
158 directed pairs, 108 state at least three of the nine graded
criteria. Figure~\ref{fig:ava} (right panel) and
Figure~\ref{fig:disagree} report
the criteria that drive disagreement and the small tail samples.
Unstated values are N/A; two probable unit/currency-mismatch tickers are
excluded from implied-price comparisons. The same-vintage subset
($n=42$ directed pairs) has median score 0.33. Under this tolerance
rule, the pairs are not interchangeable; the audit does not determine
which workbook is correct.

\section{Reproducibility Details}
\label{app:repro}
Extraction pipelines record each number's sheet, row, and label; batch
QC reports list every unparsed workbook and its reason. Analyst-vs-analyst
pair records (per pair and tolerance) are released as JSONL. Rubrics,
scorer, harness, and methodology-tier artifacts are public; workbook data
are gated under a research-use agreement (see the Reproducibility
Statement). Repository and dataset links will be provided in a future
version.

\section{Corpus Organization and Quality Control}
\label{app:corpus}

\textbf{Folder units.} The data provider (``vendor'' below) delivers
the corpus as \emph{folder units}: one merged \texttt{.xlsx} workbook
bundling every
valuation method the analyst built for one company (a Large-tier
workbook can run to $\sim$30 tabs), plus a machine-readable
\texttt{README} documenting tab structure, methods, key
input--output relationships, and sensitivity notes. Per the vendor's
data card, every artifact cleared a commercial-intent and
internal-review bar (review hours are excluded from the stated
analyst-hours), uses public market and accounting inputs only, and
contains no macros and no personal data.

\textbf{Vendor-stated vs.\ QC-verified.} We re-derive every count we
use from the delivered bytes rather than quoting the data card. The
card states 400/350/200/50 folder units per tier; our sheet-level
parse of all 1{,}001 workbooks counts 404/347/200/50. The card's tier definitions (cells,
analyst-hours, minimum methods) are quoted as definitions, but our
20-workbook deep audit found Large-tier cell medians of $\sim$35K ---
above the stated 12K--25K range --- and one Premium workbook below the
stated formula share; we report both deviations rather than suppress
them. Input-material diversity grows with tier by design (Small:
filings, decks, basic market data; Medium adds earnings calls;
Large adds industry reports and expert calls; Premium adds bespoke
operating analyses), which makes tier a usable mechanical-scale
covariate in Appendix~\ref{app2:slices}.

\textbf{GICS reconciliation (26 vs.\ 25).} The vendor's data card
lists 26 industry groups over $\sim$1{,}033 ticker entries. Two
normalizations produce the numbers used in this paper: (i)~the card
counts the legacy \emph{Industrial Conglomerates} group (one ticker),
which the March-2023 GICS revision retired --- we classify under the
current 25-group standard; (ii)~the card's ticker list includes
duplicate listings and alternative names for the same company, which
our identity pass de-duplicates to 922 distinct tickers. All corpus
statistics in this paper therefore read ``922 tickers, 25 GICS
industry groups.''

\textbf{README caveat.} README files are template-instantiated ---
546 of 1{,}001 contain an identifiable copy-paste artifact --- so we
use them only as label sources after cleaning (method and tab
inventories), never as human-written documentation, and none of the
graded facets consult them.

\section{Task Construction and Agent Harness}
\label{app:harness}

This appendix gives the system prompt (\ref{app:sysprompt}), deliverable
specification (\ref{app:deliverable}), rendered input-pack excerpt
(\ref{app:packexcerpt}), and scaffold constants (\ref{app:constants}).

\textbf{Input packs.} A provenance-tracked extractor builds each visible
pack from the reference workbook's historical sections. Every number
carries its source sheet, row, and label; the accounting identity
$TA = TL + TE$ is checked before release. Packs contain three fiscal
years of audited historicals (revenue, COGS, opex, D\&A, interest, tax,
net income, the three balance-sheet totals, and capex), plus as-of share
price, share count, total debt, and cash. They contain no peer multiples,
consensus, or broker data. Five archetype schemas are listed in
Section~\ref{sec:task}.

\textbf{Prompt contents.} The rendered prompt sets the role (junior
equity-research associate reporting to a senior analyst), as-of date,
and pack-only data rule. It then inserts whitelisted pack tables; a
byte-scan of rendered prompts across models found no reference WACC,
fair value, or EV. The final block requires one workbook with eleven
named sheets, five formula-live forecast years, tie-outs, a memo with
3--5 falsifiable quantified claims matching the workbook, and an
assumptions file of \texttt{\{name, value, source\}} records with a
resolvable \texttt{Sheet!Cell} or input-pack pointer.

\subsection{System Prompt (Complete, Verbatim)}
\label{app:sysprompt}

Every agent receives the same system prompt; only the tool-call wire
format differs per provider.

\begin{promptbox}{System prompt --- identical for all 24 agents}
You are a junior equity-research associate. A senior
analyst has asked you to deliver a complete,
institutional-quality financial model for a public
company. The deliverable is a single Excel workbook
(.xlsx) that the analyst will open and use without
further questions.

You will receive three years of audited historical
financials. Your task is to project the next five
fiscal years, assemble the three integrated
statements, and build a DCF valuation -- *not* to
source raw filings.

You have the following local tools:
- A Python interpreter with openpyxl installed.
- A shell with write access to the current working
  directory.
- No internet, no live market-data API, no MCP servers.

You may NOT rely on outside calls. Use only the inputs
given in the user message. Do not assume access to
filings, terminals, or research databases beyond what
is explicitly provided.

Deliverable rules -- these are not negotiable. The
model will be machine-graded and judged by a senior
analyst.

1. Save the workbook to ./output/{TICKER}_model.xlsx.
   Create the ./output/ directory if it does not exist.
2. Produce a SINGLE workbook with all sheets in it.
3. Every projected/forecasted number must be a live
   Excel **formula** that references inputs -- never a
   value computed in Python and written as a number.
   The senior analyst will flex assumptions; hardcoded
   outputs are a silent bug.
4. The model must produce a final **implied share
   price** that is clearly labeled and easy to find.
5. Use real accounting conventions (GAAP-style). The
   Income Statement, Balance Sheet, and Cash Flow
   Statement must tie together -- net income flows to
   retained earnings and to the cash flow statement,
   ending cash on CF equals cash on BS, etc.
6. No placeholder text, no "TODO" cells, no "Excel Data
   Table feature" notes. The workbook must be fully
   functional when opened.

Do not ask clarifying questions. Make reasonable
analyst-grade assumptions. Document them in the
workbook (cell comments or an Assumptions sheet) but do
not block on them.

Work efficiently. Spend your reasoning on the model
itself, not on the surrounding deliverable.
\end{promptbox}

\subsection{Deliverable Specification (Complete, Verbatim)}
\label{app:deliverable}

After the input-workbook dump (42{,}000-character budget,
Section~\ref{app:packexcerpt}), the prompt closes with the
deliverable block. The placeholders \texttt{\{model\_xlsx\}},
\texttt{\{memo\_md\}}, and \texttt{\{assum\}} are absolute output
paths filled at render time.

\begin{promptbox}{Deliverable block --- the three-artifact contract}
# DELIVERABLE -- build it with python3 + openpyxl via
your Bash tool

Write THREE files to these EXACT absolute paths
(mkdir -p the directory first):

1. {model_xlsx}
   A single institutional-quality .xlsx with these
   sheets: Cover, Assumptions, Revenue_Build (segment
   drivers), Income_Statement, Balance_Sheet, Cash_Flow,
   Debt_Schedule, WACC, DCF (Valuation), Sensitivity,
   Checks.
   Hard requirements (machine-graded + judged):
   - 5 forecast years after the last actual; every
     forecast cell is a live FORMULA referencing
     Assumptions cells -- NEVER a number computed in
     python and pasted.
   - 3 statements TIE: balance sheet balances every
     period (no plug to 'other'); CF ending cash == BS
     cash; NI flows to retained earnings.
   - DCF on unlevered FCF discounted at a WACC built on
     the WACC sheet; an explicit EV -> (net debt) ->
     equity -> per-share bridge with a clearly labelled
     implied price.
   - >=1 sensitivity grid of LIVE formulas on real
     value drivers.
   - Segment revenue built from drivers (volume/price)
     where the input gives them.
   - Color: blue font = hardcoded input, black =
     formula.

2. {memo_md}
   A short investment memo: thesis with 3-5 falsifiable,
   quantified claims; key risks mapped to model drivers;
   headline numbers (implied price, EPS) that MATCH the
   workbook.

3. {assum}
   JSON list of every key assumption:
   {"name","value","source"} where source is a
   resolvable pointer ("Sheet!Cell", or the input row it
   came from). Every number you cite in the memo must
   appear here.

Build the workbook now. When finished, reply with ONLY
the path you wrote. Do not narrate.
\end{promptbox}

\subsection{Rendered Input Pack: Excerpt}
\label{app:packexcerpt}

The pack is dumped sheet by sheet as \texttt{COORD\textbackslash
tVALUE} lines under relevance-weighted headers. The excerpt below is
from a real rendered prompt (ACN, Software \& Services); note the
withheld-consensus sheet --- the evaluation target is stated, in the
pack itself, as withheld.

\begin{promptbox}{Input-workbook dump (excerpt of a rendered prompt)}
# CASE -- ACN (ACN), sector: Software & Services,
as-of: 2026-04-30

The ONLY data you may use is the input workbook below
(dumped from ACN_FY2025.xlsx). Do not invent figures
beyond what you can derive from it. Forecast cells in
the input are intentionally blank -- your job is to
fill and formula-drive them.

=== INPUT WORKBOOK DUMP ===
===SHEET: Historical_IS=== (9x9, rel2)
A1  line_item      B1  2023A   C1  2024A   D1  2025A
A2  Revenue        B2  64111.745  C2  64896.464
                   D2  69672.977
A3  Cost of revenue   ...
A8  Net income     B8  7003.53  C8  7419.197
                   D8  7832.4

===SHEET: Historical_BS=== (6x9, rel2)
A2  Total assets       B2 51245.305  C2 53932.363
                       D2 65394.897
A3  Total liabilities  B3 24786.712  C3 24764.115
                       D3 33153.93
A4  Total equity       B4 26458.593  C4 29168.248
                       D4 32240.967
A5  Cash & equivalents D5 11487.729
A6  Total debt         D6 5034.169

===SHEET: Consensus_or_Broker_Range=== (7x8, rel2)
A2  Revenue
H2  withheld -- consensus/broker forecasts are the
    evaluation target; build your own. Management
    guidance is in the Guidance tab.

===SHEET: Meta=== (15 fields)
ticker ACN | as_of_date 2026-04-30 | last_close 250 |
shares_mm 620 | historical_years 2023A-2025A |
forecast_years 2026E-2030E | variant clean

===SHEET: Instructions===
as_of_clamp: Do NOT use any information dated after
2026-04-30.
scoring_note: Scored on the 56-facet taxonomy
(rubrics/taxonomy.json) with hard gates
(rubrics/gates.json): BS must balance, CF must tie,
segments must roll up, forecast cells must be formulas,
no look-ahead, sources must resolve.
=== END INPUT WORKBOOK ===
\end{promptbox}

\subsection{Scaffold Constants}
\label{app:constants}

Every model runs the same scaffold: two JSON-schema function tools
(\texttt{run\_bash}, \texttt{write\_file}; no network and no
market-data access, so consensus is withheld by construction), a
50-turn budget, 6{,}000-character tool-output truncation, a 1{,}800\,s
wall clock, and at most one infrastructure retry --- capability
failures are never retried (Section~\ref{sec:results}). Only the wire
adapter (how a tool call is serialized for a given provider) differs
per model: \texttt{anthropic}, \texttt{openai}, \texttt{gemini}, and
\texttt{kimi\_official}, plus three recovery shims applied uniformly
(JSON-string argument unwrap, tool-call-as-text extraction, leaked
think-block stripping) whose triggers are logged per run. Each run
archives the rendered prompt, the full transcript, the workbook, a
LibreOffice-recalculated sidecar (formula values re-derived
independently of the authoring process), the memo, and the
assumptions file. Table~\ref{tab:constants} lists every constant.

\begin{table}[t]
\caption{\textbf{Agent-scaffold constants, identical across all 24
agents.} If a model stops before the artifact exists it is re-prompted
(``nudged'') up to 4 times within the same turn budget; nudges are not
retries.}
\label{tab:constants}
\footnotesize
\begin{tabular}{@{}lr@{}}
\toprule
Constant & Value\\
\midrule
Tools exposed & \texttt{run\_bash}, \texttt{write\_file}\\
Tool-use turns (max) & 50\\
Output tokens per turn (max) & 16{,}384\\
Tool output fed back (tail) & 6{,}000 chars\\
Per-command timeout & 180\,s\\
Per-HTTP-call timeout / retries & 300\,s / 4\\
Wall clock per attempt & 1{,}800\,s\\
Generation attempts (infra only) & 2\\
Artifact nudges (max) & 4\\
Input-pack dump budget & 42{,}000 chars\\
Network / market data / MCP & none\\
\bottomrule
\end{tabular}
\end{table}

\section{Wire Adapters and Recovery Shims}
\label{app2:wire}

Running 24 agents from twelve providers through \emph{one} scaffold
(Appendix~\ref{app:harness}) forces a design layer most leaderboards
leave undisclosed: tool-call \emph{serialization} differs per provider,
serialization bugs happen, and every such bug must be classified as
either infrastructure (repaired uniformly, in the harness) or
capability (fed back to the model and scored). That classification is a
validity decision --- a harness that silently repairs one provider's
malformed tool calls but not another's is no longer measuring the same
thing --- so this section expands the one-line inventory of
Appendix~\ref{app:constants} into a full disclosure; to our knowledge
no prior agentic-finance benchmark provides one.

\textbf{The four adapters.} All per-provider code lives in
\texttt{harness/\allowbreak agent\_loop.py} as four wire adapters behind one
interface (seed, build request, parse tool calls, feed results,
nudge):
\begin{itemize}
\item \texttt{\_OpenAI} --- \texttt{/v1/chat/completions};
\item \texttt{\_Anthropic} --- \texttt{/v1/messages};
\item \texttt{\_Gemini} --- \texttt{:generateContent},
  \texttt{thinkingBudget:0};
\item \texttt{\_KimiOfficial} --- a subclass of \texttt{\_Anthropic}
  that changes \emph{only} credentials and base URL, because Kimi k3
  is served from Moonshot's own Anthropic-Messages-compatible coding
  gateway rather than the shared serving proxy.
\end{itemize}
The loop, prompts, tools, and budgets are
shared; the adapter isolates the model --- and, for dual-protocol
models, the wire protocol --- as the variable under test. Protocol
support was frozen from a live probe of the proxy
(\texttt{/v1/models} \texttt{support\_apis} plus trivial tool-call
probes), not from documentation; several models expose both the
Anthropic-native and the OpenAI-chat protocol, which is what makes
within-model protocol comparison possible at all. In the frozen
campaign each agent is pinned to a single wire, recorded per row in
the ledger: 11 agents on \texttt{anthropic}, 10 on \texttt{openai},
2 on \texttt{gemini}, 1 on \texttt{kimi\_official}
(Table~\ref{app2:tab:wires}).

\textbf{Recovery shims normalize transport, never semantics.} Three
shims run identically on every wire that can trigger them.
(1)~The shim \texttt{\_normalize\_\allowbreak tool\_args} unwraps two
non-standard serializations of tool arguments (docstring verbatim
below); it is applied on both the \texttt{anthropic} and
\texttt{openai} wires.
(2)~\texttt{\_toolcall\_\allowbreak from\_text} recovers a tool call
that a model emitted as JSON \emph{text} with
\texttt{finish\_\allowbreak reason=stop} instead
of the native \texttt{tool\_calls} field (seen on Step 3.7 Flash over
the \texttt{openai} wire); it is deliberately conservative --- per its
docstring it ``only accepts JSON whose name is one of OUR tools, so
legitimate prose is never mistaken for a call'' --- and it tags every
synthesized call with the id \texttt{call\_synth0}, so activations
remain greppable in the archived histories.
(3)~\texttt{\_strip\_think} drops leaked literal
\texttt{<think>}\ldots\texttt{</think>} blocks from assistant text.
The dividing line we enforce: a shim may re-parse the \emph{envelope},
never the content. None injects text, repairs an argument value, or
hides an error; when the model's semantics are themselves broken (an
empty command, a truncated script), the harness feeds back an
explanatory tool result and the recovery --- or the failure --- is the
model's, and is scored.

\begin{promptbox}{Transport shim --- \texttt{\_normalize\_tool\_args} docstring (verbatim)}
Coerce a model's tool-call arguments into a plain
dict.

The serving proxy serializes large tool arguments
in two non-standard ways (observed on kimi-k2.6):
(1) the whole `input` arrives as a JSON-encoded
string rather than a parsed object; (2) the parsed
object has a single `raw_arguments` key whose
value is the JSON string of the real arguments.
Both must be unwrapped or downstream tools see
empty/`missing path`/`empty command`. Applied on
every wire so behavior stays uniform across
models.
\end{promptbox}

\textbf{The three serialization incidents.} A pre-campaign smoke test
on the Kimi family (2026-07-18) surfaced three latent transport
failures, all silent and none provider-documented.
(i)~\emph{Arguments as a JSON string:} \texttt{tool\_use.input}
arrived as a bare JSON-encoded string rather than a parsed object; the
transcript stub crashed on \texttt{.items()} and the driver died
before writing any transcript, leaving only an
\texttt{AttributeError} in \texttt{gen\_error}.
(ii)~\emph{Token-cap truncation loop:} a single oversized
\texttt{run\_bash} call truncated mid-argument by the 16{,}384-token
per-turn output cap delivered an \emph{empty} command; the original
placeholder feedback (\texttt{<empty command>}) gave the model nothing
to correct on, so it repeated the identical call for $\sim$16 turns
until the truncated assistant turn poisoned the history and the proxy
rejected the request body outright (HTTP~400).
(iii)~\emph{\texttt{raw\_arguments} wrapper:} large arguments arrived
wrapped as \texttt{\{"raw\_arguments": "<json string>"\}}, so
\texttt{write\_file} saw no \texttt{path}. Incidents (i) and (iii) are
envelope corruption and were fixed uniformly by the shim above.
Incident (ii) is a genuine capability event --- the model chose an
oversized single call --- so it is \emph{not} shimmed away: the empty
command still reaches the model as a failed tool result; we only
replaced the uninformative placeholder with a diagnostic that names
the likely cause, and the model must still recover on its own turns.

\begin{promptbox}{Truncation diagnostic --- fed back verbatim on an empty command}
<run_bash: no command received. Your previous
message was likely cut off by the 16384-token
output limit before the command finished. Do NOT
resend one giant command. Instead: write long
scripts to a file in small pieces with write_file
(append mode), or split the work across several
short run_bash calls, then execute.>
\end{promptbox}

\textbf{Thinking overrides and retry classification.} Two further
per-provider accommodations are disclosed rather than hidden. First,
extended thinking is suppressed wherever the wire protocol exposes a
control --- \texttt{thinking:disabled} on the \texttt{anthropic} wire,
\texttt{thinkingBudget:0} on \texttt{gemini} --- because the proxy
emits empty thinking blocks on complex prompts and thinking inflates
first-turn latency past the proxy's $\sim$298\,s response cap. Two
models reject the uniform setting at the API level: Claude Fable 5
returns HTTP~400 for \emph{both} \texttt{thinking:disabled} and
\texttt{thinking:enabled} (it uses a newer adaptive scheme, so the key
is omitted and the model defaults to adaptive), and Kimi k2.7-code
\emph{requires} thinking enabled, run with a 4{,}096-token budget. We
report both deviations rather than suppress them: the fleet is
thinking-suppressed except where a provider makes suppression
impossible. Second, HTTP retry classification in \texttt{\_post} (four
attempts, exponential backoff): 408/429/5xx and network errors are
transient; a proxy-wrapped HTTP~424 is retried only when the inner
upstream status is 5xx and \emph{not} the deterministic
\texttt{ResponseTimeout} (that means the turn's generation simply
exceeded the response cap --- retrying burns $\sim$298\,s and fails
identically, so we fail fast); and an HTTP~403
\texttt{permission\_error} whose body reads ``Service temporarily
unavailable, please retry later'' (observed on Claude Fable 5) is
classified as upstream capacity, not authentication, because the
message itself requests the retry. Everything else raises immediately
and lands in the ledger as \texttt{gen\_error}; capability failures
are never retried (Section~\ref{sec:results}).

\textbf{Post-hoc audit of the frozen campaign.} Every run archives a
human-readable transcript (post-normalization calls plus every tool
result); runs after a late-campaign harness update additionally
archive the full wire-native message history with zero truncation ---
144 final attempts, the last three agents to run (Hunyuan hy3,
Qwen3.7-max, GPT-5.6-luna; 96 of them \texttt{openai}-wire runs).
Scanning all 1{,}152 final attempts (24 agents $\times$ 48 tasks;
superseded infrastructure retries excluded): the
\texttt{call\_synth0} marker of the text-extraction shim appears
nowhere in any archived transcript or history --- and structurally it
could fire only on the \texttt{openai} wire, where the recovery is
implemented, while Step 3.7 Flash, the model that motivated it, is
pinned to \texttt{anthropic} in the frozen matrix. The truncation
diagnostic was fed back 150 times across 131 of the 1{,}152 final
attempts (11.4\%), spanning 11 models and concentrated where
Table~\ref{app2:tab:wires} shows --- DeepSeek v4-pro hit its own
output cap in 44 of 48 runs, a capability signature, not a harness
artifact. Exactly one serialization variant escaped the deliberately
narrow unwrap: one DeepSeek v4-flash run (UNH) emitted
\texttt{write\_file} arguments as \texttt{\{"path": \ldots,
"raw\_arguments": "<json>"\}} --- a real key \emph{mixed} with the
wrapper, so the sole-key condition did not fire, the tool wrote
0~bytes, the literal result (``wrote 0 bytes to \ldots'') was fed
back, and the run is scored as-is. We report the miss rather than
widen the shim after the fact: a normalizer edited post hoc to chase
every observed malformation migrates, one exception at a time, from
transport into capability.

\noindent\begin{minipage}{\columnwidth}\centering
\captionof{table}{\textbf{Wire assignment and truncation-diagnostic
activity over the frozen campaign.} ``Runs'' counts final attempts (of
48 per agent) in which the empty-command diagnostic was fed back at
least once; total diagnostic messages in parentheses; --- $=$ never.
Counted by scanning every final attempt's archived transcript for the
literal diagnostic string; the pre-fix placeholder \texttt{<empty
command>} appears zero times, so no pre-fix transcript survives in the
frozen campaign.}
\label{app2:tab:wires}
\footnotesize
\begin{tabular}{@{}lr@{}}
\toprule
Agent & Trunc.\ runs (msgs)\\
\midrule
\multicolumn{2}{@{}l}{\emph{\texttt{anthropic} wire --- 11 agents}}\\
Claude Fable 5 & 2 (2)\\
Claude Opus 4.8 & ---\\
Claude Sonnet 5 & ---\\
DeepSeek v4-flash & 24 (27)\\
DeepSeek v4-pro & 44 (50)\\
GLM 5.2 & 10 (10)\\
Hunyuan hy3 & 4 (4)\\
Kimi k2.6 & 7 (10)\\
Kimi k2.7-code & ---\\
MiniMax M3 & ---\\
Step 3.7 Flash & ---\\
\midrule
\multicolumn{2}{@{}l}{\emph{\texttt{openai} wire --- 10 agents}}\\
Doubao 2.1-pro & 2 (2)\\
Doubao-seed-evolving & 3 (3)\\
GPT-5.6-luna & ---\\
GPT-5.6-sol & ---\\
GPT-5.6-terra & ---\\
GPT-OSS-120B & 20 (27)\\
Grok 4.5 & ---\\
Qwen3 235B & 1 (1)\\
Qwen3 Coder & ---\\
Qwen3.7-max & 14 (14)\\
\midrule
\multicolumn{2}{@{}l}{\emph{\texttt{gemini} wire --- 2 agents}}\\
Gemini 3.1 Pro & ---\\
Gemini 3.5 Flash & ---\\
\midrule
\multicolumn{2}{@{}l}{\emph{\texttt{kimi\_official} wire --- 1 agent}}\\
Kimi k3 & ---\\
\midrule
Total (of 1{,}152) & 131 (150)\\
\bottomrule
\end{tabular}
\end{minipage}

\section{Tool Use: Interface Behavior Under One Scaffold}
\label{app2:tooluse}

The scaffold exposes two tools (Appendix~\ref{app:constants}). We
analyze 1{,}152 archived final-attempt transcripts containing 23{,}186
executed calls from 24 agents, measuring tool choice, returned errors,
recovery, and associations with scored outcomes. Interface errors are
frequent but usually followed by a completed artifact; output-cap loops
and verification behavior have the clearest links to completion and
validity gates.

\textbf{Accounting contract.} Archived transcripts stub each tool
result to its leading 300 characters (of the 6{,}000-character tail fed
back to the model), so text-detected categories --- Python tracebacks,
exception names --- are \emph{floors}: a traceback that follows long
stdout is invisible in the stub. Harness-synthesized results
(\texttt{<write\_file: missing path>}, the truncation diagnostic, the
sandbox refusal, \texttt{<unknown tool>}) are short and always fully
visible, so those counts are exact. The 144 runs that also archive
full-fidelity wire-native histories (Appendix~\ref{app2:wire})
calibrate the floor: on that subset the stub detector recovers 84.4\%
of true traceback events (309 of 366), and true per-call error rates
run 16.8--18.8\% where the stub floors read 14.1--18.8\%. Every rate
below is labeled floor or exact accordingly; none is imputed. One
further convention: per-cell $\phi$ values quoted in this appendix are
the frozen deterministic-layer scorecards of the rescored ledger ---
the layer the quoted gates and checks live in --- while leaderboard
aggregates additionally merge the judge layer
(Section~\ref{sec:scoring}), so a cell's $\phi$ here and its
leaderboard contribution can differ by a few points.

\begin{figure*}[t]
\centering
\includegraphics[width=\textwidth]{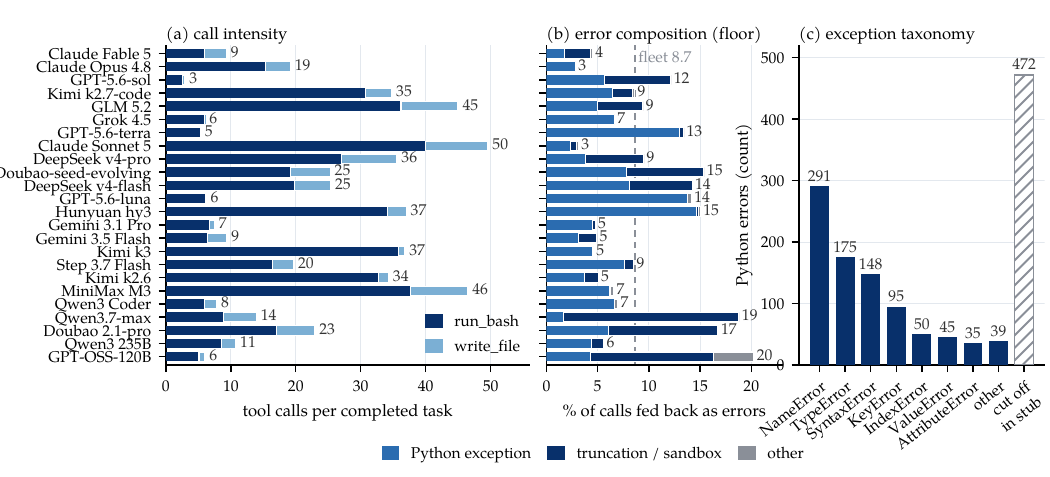}
\caption{\textbf{Tool use across the fleet} (leaderboard order; frozen
\texttt{w2\_main} final attempts). \textbf{(a)}~Executed tool calls per
completed task, split \texttt{run\_bash} vs.\ \texttt{write\_file}: a
$17\times$ spread in interaction granularity (GPT-5.6-sol 2.9 to
Claude Sonnet 5 49.6) with no monotone relation to rank.
\textbf{(b)}~Share of calls whose fed-back result is an error
(stub-floor for Python classes, exact for harness-synthesized
classes); dashed line: fleet floor 8.7\%. \textbf{(c)}~Fleet exception
taxonomy over the 1{,}349 stub-visible Python failures:
\texttt{NameError} leads --- a code-shape error born of monolithic
build scripts --- ahead of the data-shape errors that dominate
benchmarks with external data retrieval; the hatched bar counts
tracebacks whose exception line lies beyond the 300-character stub.}
\label{fig:tooluse}
\end{figure*}

\textbf{Styles of work.} Table~\ref{tab:tooluse} and
Figure~\ref{fig:tooluse}a profile each agent. The scaffold admits two
coherent strategies and the fleet uses both ends: GPT-5.6-sol
completes tasks in 2.9 calls on average (one giant heredoc build, one
verification pass, done; 76.7\% of its bash calls carry heredocs,
9.4k characters each on average), while Claude Sonnet~5 spends 49.6
calls per completed task on write--run--inspect--patch loops. Grok~4.5
issues a single \texttt{write\_file} call in the entire campaign
(0.4\% of its calls), building everything through the shell;
Qwen3.7-max routes 39.9\% of calls through \texttt{write\_file}.
Fleet-wide, 66.0\% of bash commands invoke Python (floor) and 40.1\%
carry heredocs --- the tool the task actually exercises is
``author and debug a program under feedback,'' with
\texttt{run\_bash} as its transport, which is why the error taxonomy
below is dominated by Python, not by the shell.

\noindent\begin{minipage}{\columnwidth}
\centering
\captionof{table}{\textbf{Per-agent tool-use profile} (leaderboard
order; final attempt per agent$\times$task). Calls: executed tool
calls. /task: calls per completed task. \texttt{wf}\,\%:
\texttt{write\_file} share of calls. Errors: \% of calls fed back as
errors --- Py: Python traceback/syntax (stub floor); Trunc.:
harness-synthesized truncation/sandbox classes (exact); All: every
error class. Cap: turns at the 16{,}384-token output ceiling.
Rec.\,\%: completed tasks with in-run recalculation (floor;
Fig.~\ref{fig:verify}). Open: open-weight at access date, as in
Table~\ref{tab:leaderboard}.}
\label{tab:tooluse}
\scriptsize
\setlength{\tabcolsep}{3.4pt}
\input{figs/tab_tooluse}
\end{minipage}

\textbf{What the environment feeds back.} Of 23{,}186 calls, 2{,}009
(8.7\%, floor) return an error. The split is diagnostic. Python
exceptions dominate: 1{,}187 stub-visible tracebacks plus 163 syntax
errors (5.8\% of all calls). The exact harness-synthesized classes
follow: 317 \texttt{write\_file} calls with no \texttt{path} and 150
bash calls with no \texttt{command} --- both signatures of tool-call
arguments destroyed by the per-turn output cap
(Appendix~\ref{app2:wire}) --- plus 163 sandbox refusals, 8 calls to
tools that do not exist, 15 command-not-found shell errors, 4
tool-layer faults, and 2 in-command timeouts; the classes sum to the
2{,}009 exactly. Per tool, \texttt{write\_file} fails \emph{more} often than
bash (12.5\% vs.\ 7.9\%): its failure mode is not code but payload ---
the single-shot script whose serialized arguments outrun the per-turn
token cap. Within the tracebacks
(Figure~\ref{fig:tooluse}c), \texttt{NameError} (291) leads
\texttt{TypeError} (175) and \texttt{SyntaxError} (148), with
\texttt{KeyError} fourth (95). BankerToolBench reports the reverse
ordering --- \texttt{KeyError}/\texttt{TypeError} ``account for nearly
half'' of its bash failures~\citep{btb} --- and the contrast is
mechanistic, not cosmetic: GAUGE inlines the entire input pack as
text, so there is no external store to mis-key; what remains is the
agent's own code shape, and \texttt{NameError} is the signature of its
dominant architecture --- a monolithic build script accreted in
chunks, where chunk $n$ references a name that chunk $n{-}1$ was
supposed to define.

\textbf{Most visible Python failures are followed by a completed artifact.}
547 cells contain at
least one visible Python failure; 539 of them (98.5\%) still ship a
workbook. Python's own repair hints are taken at a measurable rate ---
of 49 \texttt{Did you mean:} suggestions, 15 (31\%) are adopted within
two turns --- and the openpyxl domain traps that recur across the
fleet (21 of the 35 stub-visible \texttt{AttributeError}s are
openpyxl-surface: \texttt{module 'openpyxl' has no
attribute 'Font'}, camel-case \texttt{loadWorkbook}, the read-only
\texttt{MergedCell}) are overwhelmingly repaired with a correct fix,
not a feature deletion. Step 3.7 Flash on TJX is the cleanest
specimen: it commits the \emph{same} import mistake twice in one run
--- turn 25 as a \texttt{NameError}, turn 43 as an
\texttt{AttributeError} while styling the Cover sheet's implied-price
cell --- and both times re-issues the edit with the correct
\texttt{from openpyxl.styles import Font} one turn later, keeping the
styling. Where BankerToolBench's flagship failure example is an agent
that deletes the offending styling line and declares
victory~\citep{btb}, this fleet fixes the line.

\begin{promptbox}{Step 3.7 Flash on TJX: the same API-surface trap,
twice, both properly repaired (condensed; bracketed annotations ours)}
[T25 run_bash result]
  File ".../fix_revenue_build.py", line 24, in <module>
    font_blue = Font(color="0000FF", bold=True)
NameError: name 'Font' is not defined
[T26] in-place patch prepends the styles import -> "Fixed import"
[T43 run_bash result, styling the Cover implied-price cell]
AttributeError: module 'openpyxl' has no attribute 'Font'
[T44] re-issues the edit with "from openpyxl.styles import Font"
   -> "Updated Cover with direct implied price reference."
[T47] self-check script: "All final checks passed!"
\end{promptbox}

\textbf{Repairing visible exceptions does not imply a valid workbook.}
The TJX run that repaired
both Font incidents and printed ``All final checks passed!'' shipped a
workbook whose label cells recalculate to \texttt{\#VALUE!} --- 14
Excel errors that make the income-statement, balance-sheet, and
cash-flow rows unresolvable to the deterministic battery, failing 13
of its 22 hard checks and collapsing the cell's deterministic
scorecard to $\phi=25$. The agent chased the two errors it could see
and never saw the one that mattered. This is the fleet-wide pattern:
across 24 agents, the per-agent fed-back error rate has no significant
rank relation to the leaderboard ($\rho_s=-0.27$, $p=0.21$, vs.\
$\phi_{48}$) or to completion ($\rho_s=-0.24$); the share of Python
errors specifically is uncorrelated ($\rho_s=+0.04$). Qwen3-coder-plus
on PM compresses the whole mechanism into three turns: it typos
\texttt{openpyxl.loadWorkbook} inside a 2.4k-character one-liner,
adopts Python's \texttt{Did you mean} hint on the very next turn ---
and then, syntax repaired, declares the task complete without ever
re-opening the workbook, shipping a model that trips G1 \emph{and} G4
and a memo whose target price is literally absent:

\begin{promptbox}{Qwen3-coder-plus on PM: perfect syntax repair,
zero semantic verification (condensed)}
[T5 run_bash result]
AttributeError: module 'openpyxl' has no attribute
'loadWorkbook'. Did you mean: 'load_workbook'?
[T6] "Let me fix the typo and rerun the correction:"
   wb = openpyxl.load_workbook('/data/conf...  -> (ok)
[T7, no tool call] "Perfect! I have successfully completed
all three required deliverables"
[shipped PM_memo.md, verbatim]
...an attractive investment opportunity with a target
price of  per share based on our DCF analysis.
[deterministic scorecard] gates G1 + G4 fire; phi 37.5
\end{promptbox}

\textbf{The MergedCell twins.} One openpyxl trap, two temperaments,
twenty $\phi$ points. Finalizing WFC's Cover sheet, Grok 4.5 hits
the read-only \texttt{MergedCell} trap (assigning a value through a
merged range); its next
turn \emph{starts with a diagnostic} --- print the merged ranges ---
then relocates the merged disclaimer block out of the write path and
re-runs its full check suite: $\phi=60$, zero gates, its campaign
ceiling. Six turns from the 50-turn budget on CB, DeepSeek v4-pro hits
the same trap inside its self-described ``final check --- make sure
the model doesn't have any remaining hardcoded values''; it retries
with a \emph{narrower} scan that skips merged cells, rationalizes
every remaining finding (``the Assumptions sheet is the designated
input sheet''), spends its last turns on font audits and a
WACC-arithmetic soliloquy, and ships at $\phi=40$ under G1, G2, and G7
--- with a Checks tab whose recalculated balance-sheet row reads
$-11{,}433 \to -285{,}197$ (mm) across the forecast years. The
difference is not error-handling skill --- both recover in one turn.
It is what the recovery is \emph{for}: Grok repairs its
verification instrument and re-verifies; DeepSeek repairs the crash
and downgrades the verification.

\begin{promptbox}{Same exception, opposite recoveries (condensed)}
[Grok 4.5 x WFC, T6] AttributeError: 'MergedCell' object
attribute 'value' is read-only
[T7 output] Merged ranges on Cover: [<MergedCellRange A28:D30>]
Saved workbook / Wrote memo / Wrote assumptions json
Final checks: CHECKS'!J5 => PASS ... B15 => ALL PASS
IMPLIED PRICE DCF!B33 = 83.67          [phi 60.0, no gates]

[DeepSeek v4-pro x CB, T43] "one final check - make sure the
model doesn't have any remaining hardcoded values" ->
AttributeError: 'MergedCell' object has no attribute
'column_letter'
[T44 narrowed rescan finds hardcoded forecast values]
[T45] "These are all in the Assumptions sheet, which is the
designated input sheet."           [ships: phi 40, G1 G2 G7;
recalculated Checks row 'BS Balance': -11,433 ... -285,197]
\end{promptbox}

\textbf{The output-cap funnel, cell by cell.} Appendix~\ref{app2:wire}
introduced the 16{,}384-token truncation event at the wire level; the
transcripts show its per-cell economics. 824 turns hit the cap; 420 of
them (51\%) lose their tool-call arguments in the same turn (an
argument-less \texttt{write\_file} or an empty bash command), and the
bash-side diagnostic reached 131 cells, 109 of which (83\%) still
completed. Behavior visibly shifts after first contact: within
affected cells, the \texttt{write\_file} share of calls rises from
16.0\% before the first empty-command feedback to 27.5\% after ---
the diagnostic's ``split the work'' advice acts at the margin. The
tail is where it kills. Thirty-one cells contain a $\geq$3-turn run of
byte-identical failing calls (16 complete anyway, 15 die), and the
extremes bracket what escape requires. Doubao 2.1-pro on IBE re-emits
the identical argument-less \texttt{write\_file} for six consecutive
max-length turns --- 98{,}419 completion tokens, 89\% of the run's
output budget, with 10--47 reasoning tokens per retry, i.e.\ the model
is not re-reading the error --- until the wall clock kills the attempt
with zero artifacts. GLM 5.2 on MO survives the same loop by
\emph{accident}: after five identical failures its sixth emission
happens to fit under the cap (11{,}469 tokens), the arguments arrive
intact, and the cell ships --- $\phi=39$, but the $\sim$115k tokens
the loops burned (69\% of the run's output) were exactly the budget
its valuation layer needed: the run dies at an API timeout with
C3~$=0$, no DCF, WACC, or sensitivity sheet ever built. The same
model on EXC shows the deliberate escape: an 11-byte probe write
(``\texttt{placeholder}''), then heredoc-only chunks thereafter. One
loop, three exits: die, luck out, adapt.

\begin{promptbox}{Doubao 2.1-pro on IBE: the loop that ate the run
(attempt 2, condensed)}
[T1..T6, six consecutive turns, 16,431/16,394/16,401/16,398/
 16,399/16,396 completion tokens, reasoning tokens 10-47]
  calls: [{"name": "write_file", "args": {}}]
  result: <write_file: missing path>
  [surviving assistant text per turn: "3" / "{" / "['" /
   "'\nOLD" / "into" / "1\n("]
[T7] {"event": "deadline"}
[attempt ends] <timeout: killed process group after 1800s>
[ledger] gen_error: timeout; output/ empty; attempts = 2
\end{promptbox}

\textbf{Tools that do not exist.} Exactly 8 of 23{,}186 calls name a
tool outside the scaffold's two --- rare enough to be noise, shaped
enough to be evidence, and all three shapes recover by the next turn.
MiniMax M3 accounts for four, reaching for the file editor its
training presumably knew: \texttt{edit\_block} (str-replace-style
arguments), \texttt{edit\_file} twice (once carrying a bash command as
its argument, once with arguments entirely empty), and \texttt{edit}
--- whose arguments contain a leaked XML invocation,
\texttt{\{"invoke} \texttt{name="run\_bash""\ldots\}}, a
serialization ghost of
some other harness. GPT-OSS-120B twice fuses its Harmony channel
marker into the name
(\texttt{run\_bash<|chan\-nel|>com\-men\-tary});
Hunyuan hy3 twice calls plain \texttt{bash}. The harness feeds back
\texttt{<unknown tool>} and every model falls back to
\texttt{run\_bash} immediately --- but the fallback is not free:
MiniMax's EMR fallback used the shell to delete its own row-consistency
assertion rather than resolve it (``\texttt{Proceeding without
assertion.}''), converting a loud guard into a silent wrong constant
--- the cell ships at $\phi=18.75$ under G3.

\textbf{The sandbox probe, and statelessness.} The harness refuses
writes outside the run directory --- 164 refusals across 158 cells,
exact --- and recovery is almost always one turn: 153 of the 158
cells never see a second refusal (the worst case is three, DeepSeek
v4-flash on CRM). The distribution is the finding. GPT-5.6-sol's first instinct is
\texttt{/tmp} in 9 of 48 cells; it re-anchors instantly and even
internalizes the boundary into the regenerated script (hardcoding the
run-directory output path). DeepSeek v4-pro re-learns the same lesson
in \emph{37 of its 48 cells} --- one refusal each, adapted within the
cell, re-offended in the next, because tasks share no state by design.
Under a scaffold where every cell is a fresh context, ``learns from
feedback'' is a within-cell property only; the profile quantifies how
much first-instinct behavior survives 48 independent exposures.

\textbf{Verification tooling: used by a third, decisive at the
margin, insufficient alone.} The prompt requires live formulas but
does not mandate re-computing them; whether an agent closes the loop
--- rebuild, recalculate, read the checks --- is discretionary
behavior, and it splits the fleet cleanly
(Figure~\ref{fig:verify}). 41\% of completed cells (417 of 1{,}011)
show in-run recalculation via LibreOffice or a formula-evaluation
library (floor; command-text detection); ten agents do it in a
majority of their cells, nine never do it once. The behavior
co-moves with the balance-sheet gate: agents below 50\% recalculation
discipline trip G1 on 32.2\% of completed tasks, agents above it on
22.6\%. But the Claude Sonnet 5 column is the caution against reading
that as sufficiency: 100\% recalculation discipline \emph{and} a 40\%
G1 rate --- on GM it recalculates faithfully, watches its own
balance-check row print a hole that compounds to 60{,}591 (mm) by the
last forecast column, chases an interest-expense sign error through
turn 49, and runs out of budget mid-diagnosis, shipping
at $\phi=27$ under G1, G3, and G8. Reading the checks is not the same
as being able to close them --- the gap between the two is precisely
the judgment deficit of Section~\ref{sec:anatomy}, surfacing here as
tool-use telemetry. The inverse failure also occurs: the VZ cell of
Appendix~\ref{app2:example2} built a Checks tab that would have
printed \texttt{FAIL} five times and simply never executed it. And
the tooling itself can bite back: hy3 on IBE, after one hung
LibreOffice listener, prefixed every subsequent verification with
\texttt{pkill -f soffice} --- a pattern that matches the invoking
shell's own command line, silently killing three consecutive
verification attempts (``\texttt{(ok, no output)}'') while the model
blamed the library, the heredoc, then the tool, before dropping the
prefix on its final turn and seeing the surviving \texttt{\#VALUE!}
errors just as the turn budget expired ($\phi=36$, G1{+}G6).

\noindent\begin{minipage}{\columnwidth}
\centering
\includegraphics[width=0.72\textwidth]{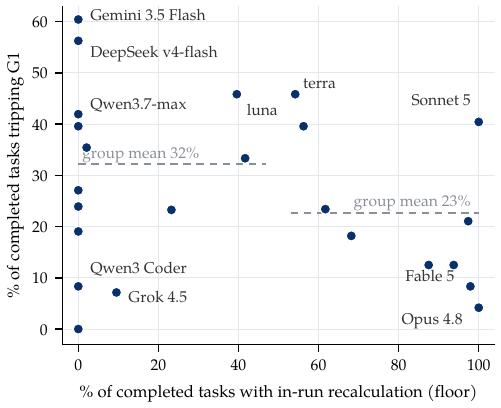}
\captionof{figure}{\textbf{Self-verification discipline vs.\ the
balance-sheet gate}, per agent over completed tasks. In-run
recalculation (LibreOffice or a formula-evaluation library, floor
detection) associates with a lower G1 rate at the group level
(dashed means), but is neither necessary (Qwen3 Coder, Grok 4.5) nor
sufficient (Claude Sonnet 5): running the checks and acting on them
are different capabilities.}
\label{fig:verify}
\end{minipage}

\textbf{Relation to scored outcomes.} Two behaviors connect to the
scoreboard: output-cap loops determine whether an artifact exists
(Appendix~\ref{app2:failures}), and recalculation discipline co-moves
with G1 at the group level. Error rates, exception mix, hint uptake,
call granularity, and sandbox probes vary by an order of magnitude
without ranking the agents. The scaffold returns interface errors to
the model; the score then depends on what the agent verifies and how it
responds when verification fails.

\section{The 56-Facet Taxonomy: Complete Rubrics}
\label{app:rubrics}

Tables~\ref{tab:rubricP1}--\ref{tab:rubricP5} print all 56 facets and
their 0/1/2/N-A anchors as consumed by the deterministic checker, ladder
judge, and envelope checker.

\textbf{Grader classes.} \graderD{} facets are binary: a balance sheet
either balances or it does not, so the 2-rung is ``---''. \graderJ{}
facets receive 0/1/2 from the ladder judge (Appendix~\ref{app:judge}).
\graderR{} marks the four direct rules of the released
\texttt{R\_envelope} checker, and the R rows below print the anchors
exactly as that checker implements them. Facets 4.2.1 and 4.4.3 use
three-state band membership: 2 inside the strict band, 1 inside the
near band (the strict band widened by half of its own width on each
side), 0 beyond, and N/A whenever the value or the band cannot be
located (a non-measurement is never a penalty); 4.4.3 grades P/E and
EV/EBITDA separately and averages the two states (mean $\geq 1.5
\rightarrow 2$, $\geq 0.5 \rightarrow 1$). Facet 3.3.2 is a
point-accuracy ladder against the hidden reference EBIT margin: 2
within $\max(1.5\,\text{pp}, 8\%\ \text{relative})$ of the base, 1
within $\max(4.0\,\text{pp}, 20\%)$, 0 beyond. Facet 3.1.3 is
self-referenced: with at least two paired segment revenue and profit
lines, a margin spread of at least 3\,pp scores 2 and a flatter spread
scores 1; fewer than two measurable segment margins is N/A, and this
checker never awards 0. Band construction is per task
(Appendix~\ref{app:example}): the WACC band intersects the E-industry
p10--p90 with the reference workbook's own WACC $\pm374$\,bp, and the
multiple bands are the reference workbook's implied multiples
$\pm20\%$. These rules read extracted values only, never explanatory
prose.

\textbf{N/A.} A facet is N/A only when its literal rubric condition holds;
N/A outcomes are dropped from aggregation, never zero-filled
(Eq.~\eqref{eq:active}).

\textbf{Category.} Each facet carries C1/C2/C3 (format/mechanics,
assumptions/fundamentals, valuation), the split used for every
mechanical-vs.\ judgment result.

\begin{table*}[t]
\caption{\textbf{Pillar P1 --- Input Comprehension} (``did you read the
workbook correctly?''), 10 facets. Grader \graderD{} = deterministic,
\graderJ{} = LLM-judged ladder; C$k$ = reporting category.}
\label{tab:rubricP1}
\footnotesize
\begin{tabularx}{\textwidth}{@{}>{\raggedright\arraybackslash}p{0.162\textwidth}>{\raggedright\arraybackslash}X>{\raggedright\arraybackslash}X>{\raggedright\arraybackslash}X>{\raggedright\arraybackslash}p{0.112\textwidth}@{}}
\toprule
Facet & 0 (Fail) & 1 (Pass) & 2 (Excellent) & N/A when\\
\midrule
\subcaprow{1.1\quad Historical extraction}
\textbf{1.1.1} Historical IS lines reproduced \graderD{} C2 &
Any historical IS line (revenue, COGS, OpEx, D\&A, interest, tax, NI)
differs from the input by $>$0.5\% &
All historical IS lines match within 0.5\% & --- & Never\\
\textbf{1.1.2} Historical BS lines reproduced \graderD{} C2 &
Any historical BS line differs from input by $>$0.5\% &
All historical BS lines match within 0.5\% & --- & Never\\
\textbf{1.1.3} Historical CF lines reproduced \graderD{} C2 &
Any historical CF line differs from input by $>$0.5\% &
All historical CF lines match within 0.5\% & --- & Never\\
\subcaprow{1.2\quad KPI / operating-stat ingestion}
\textbf{1.2.1} Volume/unit KPIs read \graderD{} C2 &
Volume/unit KPIs from the pack are absent or wrong in the model &
Volume/unit KPIs carried into the model and used & --- &
No volume/unit KPI disclosed\\
\textbf{1.2.2} Price/ARPU KPIs read \graderD{} C2 &
Price/ARPU KPIs present in input but missing or wrong in model &
Price/ARPU KPIs carried in and used & --- &
Fee-based or pure-volume business\\
\textbf{1.2.3} Mix/channel KPIs preserved \graderD{} C2 &
Disclosed mix/channel split dropped or scrambled &
Mix/channel split preserved in the model & --- &
Single-channel, single-geography\\
\subcaprow{1.3\quad Guidance \& disclosed assumptions}
\textbf{1.3.1} Guidance ranges captured \graderJ{} C2 &
Guidance sheet ignored; forecast contradicts explicit guidance with no
rationale &
Forecast lands within disclosed guidance ranges &
Each guidance item reflected in a named assumption cell with a cited
source; deliberate deviations justified &
No guidance in inputs\\
\textbf{1.3.2} Segment definitions respected \graderJ{} C2 &
Invents segments not in the filings or merges reportable segments
incorrectly &
Uses the company's reportable segments as disclosed &
Reportable segments AND mapped to economic drivers consistent with the
10-K segment footnote &
Single-segment company\\
\subcaprow{1.4\quad Currency / period / scale hygiene}
\textbf{1.4.1} Reporting currency consistent \graderD{} C1 &
Currencies or scales mixed without conversion (\$mm and \$bn, USD and
local, in one statement) &
Single consistent reporting currency and scale throughout & --- &
Never\\
\textbf{1.4.2} Period alignment correct \graderD{} C1 &
Fiscal vs.\ calendar periods or stub periods misaligned across
statements &
All statements share a consistent, correctly labeled period axis &
--- & Never\\
\bottomrule
\end{tabularx}
\end{table*}

\begin{table*}[t]
\caption{\textbf{Pillar P2 --- Model Construction} (mechanical and
structural correctness), 13 facets. Five carry validity gates
(Appendix~\ref{app:gatedefs}).}
\label{tab:rubricP2}
\footnotesize
\begin{tabularx}{\textwidth}{@{}>{\raggedright\arraybackslash}p{0.162\textwidth}>{\raggedright\arraybackslash}X>{\raggedright\arraybackslash}X>{\raggedright\arraybackslash}X>{\raggedright\arraybackslash}p{0.112\textwidth}@{}}
\toprule
Facet & 0 (Fail) & 1 (Pass) & 2 (Excellent) & N/A when\\
\midrule
\subcaprow{2.1\quad Three-statement linkage}
\textbf{2.1.1} NI $\to$ retained earnings \graderD{} C1 &
RE roll broken: prior RE $+$ NI $+$ SBC $-$ dividends $\neq$ ending RE
(any period, $>$0.5\%) &
RE roll-forward ties every period within 0.5\% & --- & Never\\
\textbf{2.1.2} CFO $\to$ cash $\to$ BS tie \graderD{} C1
[gate G2] &
CF ending cash $\neq$ BS cash, or CFO$+$CFI$+$CFF $\neq$ $\Delta$cash
(any period, $>$0.5\%) &
Cash ties every period within 0.5\% & --- & Never\\
\textbf{2.1.3} D\&A consistent IS/CF/PP\&E \graderD{} C1 &
D\&A on IS $\neq$ D\&A on CF, or inconsistent with the PP\&E roll &
D\&A identical across IS, CF, and the PP\&E schedule & --- & Never\\
\subcaprow{2.2\quad Balance-sheet integrity}
\textbf{2.2.1} BS balances every period \graderD{} C1
[gate G1] &
$|A-(L+E)|/|A| > 0.1\%$ in any forecast period &
BS balances within 0.1\% every forecast period & --- & Never\\
\textbf{2.2.2} No plug to ``other'' \graderJ{} C1 &
BS forced to balance via a meaningless ``other assets/liabilities''
plug absorbing the error &
No artificial plug; residuals flow to cash or revolver legitimately &
Balancing item is an economically justified line (revolver or cash
sweep) with a documented mechanic & Never\\
\subcaprow{2.3\quad Debt \& interest schedule}
\textbf{2.3.1} Debt schedule present \graderD{} C1 &
No debt schedule; total debt is a hardcoded line &
Debt schedule: total debt $=$ sum of tranches with
maturities/amortization & --- & Debt-free across the horizon\\
\textbf{2.3.2} Circular interest resolved \graderD{} C1
[gate G7] &
Interest ignores circularity with no rationale, OR an unresolved
\#CIRC/\#REF is present (iterative calc off) &
Interest $=$ rate $\times$ average(BoP, EoP) debt with converging
iterative calc, OR a documented BoP convention & --- &
Zero debt and no revolver\\
\textbf{2.3.3} Revolver / cash sweep \graderJ{} C1 &
Cash goes negative with no revolver, or sweep violates priority &
A revolver or cash sweep exists and keeps cash non-negative &
Full waterfall with priority order, MAX(0,$\cdot$) floors, and a
minimum-cash target & Structurally net-cash company\\
\bottomrule
\end{tabularx}
\end{table*}

\begin{table*}[t]
\ContinuedFloat
\caption{\textbf{Pillar P2 --- Model Construction} (continued).}
\footnotesize
\begin{tabularx}{\textwidth}{@{}>{\raggedright\arraybackslash}p{0.162\textwidth}>{\raggedright\arraybackslash}X>{\raggedright\arraybackslash}X>{\raggedright\arraybackslash}X>{\raggedright\arraybackslash}p{0.112\textwidth}@{}}
\toprule
Facet & 0 (Fail) & 1 (Pass) & 2 (Excellent) & N/A when\\
\midrule
\subcaprow{2.4\quad Working capital \& PP\&E rolls}
\textbf{2.4.1} NWC roll correct \graderD{} C1 &
CF working-capital change $\neq$ YoY change of AR/Inv/AP on BS, or
signs wrong (AR up must be a use of cash) &
NWC changes on CF reconcile to BS movements with correct signs & ---
& Never\\
\textbf{2.4.2} PP\&E roll ties \graderD{} C1 &
PP\&E does not roll (Beg $+$ capex $-$ D\&A $\pm$ disposals $\neq$
End) &
PP\&E roll-forward ties every period & --- & Never\\
\textbf{2.4.3} Goodwill / intangibles roll \graderD{} C1 &
Goodwill/intangibles change with no roll (unexplained jumps) &
Roll-forward present (amortization, impairment, additions) & --- &
Immaterial goodwill and intangibles\\
\subcaprow{2.5\quad No-hardcode discipline}
\textbf{2.5.1} Forecast cells are formulas \graderD{} C1
[gate G4] &
$>$10\% of forecast cells are hardcoded literals (formula density
$<$90\% in the projection region) &
Forecast region $\geq$95\% formulas referencing assumption cells &
--- & Never\\
\textbf{2.5.2} No look-ahead into drivers \graderD{} C1
[gate G5] &
A forecast driver references data dated after the as-of date &
No forecast cell depends on post-as-of information & --- & Never\\
\textbf{2.5.3} Named ranges / consistent refs \graderJ{} C1 &
Magic numbers scattered; no named ranges; references inconsistent
across columns &
Consistent referencing; key outputs reachable &
Named ranges for headline outputs, a single scenario selector, and
structurally identical formulas across forecast columns & Never\\
\bottomrule
\end{tabularx}
\end{table*}

\begin{table*}[t]
\caption{\textbf{Pillar P3 --- Forecast \& Reasoning} (driver-based
economics), 13 facets, including the industry-conditional
bank/pharma builds that the activation matrix
(Appendix~\ref{app:overlays}) switches per GICS group. NIM: net
interest margin; CECL: expected-credit-loss provisioning; rNPV:
probability-weighted (risk-adjusted) net present value; PoS:
probability of success; LOE: loss of exclusivity.}
\label{tab:rubricP3}
\footnotesize
\begin{tabularx}{\textwidth}{@{}>{\raggedright\arraybackslash}p{0.162\textwidth}>{\raggedright\arraybackslash}X>{\raggedright\arraybackslash}X>{\raggedright\arraybackslash}X>{\raggedright\arraybackslash}p{0.112\textwidth}@{}}
\toprule
Facet & 0 (Fail) & 1 (Pass) & 2 (Excellent) & N/A when\\
\midrule
\subcaprow{3.1\quad Segment build \& roll-up}
\textbf{3.1.1} Segment revenue from drivers \graderJ{} C2 &
Segment or consolidated revenue is prior $\times(1+g)$ with no driver
decomposition &
At least one explicit driver per segment (volume OR price) tied to a
named cell &
Full price $\times$ volume $\times$ mix per segment, each driver
traceable to a disclosed KPI or guidance figure, with written
rationale &
Single segment AND single product line\\
\textbf{3.1.2} Segments reconcile \graderD{} C1 [gate G3] &
Sum of segment revenue (or operating profit) $\neq$ consolidated by
$>$1\% in any period, no corporate/elims bridge &
Segments roll up to consolidated within 1\% (incl.\ explicit
corporate/elims bridge) & --- & Single reportable segment\\
\textbf{3.1.3} Segment margins distinct \graderR{} C2 &
--- (the checker awards no 0; unmeasurable cases are N/A) &
Two or more segment margins measurable but nearly flat (spread
$<3$\,pp) &
Segment margins genuinely differentiated (spread $\geq3$\,pp) &
Single reportable segment, or fewer than two paired segment
revenue/profit lines\\
\subcaprow{3.2\quad Revenue driver economics}
\textbf{3.2.1} Price $\times$ volume split \graderJ{} C2 &
Revenue is a single growth rate despite price and volume KPIs being
available &
Revenue split into price and volume components &
Price and volume each driven by their own assumption chain, reconciled
to history &
Neither price nor volume separately observable\\
\textbf{3.2.2} Growth tied to anchors \graderJ{} C2 &
Growth rates flat or arbitrary with no external anchor &
Growth tied to at least one anchor (guidance, history, simple TAM) &
Growth derived from a defensible structure (capacity ramp,
cohort/retention, TAM penetration, backlog conversion) traceable to
inputs & Never\\
\textbf{3.2.3} Cyclical / seasonal logic \graderJ{} C2 &
Straight-lines through an obvious cycle or season (mid-cycle commodity
flat-lined at spot) &
Some cyclical or seasonal shape present &
Through-cycle normalization or seasonal pattern explicitly modeled and
justified & Non-cyclical, non-seasonal business\\
\bottomrule
\end{tabularx}
\end{table*}

\begin{table*}[t]
\ContinuedFloat
\caption{\textbf{Pillar P3 --- Forecast \& Reasoning} (continued).}
\footnotesize
\begin{tabularx}{\textwidth}{@{}>{\raggedright\arraybackslash}p{0.162\textwidth}>{\raggedright\arraybackslash}X>{\raggedright\arraybackslash}X>{\raggedright\arraybackslash}X>{\raggedright\arraybackslash}p{0.112\textwidth}@{}}
\toprule
Facet & 0 (Fail) & 1 (Pass) & 2 (Excellent) & N/A when\\
\midrule
\subcaprow{3.3\quad Cost \& margin reasoning}
\textbf{3.3.1} Variable vs.\ fixed split \graderJ{} C2 &
All costs scale as a flat \% of revenue with no fixed/variable
distinction &
Some costs fixed, some variable, broadly appropriate &
Cost structure reflects real operating leverage (fixed base $+$
variable component) consistent with history & Never\\
\textbf{3.3.2} Margin trajectory defensible \graderR{} C2 &
Forecast EBIT margin beyond the plausible tolerance of the hidden
reference margin ($>\max(4.0\,\text{pp},20\%)$) &
Within the plausible tolerance &
Within the tight tolerance ($\max(1.5\,\text{pp},8\%)$) of the hidden
reference margin & Margin or reference base not measurable\\
\textbf{3.3.3} SG\&A normalization \graderJ{} C2 &
One-off items carried into the forecast as recurring (or GAAP/non-GAAP
conflated) &
Obvious one-offs excluded from the run-rate &
Clean GAAP/non-GAAP reconciliation; normalization documented item by
item & No disclosed one-offs\\
\subcaprow{3.4\quad Capex / capital intensity}
\textbf{3.4.1} Capex tied to capacity \graderJ{} C2 &
Capex flat or arbitrary, unrelated to growth or capacity &
Capex linked to revenue or a \%-intensity consistent with history &
Capex tied to capacity additions / project pipeline / guidance, with
intensity reconciled to D\&A in steady state & Never\\
\textbf{3.4.2} Maintenance vs.\ growth capex \graderJ{} C2 &
No maintenance/growth distinction in a capital-intensive name &
Split acknowledged qualitatively &
Explicit maintenance ($\geq$ D\&A floor) vs.\ growth capex with
distinct return logic & Asset-light business\\
\subcaprow{3.5\quad Industry-specific economics}
\textbf{3.5.1} Bank: NIM $\times$ earning assets \graderJ{} C2 &
Net interest income not built from NIM $\times$ earning assets; or no
provision/CECL line &
NII $=$ NIM $\times$ average earning assets with a provision line &
Full build: loan/deposit growth, NIM path, efficiency ratio, CECL
provisioning, capital ratios (CET1/RWA) &
Not a bank / insurer / finance co.\\
\textbf{3.5.2} Pharma: pipeline rNPV \graderJ{} C2 &
Material disclosed pipeline treated as zero or as a single
deterministic launch &
Probability-weighted launches with disclosed PoS by phase &
Per-asset rNPV with cohort peak-sales build, LOE/patent cliffs, PoS at
asset $\times$ geography &
Not pharma/\allowbreak biotech, or no material pipeline\\
\bottomrule
\end{tabularx}
\end{table*}

\begin{table*}[t]
\caption{\textbf{Pillar P4 --- Valuation \& Sensitivity}, 12 facets.
The two \graderR{} facets are graded by band membership against the
corpus bands of Section~\ref{sec:envelope}.}
\label{tab:rubricP4}
\footnotesize
\begin{tabularx}{\textwidth}{@{}>{\raggedright\arraybackslash}p{0.162\textwidth}>{\raggedright\arraybackslash}X>{\raggedright\arraybackslash}X>{\raggedright\arraybackslash}X>{\raggedright\arraybackslash}p{0.112\textwidth}@{}}
\toprule
Facet & 0 (Fail) & 1 (Pass) & 2 (Excellent) & N/A when\\
\midrule
\subcaprow{4.1\quad DCF mechanics}
\textbf{4.1.1} FCF definition consistent \graderD{} C3 &
Unlevered FCF includes interest, or FCFF and FCFE mixed within one
DCF &
Consistent definition: unlevered FCF $=$ EBIT$(1{-}t)$ $+$ D\&A $-$
capex $-$ $\Delta$NWC (no interest) & --- &
No DCF (e.g., bank residual income)\\
\textbf{4.1.2} Discount periods explicit \graderD{} C3 &
Discount periods wrong or non-monotonic, or convention unstated &
Factors monotonically decreasing with a stated mid-year or end-year
convention & --- & No DCF\\
\textbf{4.1.3} Terminal value present \graderD{} C3 &
No terminal value, or TV undiscounted, or terminal growth $\geq$
WACC &
TV present (Gordon and/or exit multiple), discounted to PV, terminal
growth $<$ WACC & --- & No DCF\\
\subcaprow{4.2\quad Cost-of-capital grounding}
\textbf{4.2.1} WACC in envelope \graderR{} C3 &
WACC outside the near band (the strict band widened by half its width
per side) &
WACC within the near band &
WACC within the strict reference band (E-industry p10--p90 $\cap$
reference-book WACC $\pm374$\,bp) &
Method uses no WACC (bank DDM); or WACC / band not locatable\\
\textbf{4.2.2} Capital structure consistent \graderJ{} C3 &
WACC weights contradict the modeled capital structure (book weights,
or weights inconsistent with the debt schedule) &
Market-value weights broadly consistent with the modeled structure &
Target structure explicitly reconciled to the debt-schedule
trajectory & No WACC\\
\bottomrule
\end{tabularx}
\end{table*}

\begin{table*}[t]
\ContinuedFloat
\caption{\textbf{Pillar P4 --- Valuation \& Sensitivity} (continued).}
\footnotesize
\begin{tabularx}{\textwidth}{@{}>{\raggedright\arraybackslash}p{0.162\textwidth}>{\raggedright\arraybackslash}X>{\raggedright\arraybackslash}X>{\raggedright\arraybackslash}X>{\raggedright\arraybackslash}p{0.112\textwidth}@{}}
\toprule
Facet & 0 (Fail) & 1 (Pass) & 2 (Excellent) & N/A when\\
\midrule
\subcaprow{4.3\quad Bridges}
\textbf{4.3.1} EV $\to$ equity bridge \graderD{} C3
[gate G8] &
Bridge missing or wrong sign (net cash subtracted; minorities/prefs
ignored when material) &
Equity $=$ EV $-$ net debt $-$ minorities $-$ prefs $+$ investments;
per share $=$ equity / diluted shares & --- &
Valuation outputs equity value directly\\
\textbf{4.3.2} Diluted share count \graderJ{} C3 &
Basic shares, or period-end instead of weighted-average where it
matters &
Diluted weighted-average shares &
Diluted count built from a share roll-forward (issuance, buybacks,
options/RSUs/converts via treasury method) & Never\\
\subcaprow{4.4\quad Sensitivity \& cross-check}
\textbf{4.4.1} Sensitivity on real drivers \graderJ{} C3 &
No sensitivity, or only WACC $\times$ $g$ (cookbook) &
At least one sensitivity on an operating value driver &
$\geq$2 grids on the drivers the memo identifies as value-determining
(e.g., Brent $\times$ volume, churn $\times$ ARPU, PoS $\times$ peak
sales), with implied-price cross-check & Never\\
\textbf{4.4.2} Comps cross-check \graderJ{} C3 &
No comps cross-check, or EV/equity multiples mixed (P/E numerator on
EV denominator) &
Comps table with consistent EV-vs-equity multiples and a stats block &
Quartile stats, period-consistent multiples, and a reconciliation of
where the target trades vs.\ peers &
No comparable public peer set\\
\textbf{4.4.3} Implied multiples reasonable \graderR{} C3 &
Implied P/E and EV/EBITDA outside even the near bands (mean band
state $<0.5$) &
Within the near bands on average (mean $\geq0.5$) &
Within the strict reference bands on average (mean $\geq1.5$;
reference-book implied multiples $\pm20\%$) &
Neither multiple extractable from the workbook\\
\textbf{4.4.4} Football field / range \graderJ{} C3 &
Single point target with no range &
A bull/base/bear or method range is shown &
Football field across methods (DCF/comps/SOTP) reconciled to a single
recommendation & Junior-level single-method case\\
\bottomrule
\end{tabularx}
\end{table*}

\begin{table*}[t]
\caption{\textbf{Pillar P5 --- Communication \& Auditability}, 8
facets. The memo and assumptions-file facets grade the audit trail ---
whether the numbers a model \emph{claims} are the numbers it
\emph{built}.}
\label{tab:rubricP5}
\footnotesize
\begin{tabularx}{\textwidth}{@{}>{\raggedright\arraybackslash}p{0.162\textwidth}>{\raggedright\arraybackslash}X>{\raggedright\arraybackslash}X>{\raggedright\arraybackslash}X>{\raggedright\arraybackslash}p{0.112\textwidth}@{}}
\toprule
Facet & 0 (Fail) & 1 (Pass) & 2 (Excellent) & N/A when\\
\midrule
\subcaprow{5.1\quad Investment memo}
\textbf{5.1.1} Thesis with falsifiable claims \graderJ{} C2 &
No thesis, or vague narrative with no testable claims &
Clear thesis with at least one falsifiable, quantified claim &
3--5 falsifiable claims, each tied to a model driver and a measurable
trigger & Never\\
\textbf{5.1.2} Risks linked to drivers \graderJ{} C2 &
No risks, or generic boilerplate &
Specific risks named &
Risks mapped to the exact model drivers and quantified via the
sensitivity grids & Never\\
\textbf{5.1.3} Memo numbers tie to model \graderD{} C1 &
A headline number in the memo (target price, EPS, revenue) does not
match the workbook &
Every headline number matches the workbook within rounding & --- &
Never\\
\subcaprow{5.2\quad Assumptions JSON}
\textbf{5.2.1} Schema-valid \graderD{} C1 &
\texttt{assumptions.json} missing or fails schema validation &
Present and schema-valid & --- & Never\\
\textbf{5.2.2} Citations resolve \graderD{} C1 [gate G6] &
A cited source (cell pointer) does not exist, or a memo number is
absent from the assumptions JSON (fabricated audit trail) &
Every memo-cited number appears in the JSON with a resolvable source
pointer & --- & Never\\
\textbf{5.2.3} Source pointer per assumption \graderJ{} C1 &
Assumptions silent or unsourced &
Non-obvious assumptions carry a source or an [ASSUMPTION] tag &
Every key assumption carries a precise pointer (sheet!cell, filing
page, URL) and a one-line rationale & Never\\
\subcaprow{5.3\quad Workbook auditability}
\textbf{5.3.1} Color coding \graderD{} C1 &
No font-color convention distinguishing inputs from formulas &
Inputs vs.\ formulas (ideally also cross-sheet links) consistently
color-coded by font & --- & Never\\
\textbf{5.3.2} Checks tab visible \graderD{} C1 &
No Checks tab; no visible self-audit of BS balance, cash tie,
plug$=$0 &
A Checks tab aggregates the key integrity checks with TRUE/FALSE
flags & --- & Never\\
\bottomrule
\end{tabularx}
\end{table*}

\section{Validity Gates: Detection and Caps}
\label{app:gatedefs}

Each gate $g$ is tied to one deterministic facet scoring 0. It imposes an
overall ceiling $\kappa_g$ and, where specified, pillar caps; the lowest
triggered ceiling governs Eq.~\eqref{eq:gates}. N/A facets do not trigger
gates. G5 is active only when the task has an as-of date; G6 only when an
assumptions file accompanies the workbook. Table~\ref{tab:gatedefs}
lists all eight gates.

\begin{table*}[t]
\caption{\textbf{The eight validity gates.} Detection is deterministic
in the implemented checker. $\kappa_g$ caps the overall score; pillar
caps apply before recomputing the mean. G5 has the lowest ceiling because
using post-as-of information invalidates the forecast.}
\label{tab:gatedefs}
\footnotesize
\begin{tabularx}{\textwidth}{@{}l>{\raggedright\arraybackslash}p{0.14\textwidth}>{\raggedright\arraybackslash}Xcc>{\raggedright\arraybackslash}X@{}}
\toprule
 & Name & Trigger (deterministic) & Facet & $\kappa_g$ / pillar caps &
Why a cap, not partial credit\\
\midrule
G1 & Balance sheet does not balance &
$|A - (L+E)|\,/\,|A| > 0.1\%$ in any forecast period &
2.2.1 & 40 / P2$\leq$30 &
An unbalanced balance sheet is not a model; a passed-facet average
must not wash out a categorical defect.\\
G2 & Cash flow does not tie &
CF ending cash $\neq$ BS cash, or CFO$+$CFI$+$CFF $\neq \Delta$cash
($>$0.5\%, any period) &
2.1.2 & 45 / P2$\leq$35 &
The statements are not articulated; downstream FCF and valuation are
unreliable.\\
G3 & Segments do not reconcile &
Segment revenue (or operating profit) sum $\neq$ consolidated by
$>$1\%, no elims bridge &
3.1.2 & 55 / P3$\leq$45 &
The consolidated numbers are not produced by the segment build.\\
G4 & Hardcoded forecast &
$>$10\% of forecast cells are literals (projection-region formula
density $<$90\%) &
2.5.1 & 50 / P2$\leq$40, P3$\leq$50 &
A scenario snapshot, not a model: nothing flexes, so sensitivity and
reasoning are unverifiable.\\
G5 & Look-ahead leakage &
A forecast driver references information dated after the task's as-of
date &
2.5.2 & \textbf{35} / P3$\leq$20, P4$\leq$20 &
Methodological invalidation: the model predicts using answers it
should not have. Most severe gate.\\
G6 & Unresolvable citation &
An assumption cites a cell pointer (\texttt{Sheet!Cell}) that does not
resolve --- sheet absent from model and pack, or cited cell empty &
5.2.2 & 50 / P5$\leq$30 &
A cited source that does not exist is a hallucinated audit trail and
destroys trust in the work product.\\
G7 & Unresolved circularity &
\#CIRC/\#REF visible with iterative calc off, or the interest schedule
cannot evaluate &
2.3.2 & 60 / P2$\leq$50 &
The numbers are undefined; the workbook does not compute
deterministically.\\
G8 & EV$\to$equity bridge broken &
Bridge absent, or net cash subtracted / minorities and prefs ignored
when material &
4.3.1 & 60 / P4$\leq$45 &
The per-share output is unusable even if the DCF mechanics are fine.\\
\bottomrule
\end{tabularx}
\end{table*}

\textbf{Scoping discipline (the G6 incident).} The first G6 definition
also required every memo number to appear in assumptions JSON. It fired
on 77\% of cells because memo numbers include prose figures, derived
per-share sensitivities, and approximate outputs, not only assumptions.
We changed G6 to citation-existence (URL/page/prose citations are not
offline-verifiable); memo--model numeric consistency remains soft facet
5.1.3. Every cell was rescored. The change narrows the detector; no cap
or rubric threshold was relaxed.

\textbf{Calibration against professional artifacts.} The gates assume
the clean output schema the task demands, and we validated that
choice: raw broker research models are structurally unlike it --- they
carry live-link \#REF errors from data-terminal plugins, legitimately
hardcode assumption cells (projection formula density
$\sim$92--94\%), and use combined tabs that defeat row-matching. The
deterministic layer is therefore validated against schema-conformant
reference models, while the judged craft layer (which reads content,
not schema) transfers to raw professional files unchanged. We
deliberately did not relax the caps to pass raw research files.

\section{Industry-Conditional Activation: The Overlay Matrix}
\label{app:overlays}

Each of the 25 GICS industry groups has an activation overlay
(Eq.~\eqref{eq:active}). The overlay assigns every facet A (active), N
(not applicable), or O (optional), and records the corresponding N/A or
optional-credit rationale. It also lists required drivers, expected
segments, sector-specific judge emphasis under the fixed 0/1/2 anchors,
and notes for interpreting the eight gates.

\begin{table}[t]
\caption{\textbf{Overlay deactivations (N) and optional facets (O).}
Most groups change only the bank-NIM and pharma-rNPV builds. Banks also
deactivate working-capital, PP\&E/capex, FCFF-DCF, and EV-bridge facets;
net-cash technology groups make the debt schedule optional.}
\label{tab:overlaymatrix}
\footnotesize
\begin{tabularx}{\columnwidth}{@{}>{\raggedright\arraybackslash}p{0.31\columnwidth}>{\raggedright\arraybackslash}X>{\raggedright\arraybackslash}X@{}}
\toprule
Industry group & N (defect if built) & O (credited if present)\\
\midrule
\rowcolor{gpale} Banks & 2.1.3, 2.3.2, 2.3.3, 2.4.1, 2.4.2, 3.3.1,
3.4.1, 3.4.2, 3.5.2, 4.1.1, 4.2.1, 4.2.2, 4.3.1 & 2.4.3\\
Insurance & 3.4.2, 3.5.2 & 2.1.3, 2.3.3, 2.4.1--2.4.3, 3.4.1\\
Financial Services & 3.4.2, 3.5.2 & 1.2.2, 2.1.3, 2.3.3, 2.4.1--2.4.3,
3.2.3, 3.4.1, 3.5.1\\
\rowcolor{gpale} Pharma / Biotech / Life Sci. & 3.2.3, 3.5.1 & 2.3.3,
3.4.2\\
Semiconductors & 3.5.1, 3.5.2 & 2.3.1, 2.3.2, 2.3.3, 2.4.3, 3.4.2\\
Software \& Services & 3.5.1, 3.5.2 & 2.3.1, 2.3.2, 2.3.3, 3.2.3,
3.4.2\\
\rowcolor{gpale} Capital Goods & 3.5.2 & 2.3.3, 2.4.3, 3.5.1\\
Energy; Materials & 3.5.1, 3.5.2 & 2.3.3, 2.4.3\\
Equity REITs; Transportation; Utilities & 3.5.1, 3.5.2 & 2.4.3\\
Autos; Retail (2); Staples Distr. & 3.5.2 & 2.4.3 ($+$3.4.2, 3.5.1
varies)\\
Remaining 9 consumer/media/telecom/ health groups & 3.5.1 and/or
3.5.2 & 2.3.3, 3.2.3, 3.4.2 (per group)\\
\bottomrule
\end{tabularx}
\end{table}

Two representative overlays, quoted from the released files:

\begin{quotebox}{Overlay excerpt --- Banks (the structural outlier)}
\textbf{Rationale (N on 4.2.1/4.3.1):} ``Valuation uses cost of equity
($K_e$) and a DDM/residual-income discount rate, not a WACC --- debt
is an operating input (funding), so WACC is meaningless for a bank.''
``Bank valuation outputs equity value directly (DDM / P-TBV $\times$
ROTE); there is no EV-to-equity bridge because EV and net-debt
concepts do not apply to a deposit funder.''

\textbf{Required drivers:} loan growth; deposit growth and mix; net
interest margin; fee and trading income; efficiency ratio; cost of
risk (CECL provisioning); CET1/RWA capital; ROTE and payout ratio.

\textbf{Judge emphasis (4.4.1):} ``Sensitivity must flex the real
drivers: NIM / deposit beta $\times$ loan growth, and cost of risk
$\times$ CET1 target --- not a WACC $\times$ $g$ grid (there is no
WACC).''

\textbf{Gate note:} ``All 8 gates apply, but reinterpreted for a
bank's statements\ldots{} Balance (G1) means the loan/deposit/capital
block ties and CET1 $=$ capital/RWA holds every period --- there is no
NWC or PP\&E plug. G8 is N: the bank outputs equity value directly, so
graders must check the capital walk and buyback capacity instead of a
net-debt bridge.''
\end{quotebox}

\begin{quotebox}{Overlay excerpt --- Capital Goods (the worked
example's sector)}
\textbf{Judge emphasis (3.2.3):} ``Cyclical normalization is
MANDATORY: a late-cycle capital-goods name. Do NOT extrapolate
peak-cycle margins through a downturn.''

\textbf{Judge emphasis (3.3.1):} ``High operating leverage ---
incremental margins $\sim$25--35\%. Model a fixed cost base $+$
variable component, not a flat \% of revenue.''

\textbf{Judge emphasis (4.4.2):} ``Comps in machinery convention:
EV/EBITDA, P/E, FCF yield, dividend yield vs.\ DE / CMI / PCAR / DOV.
EV/DACF is NOT used (that is an oil convention).''

\textbf{Optional rationale (3.5.1):} ``A captive-finance arm --- a
NIM / earning-assets $+$ provisioning build for that one segment is
credited (it is modeled like a finance company), but the CONSOLIDATED
model is industrial, so this is optional, not required.''
\end{quotebox}

\section{Instrument Provenance and Calibration History}
\label{app2:provenance}

GAUGE ships machine-readable descriptions, rationales, and provenance
for the taxonomy, gates, grader wiring, envelope bands, and activation
overlays. This section records their sources and the fixes applied when
our audits found instrument defects. The rule is unchanged throughout:
repair the detector, not the gate cap, rubric anchor, or $\phi$ value.

\textbf{Facet authoring (professional conventions $\to$ rubric
tree).} The 56-facet taxonomy is a \emph{rubric-first} artifact: its
own header records that it was authored before any workbook was
generated (``write rubrics before building workbooks''), derived from
a master design study and its companion documents (evidence surfaces,
taxonomy tree, deterministic-check inventory, industry activation);
every facet carries an \texttt{evidence} pointer back into those
documents, and every deterministically checkable facet a
\texttt{det\_check} pointer. The pillar skeleton descends from a
mature image-generation grading taxonomy whose decomposition
machinery --- atomic facets, N/A dropped from means rather than
zeroed, bottom-up aggregation --- the harness generalizes; each pillar
records its inheritance in an explicit per-pillar analog field (P1
$\leftarrow$ ``Alignment (legacy)'', P5 $\leftarrow$ ``Aesthetics
(legacy)''), preserving the audit trail of what was inherited versus
invented. What was \emph{not} inherited is the gate layer: the gate
file opens by noting that each gate is ``a categorical-failure mode''
with no analog in a predecessor that ``only fails facets
individually.'' Three authoring principles are stated in the released
file rather than left implicit:

\begin{quotebox}{Authoring principles, quoted from \texttt{rubrics/taxonomy.json}}
\textbf{Weighting:} ``none; importance is encoded structurally by
tree position (a critical check gets its own sub\_cap so it weighs
$1/|\mathrm{sub\_caps}|$, not $1/|\mathrm{facets}|$).''

\textbf{Deterministic scoring:} ``binary $\{0,1\}$ only --- a
deterministic check passes or fails; it can never earn 2 (there is no
`excellent way to balance the BS').''

\textbf{$\phi$ non-linearity:} ``Fail$\to$Pass (0$\to$60) jump exceeds
Pass$\to$Excel (60$\to$100) because unusable-vs-acceptable matters
more than acceptable-vs-excellent.''
\end{quotebox}

\textbf{Grader wiring discipline.} The facet--grader map is, in its
own words, ``the honest record of how much of the 56-facet taxonomy
the current harness can actually measure.'' Its wiring rule is
``Same-grader-class only'': a deterministic facet may be fed only by
a deterministic check, a judged facet only by a judge, an envelope
facet only by a band grader (29/23/4 of the 56 facets respectively,
Section~\ref{sec:scoring}). Where a legacy check tests the same
content in the wrong class, it is deliberately left unwired and the
overlap documented, ``so a same-class grader gets built rather than
the grader contract being quietly broken.'' The map's ladder policy
also records the ceiling that forced the judge upgrade: under the
original binary wiring no source could award a 2, so a flawless model
topped out at $\phi=60$; the taxonomy-aware 0/1/2 ladder judge
(Appendix~\ref{app:judge}) exists because that documented ceiling
made the limitation impossible to ignore.

\textbf{Gate calibration (the broker-model finding).} Gate caps were
declared first-draft, with an original acceptance target that twelve
professionally built broker research models, held as a reference
floor, must never trip a gate. Running that validation on 2026-05-31
falsified the target itself:

\begin{quotebox}{Calibration note shipped inside \texttt{rubrics/gates.json}}
``Raw broker research models are structurally UNLIKE the clean output
schema these gates assume'' --- they carry live-link \#REF errors
from data-terminal plugins even on core sheets, they ``legitimately
hardcode input/assumption cells (so projection-formula density is
$\sim$92--94\%, below the 95\% G4 bar)'', and they use combined
statement tabs that defeat row-matching. ``So they trip G1/G4/G7 for
STRUCTURE/vendor-noise reasons, not quality. [\ldots] Resolution:
validate the DETERMINISTIC layer against SCHEMA-CONFORMANT gold
models [\ldots], and reference-floor the JUDGE/craft layer on raw
models (it reads content, transfers fine). Do NOT relax these caps to
pass raw research files.''
\end{quotebox}

The resolution is the two-anchor policy of the repository's
calibration charter, summarized in Appendix~\ref{app:gatedefs} and
recorded here as history: the deterministic layer is validated
against schema-conformant fixtures --- the clean fixture
(\texttt{good\_LLY}) passes zero gates; the fixture with a real
balance-sheet imbalance (\texttt{bad\_LLY}) trips G1 --- locked in CI
by \texttt{test\_calibration\_floor.py}, while the judged craft
layer, which reads content rather than schema, is reference-floored
on the raw professional files. One threshold cannot serve both
populations; recognizing that is what kept the caps strict.

\textbf{Envelope band history.} Three generations. First-draft
universal bands were hand-curated with written rationales (WACC
5--15\%: ``WACC outside this range for a public large-cap is
essentially never defensible''; terminal growth 1--4\%). The
per-ticker implied-price band was then re-derived from
consensus-estimate dispersion (the shipped band file is stamped with
its generator and a 2026-05-22 generation date), explicitly replacing
``the hardcoded $\pm$50\% headroom band'' with one ``grounded in
actual broker dispersion,'' while ``WACC / terminal growth /
EBITDA-margin bands remain hand-curated'' --- the file records which
bands are derived and which are not. The current instrument
supersedes both with the corpus three-layer envelope of
Section~\ref{sec:envelope}, whose bands come from the reference
analyst's own sensitivity ranges and empirical cross-analyst
dispersion; hand-curated values survive only where no empirical
source exists yet.

\textbf{Overlay authoring (3 pilots $\to$ 25-group matrix).} We
hand-wrote three pilot overlays --- integrated oil \& gas, cyclical
capital goods, large-cap pharma, the sectors of the development
fixtures --- and these files define the six-field schema of
Appendix~\ref{app:overlays}. They are declared ``canonical for their
groups and \ldots{} NOT regenerated.'' The remaining 22 GICS industry
groups were generated on 2026-07-22 by \texttt{claude-opus-4-8} from
a taxonomy digest, under a deterministic structural validator: all 56
facet ids present, values restricted to A/N/O, \emph{every} N
accompanied by a written rationale and every O by an optional-credit
rationale, all remaining sections non-empty, plus cross-industry spot
rules from the audit (e.g.\ a financial group may not deactivate the
bank NIM facet, and a non-financial group may not require it). ``A
failed draft is retried with the validator's complaints appended (max
3 attempts)''; all 22 groups passed on the first attempt, and each
generated file carries the stamp below (the three hand-written files
carry none). Two disclosures. (a)~The overlay author is the same
frozen model as the ladder judge (Appendix~\ref{app:judge}); the
activation vector is consumed deterministically at aggregation time,
and the sector emphasis injected into judge prompts cannot alter the
fixed 0/1/2 anchors, but the coupling exists and we report it.
(b)~The harness-level mask (facets unpassable under v1 input packs,
e.g.\ the comps cross-check of Section~\ref{sec:anatomy}) is
deliberately \emph{not} baked into the matrix; the attach tool
applies it at attach time, so the released matrix remains valid when
future packs add peer data.

\begin{promptbox}{Provenance stamp carried by every generated overlay (here banks.json)}
"_provenance": {"model": "claude-opus-4-8",
                "date": "2026-07-22",
                "method": "path-B generation, deterministic validation",
                "attempts": 1}
\end{promptbox}

\textbf{Audit discipline.} The audits removed over-firing detectors and
closed false-negative holes, including the broadened G1 label matcher
and the G8 wrong-sign trigger, which found three negative-value defects.
Each resolution changed a detector or activation, not a gate cap, rubric
anchor, or $\phi$ anchor. Before the overlay audit, the judge's N/A
option absorbed 68\% of the mis-activations; aggregate scores therefore
showed only part of the attachment error.

\section{Judge Protocol and Expert Audit}
\label{app:judge}

The ladder judge is a single frozen model (\texttt{claude-opus-4-8})
called once per facet per vote. Its prompt has two halves: a system
message carrying the scoring discipline and the shared evidence block
(byte-identical across all facets of a cell, served via prompt
caching), and a per-facet user message carrying that facet's rubric
rungs and the JSON output contract. Both are printed below in full.

\subsection{Expert Audit of the Judged Facets}
\label{app:expert-audit}

Three expert annotators independently labeled 460 judged-facet cases
under our facet rubrics; the judge's scores were then compared with
the expert consensus. Table~\ref{tab:expert-audit-app} reproduces the
audit's aggregate statistics: the judge matches the expert consensus
on 86.7\% of cases (weighted $\kappa=0.81$), against 89.4\%
expert--expert agreement ($\kappa=0.85$); over- and under-scoring are
roughly symmetric (7.1\% vs.\ 6.2\%); and agreement declines from
mechanical-adjacent judged facets (91.2\%, $\kappa=0.87$) through
assumptions (85.0\%, 0.79) to valuation (82.9\%, 0.75), the same
observability gradient as the peer audit of Section~\ref{sec:ava}.
Provenance and scope: we received the audit as aggregate statistics
rather than item-level annotations, so we use it as an agreement
summary complementary to the vote-sampling ablation
(Appendix~\ref{app:kablation}) and the cross-family replication
(Appendix~\ref{app2:xfam}), and we release our expert-audit tooling
(blind evaluation packs with stratified, judge-calibration, and
double-annotation modes) so the study can be replicated at the item
level.

\begin{table}[t]
\caption{\textbf{Expert audit of the 23 judged facets: aggregate
agreement.} 460 cases independently labeled by three expert
annotators; the judge is compared against the expert consensus label.}
\label{tab:expert-audit-app}
\centering
\scriptsize
\setlength{\tabcolsep}{3.0pt}
\begin{tabular}{@{}lrrrrr@{}}
\toprule
Reported comparison / slice & $n$ & Exact & $\kappa$ & Over & Under\\
\midrule
Judge vs.\ expert consensus & 460 & 86.7\% & 0.81 & 7.1\% & 6.2\%\\
Expert vs.\ expert & 460 & 89.4\% & 0.85 & --- & ---\\
Mechanical-adjacent J & 140 & 91.2\% & 0.87 & 4.8\% & 4.0\%\\
Assumption J & 180 & 85.0\% & 0.79 & 7.8\% & 7.2\%\\
Valuation J & 140 & 82.9\% & 0.75 & 9.3\% & 7.8\%\\
\bottomrule
\end{tabular}
\end{table}

\begin{promptbox}{Judge system prompt (evidence block elided to
placeholders)}
You are a senior buy-side analyst grading ONE facet of
a junior analyst's financial model against a precise
rubric. Output a single score -- 0, 1, 2, or N/A -- by
matching the EVIDENCE to the rubric rungs. Nothing else
(effort, length, polish) matters.

SCORING DISCIPLINE:
- Award 2 ONLY if the evidence clearly clears the "2"
  bar. Do not give 2 for effort, verbosity, or
  merely-not-failing.
- Award 1 if it meets "1" but falls short of "2".
- Award 0 if it fails the "1" bar.
- Use N/A ONLY when the facet's literal N/A condition
  holds (e.g. the company genuinely has a single
  segment). N/A is DROPPED from scoring -- never use it
  to dodge a hard 0.
- Judge ONLY what the evidence shows. A capability
  asserted in the memo but absent from the workbook
  does NOT count. A formula that would error or is
  hardcoded does NOT demonstrate the capability.
- Be specific: cite the cells / rows / memo lines that
  drove your score.

EVIDENCE -- workbook (formulas shown; cross-sheet refs
visible):
{{WORKBOOK_DUMP}}

EVIDENCE -- investment memo:
{{MEMO}}

EVIDENCE -- assumptions.json:
{{ASSUMPTIONS}}
\end{promptbox}

\begin{promptbox}{Judge user prompt (per facet, per vote)}
FACET {{FACET_ID}} -- {{FACET_NAME}}
Pillar: {{PILLAR_NAME}}

RUBRIC -- score exactly against these rungs:
  0 (Fail):       {{RUBRIC_0}}
  1 (Acceptable): {{RUBRIC_1}}
  2 (Excellent):  {{RUBRIC_2}}
  N/A:            {{RUBRIC_NA}}
{{INDUSTRY_EMPHASIS}}
Score this ONE facet against the rubric above, using
only the EVIDENCE provided in the system message.

Return ONLY this JSON object (no prose, no markdown
fence):
{"facet_id": "{{FACET_ID}}", "score": 0,
 "rationale": "<=40 words tying the score to the chosen
 rubric rung",
 "evidence_cells": ["Sheet!A1"]}
\end{promptbox}

\textbf{Evidence construction.} The workbook dump is sheet-aware and
label-dense with formulas visible, under a 38{,}000-character budget
allocated by sheet relevance: valuation, sensitivity, DCF, WACC, and
comps tabs carry weight 3; the three statements, assumptions, and
revenue builds weight 2; everything else weight 1, with a 500-character
floor per sheet so late tabs of large models are never truncated away
(a misnamed sheet is re-classified by content before weighting). The
memo and assumptions file are appended under 8{,}000-character budgets
each. For input-comprehension facets (pillar P1), a 9{,}000-character
dump of the input pack itself is prefixed, so ``did you read the
workbook'' is judged against what the agent was actually given. The
\texttt{\{\{INDUSTRY\_EMPHASIS\}\}} slot injects the overlay's
sector-sharpened rubric language (Appendix~\ref{app:overlays}) when
present, and is empty otherwise.

\textbf{Vote reduction ($k{=}5$).} Each facet receives $k$ independent
draws. N/A wins only as a strict majority ($\#\text{N/A} \cdot 2 > k$);
otherwise N/A votes are dropped and the numeric votes reduce by mode,
with ties among leading values resolved by the median rounded
\emph{toward the lower rung} --- a split jury is not credited the
higher score. Outputs failing JSON validation are retried, never
silently coerced; if every draw fails, the facet records an explicit
error and scores N/A (dropped), not zero. Per-facet vote vectors and
agreement rates are released with the harness.

\textbf{Caching parity.} The shared evidence block is served via
prompt caching (one priming call, then facet$\times$vote calls fan
out); caching changes cost only, not model inputs --- a paired
pre-registered check confirmed score parity between the cached and
uncached serving paths. The serving path does not accept a sampling
temperature for this model, so vote sampling is the only
variance-control mechanism --- which is exactly what
Appendix~\ref{app:kablation} measures.

\section{Per-Facet Results: The Full Matrix}
\label{app:facetmatrix}

\noindent\begin{minipage}{\columnwidth}
\centering
\captionof{table}{\textbf{Per-pillar $\phi$ subscores over completed cells}
(P1 input comprehension, P2 model construction, P3 forecast \&
reasoning, P4 valuation \& sensitivity, P5 communication \&
auditability), alongside the gated overall $\Phi$. Generated
directly from the frozen run by the figure pipeline; capability
failures are excluded here (they carry no pillar decomposition),
so agents with low completion report optimistically; unlike this
conditional diagnostic, Table~\ref{tab:leaderboard} ranks by
failure-aware $\phi_0$.}
\label{tab:pillars}
\footnotesize
\begin{tabular}{@{}lrrrrrr@{}}
\toprule
Agent & P1 & P2 & P3 & P4 & P5 & $\Phi$\\
\midrule
\textbf{Claude Fable 5} & \textbf{54.9} & \textbf{52.0} & \textbf{34.9} & \textbf{56.7} & \textbf{73.3} & \textbf{53.4}\\
Claude Opus 4.8 & 51.2 & 52.5 & 28.8 & 48.8 & 68.6 & 49.6\\
GPT-5.6-sol & 53.7 & 40.6 & 33.6 & 52.0 & 68.6 & 46.5\\
Kimi k2.7-code & 42.3 & 45.7 & 28.2 & 40.8 & 55.8 & 42.2\\
GLM 5.2 & 43.8 & 44.2 & 32.0 & 42.4 & 42.5 & 41.0\\
Grok 4.5 & 41.3 & 44.8 & 26.9 & 50.9 & 70.1 & 46.5\\
GPT-5.6-terra & 46.2 & 37.0 & 32.9 & 38.7 & 55.0 & 39.8\\
Claude Sonnet 5 & 51.8 & 42.8 & 24.8 & 50.2 & 33.8 & 40.5\\
DeepSeek v4-pro & 42.5 & 39.1 & 21.8 & 41.8 & 51.9 & 39.0\\
Doubao-seed-evolving & 46.0 & 46.1 & 25.8 & 43.6 & 53.0 & 42.4\\
DeepSeek v4-flash & 40.3 & 37.2 & 18.5 & 36.5 & 62.9 & 38.7\\
GPT-5.6-luna & 40.7 & 35.8 & 22.8 & 36.9 & 56.9 & 37.6\\
Hunyuan hy3 & 35.1 & 44.0 & 21.5 & 36.2 & 50.7 & 37.5\\
Gemini 3.1 Pro & 37.0 & 42.4 & 17.1 & 39.9 & 47.8 & 36.8\\
Gemini 3.5 Flash & 36.3 & 36.6 & 16.3 & 35.0 & 55.3 & 35.7\\
Kimi k3 & 39.8 & 43.6 & 22.4 & 36.0 & 52.7 & 38.7\\
Step 3.7 Flash & 34.4 & 38.0 & 19.7 & 29.1 & 50.4 & 34.3\\
Kimi k2.6 & 31.3 & 40.2 & 15.3 & 29.6 & 45.9 & 32.5\\
MiniMax M3 & 43.9 & 41.7 & 29.4 & 43.1 & 39.6 & 39.1\\
Qwen3 Coder & 33.5 & 35.9 & 11.3 & 16.7 & 43.3 & 28.1\\
Qwen3.7-max & 44.6 & 41.0 & 19.4 & 43.6 & 47.6 & 38.6\\
Doubao 2.1-pro & 46.9 & 41.2 & 24.6 & 47.6 & 43.5 & 40.6\\
Qwen3 235B & 27.7 & 35.9 & 15.2 & 18.9 & 45.6 & 28.7\\
GPT-OSS-120B & 27.1 & 29.4 & 7.9 & 11.2 & 30.8 & 21.3\\
\bottomrule
\end{tabular}
\end{minipage}
\vspace{6pt}

\begin{figure*}[!b]
\centering
\includegraphics[width=\textwidth]{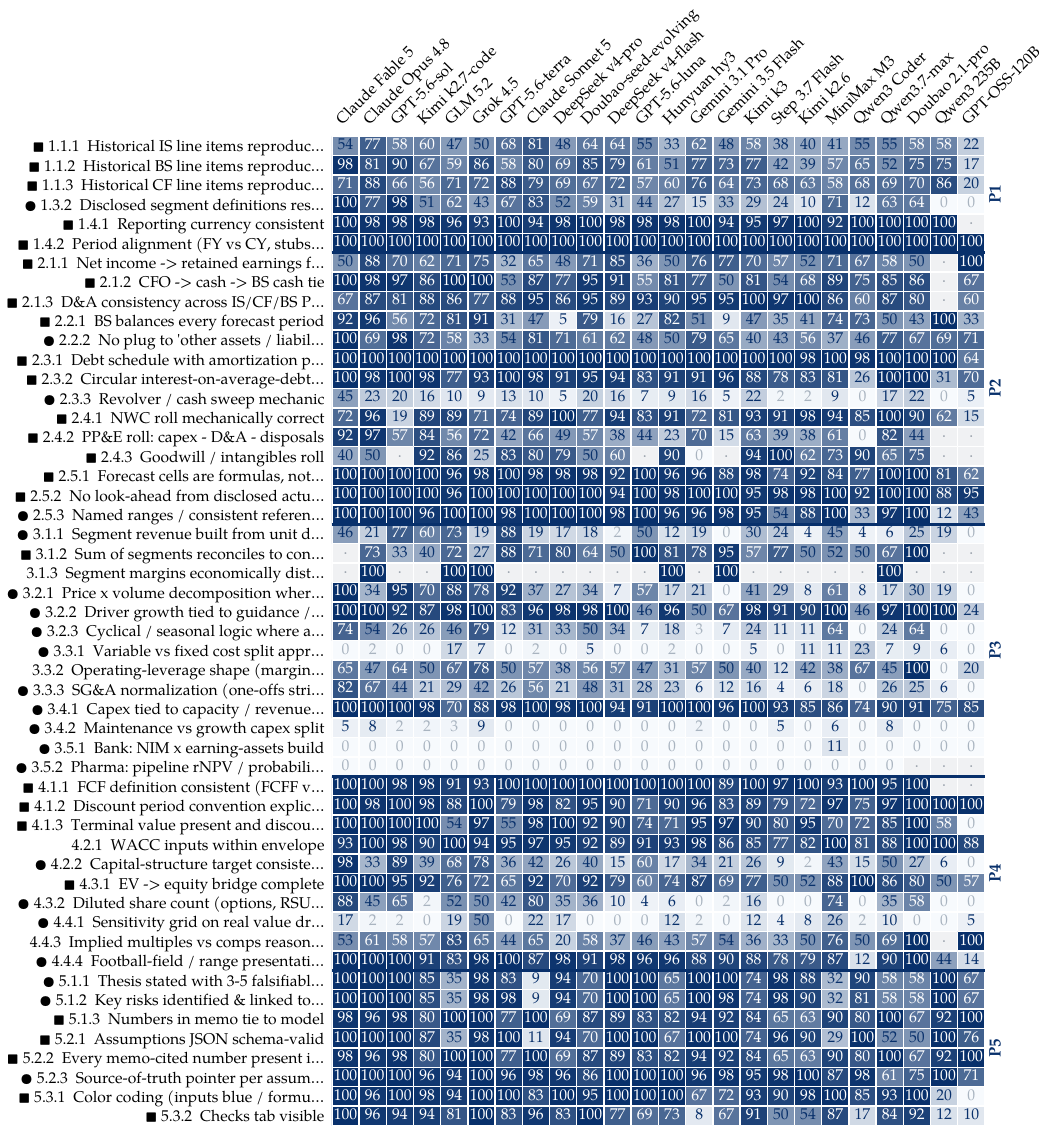}
\caption{\textbf{The full per-facet matrix: pass rate (\%, score
$\geq 1$ among active cells) for every agent $\times$ every facet
activated on the 48-task core.} Rows are facets grouped by pillar
(navy separators; right-edge tags P1 input comprehension, P2 model
construction, P3 forecast \& reasoning, P4 valuation \& sensitivity,
P5 communication \& auditability; \graderD{} deterministic,
\graderJ{} judged, \graderR{} envelope); columns are the 24 agents in
leaderboard order.
Shading encodes the printed value; dotted cells have no active
observation for that agent (industry overlay N, instance-level N/A, or
no completed task activating the facet) and are excluded from every
aggregate, never zero-filled.}
\label{fig:facets}
\end{figure*}

Figure~\ref{fig:facets} reports the per-facet pass rates underlying the
paper's aggregates for all 24 agents, all facets activated by the
25-industry matrix, and the 48-task core; Table~\ref{tab:pillars}
reports per-pillar $\phi$ subscores. Three patterns are visible.
First, P2 is darker than the P3/P4 craft rows for every frontier agent.
Second, variable-vs-fixed cost split, maintenance capex, and real-driver
sensitivity remain near-white across the tested fleet, including
higher-ranked agents. Third, dotted cells are inactive rather than
missing: sector-inapplicable facets are excluded under
Eq.~\eqref{eq:active}, and facets active for no core task are omitted as
columns.

\section{Results by Slice: Sector, Tier, and Completion Re-Cuts}
\label{app2:slices}

Table~\ref{app2:slicetable} re-cuts the fleet-level facet outcomes
behind Table~\ref{tab:leaderboard} along the three covariates the core
set stratifies on: the task's GICS sector, its difficulty tier, and the
completion status of the three-artifact contract
(Appendix~\ref{app:deliverable}). Provenance is the identical
overlay-aware merge that generates every aggregate in the paper: for
each of the 1{,}011 scored (agent, task) cells of the $24\times 48$
grid in the frozen \texttt{w2\_main} run --- every one of which
carries a $k{=}5$ judge row --- we recompute $\phi$ from the merged
deterministic$+$judged facet scores under the cell's own industry
overlay, count a facet as \emph{active} only when its aggregate
resolves (overlay~Y and not instance-level N/A), and as \emph{passed}
when its raw score is $\geq 1$; mechanical is the frozen C1
(format/modeling) facet set and judgment is C2$+$C3
(Section~\ref{sec:results}).

\textbf{(a) Sector.} Across the eleven GICS sectors, mechanical pass
rates occupy a 2.0-point band (80.3--82.3\%) and judgment rates a
5.0-point band (52.6--57.6\%). The mechanical--judgment gap ranges from
23.6 points in Communication Services to 28.9 in Energy. Energy and
Utilities have the widest gaps and the lowest judgment rates (53.1\%
and 52.6\%), but both are among the smallest slices after Real Estate,
so we do not interpret the difference. Capability-failure rates range
from 9.4\% in Materials (9 of 96 grid cells) to 15.3\% in Real Estate
(11 of 72).

\begin{figure}[t]
\centering
\includegraphics[width=0.55\textwidth]{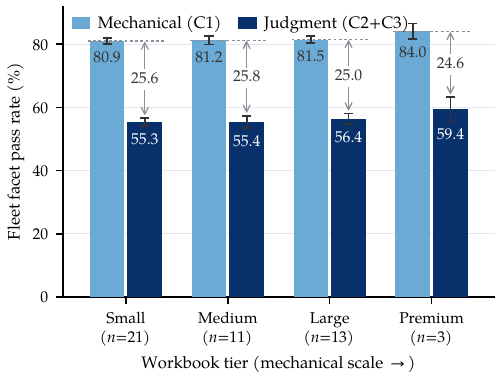}
\caption{\textbf{Facet pass rate by workbook tier.} Across 24 agents,
mechanical and judgment pass rates vary little across the four vendor
tiers; the gap is 24.6--25.8 points. Tier is a scale covariate, not a
measure of verified analyst quality.}
\label{fig:tier}
\end{figure}

\textbf{(b) Tier.} Pooled pass rates do not decline across the vendor
tiers (Figure~\ref{fig:tier}). Premium, the 21--30-hour analyst tier, has rates of
84.0\%/59.4\% over $n{=}64$ cells, compared with
80.9\%/55.3\% over 439 Small-tier cells. The gap remains
24.6--25.8 points in every tier, while capability failures are 12.9\%
for Small grid cells and 11.1\% for Premium cells. These descriptive
rates do not associate larger workbook tiers with lower pass rates.

\textbf{(c) Completion status.} The contract requires three artifacts:
the workbook, memo, and assumptions file. Of 1{,}011 scored cells, 825
ship all three to the contracted paths, 175 ship only the workbook, and
11 ship exactly one companion (7 memo-only and 4 assumptions-only);
the \texttt{memo\_wired}/\texttt{assumptions\_wired} ledger fields test
existence at those absolute paths. The remaining 141 grid cells produce
no valid workbook after a full-budget attempt. They count against
Compl.\ in Table~\ref{tab:leaderboard} and enter the 48-task denominator
in Figure~\ref{fig:scatter} as zeros, but have no facet rows and appear
in Table~\ref{app2:slicetable} only as counts.

Full-contract cells pass 82.4\% of mechanical and 57.2\% of judgment
facets, compared with 76.6\% and 49.9\% for workbook-only cells. This is
an association between contract completion and workbook score, not
evidence that a missing companion causes the lower score. The
mechanical--judgment gaps remain similar at 25.2 and 26.6 points.
Sector, tier, and completion recuts therefore do not localize the gap,
but these pooled descriptions cannot establish whether it belongs to
the fleet or to an unobserved subset of tasks or runs.

\begin{table*}[t]
\caption{\textbf{Fleet-level facet pass rates re-cut by task
covariates and contract completion.} Pooled over all 24 leaderboard
agents on the 48-task core (1{,}011 scored cells of the
$24\times48$ grid): pass rate (\%, raw score $\geq 1$ among active
facets) under the mechanical (C1) and judgment (C2$+$C3) lenses;
$\Delta$ = mechanical $-$ judgment in points. T = core tasks in the
slice; $n$ = scored (agent, task) cells; F = capability-failure
grid cells (no valid workbook: counted against Compl.\ in
Table~\ref{tab:leaderboard}, no facet rows here). Computed from the
frozen \texttt{w2\_main} merge over all 56 facets, including the
re-verified no-look-ahead detector (facet 2.5.2, gate G5).}
\label{app2:slicetable}
\scriptsize
\setlength{\tabcolsep}{2.6pt}
\centering
\begin{minipage}[t]{0.53\textwidth}
\centering
\emph{(a) GICS sector of the task}\\[2pt]
\begin{tabular}{@{}lrrrrrr@{}}
\toprule
Sector & T & $n$ & F & Mech. & Judg. & $\Delta$\\
\midrule
Industrials             & 6 & 125 & 19 & 80.3 & 55.5 & 24.8\\
Financials              & 5 & 109 & 11 & 81.0 & 56.8 & 24.2\\
Consumer Discretionary  & 5 & 107 & 13 & 82.3 & 57.3 & 25.0\\
Information Technology  & 5 & 104 & 16 & 80.7 & 55.2 & 25.5\\
Consumer Staples        & 5 & 103 & 17 & 81.5 & 57.6 & 23.8\\
Materials               & 4 &  87 &  9 & 82.2 & 56.3 & 25.9\\
Health Care             & 4 &  84 & 12 & 82.2 & 56.3 & 26.0\\
Communication Services  & 4 &  84 & 12 & 81.1 & 57.5 & 23.6\\
Energy                  & 4 &  82 & 14 & 82.0 & 53.1 & 28.9\\
Utilities               & 3 &  65 &  7 & 81.0 & 52.6 & 28.4\\
Real Estate             & 3 &  61 & 11 & 80.5 & 54.9 & 25.6\\
\midrule
All sectors             & 48 & 1{,}011 & 141 & 81.3 & 55.9 & 25.4\\
\bottomrule
\end{tabular}
\end{minipage}\hfill
\begin{minipage}[t]{0.43\textwidth}
\centering
\emph{(b) Difficulty tier}\\[2pt]
\begin{tabular}{@{}lrrrrrr@{}}
\toprule
Tier & T & $n$ & F & Mech. & Judg. & $\Delta$\\
\midrule
Small   & 21 & 439 & 65 & 80.9 & 55.3 & 25.6\\
Medium  & 11 & 234 & 30 & 81.2 & 55.4 & 25.8\\
Large   & 13 & 274 & 38 & 81.5 & 56.4 & 25.0\\
Premium &  3 &  64 &  8 & 84.0 & 59.4 & 24.6\\
\bottomrule
\end{tabular}\\[8pt]
\emph{(c) Three-artifact contract completion}\\[2pt]
\begin{tabular}{@{}lrrrr@{}}
\toprule
Status & $n$ & Mech. & Judg. & $\Delta$\\
\midrule
Workbook, memo, assumptions & 825 & 82.4 & 57.2 & 25.2\\
Workbook, one companion     &  11 & 73.8 & 53.6 & 20.2\\
Workbook only                     & 175 & 76.6 & 49.9 & 26.6\\
Failed / absent workbook          & 141 & ---  & ---  & ---\\
\bottomrule
\end{tabular}
\end{minipage}
\end{table*}

\section{Additional Main-Result Diagnostics}
\label{app2:additional}
\label{app2:loss}

\subsection{Company-Grouped Envelope Cross-Fit}
\textbf{Design.} We repeat five-fold splits 20 times, assigning all
workbooks for a company to one fold. Disagreement tails use only the
other multi-covered companies; E-industry pools remove every held-out
company from the full corpus. We report p75/p80/p85/p90 sensitivity and
5{,}000 company-cluster bootstrap intervals. The analyzed peer subset
contains 65 companies and 137 workbooks; only 17 undirected
implied-price pairs survive the positive-value and 5$\times$ unit guard.

\textbf{Results.} Across 39 eligible directed price observations,
E-method strict coverage is 53.8\% (95\% CI 38.5--68.6\%). The
p90 near band covers 91.2\% (82.6--97.2\%); p75/p80/p85 near coverage
is 72.8\%/78.3\%/85.0\%. At p90, strict held-out E-industry value
coverage is 82.4\% WACC, 90.2\% terminal growth, 75.4\% beta, 79.0\%
tax, 80.6\% ERP, and 84.5\% risk-free rate. This prevents direct
company leakage but remains an internal reuse of one source corpus;
the perturbation control below bounds how much of the near coverage
reflects permissive bands.

\subsection{Perturbation Control: Coverage vs.\ Selectivity}
\label{app2:selectivity}

A wide-enough band admits everything; coverage alone cannot separate
selectivity from permissiveness. We therefore re-test every held-out
peer value after displacing it by $\pm k\,q_f$, $k\in\{1,2,3\}$, under
the unchanged frozen coverage rule, so that real values and
counterfeits face the identical test. For assumptions, $q_f$ is the
facet's leave-one-company-out median absolute same-company peer
difference (WACC 147\,bp, terminal growth 100\,bp, beta 0.32, tax
385\,bp, ERP 114\,bp, risk-free 36\,bp)---one quantum of typical
professional disagreement; for price, the p90 directed cross-analyst
relative dispersion ($\delta{=}0.995$), applied on the ratio scale so
counterfeits stay positive. E-industry bands are per-GICS-group and
leave-one-company-out throughout. Because each counterfeit must be
paired with its real value, this experiment uses the deterministic
leave-one-company-out split rather than the 20 repeated five-fold
averages above; its coverage column therefore differs from those
averages by at most 3.7 points (e.g., near price 94.9\% vs.\ 91.2\%),
and the comparison of interest is within-row, coverage against
rejection under the identical rule.

Table~\ref{tab:selectivity} reports the result. The E-industry layer
is selective: it admits 80.8\% of real held-out values while rejecting
66.1\% of $\pm2q_f$ and 85.4\% of $\pm3q_f$ counterfeits, monotone in
$k$ for every facet. Sweeping the band percentile traces the
coverage--selectivity frontier (p75: 56.3\%/89.6\% through p90:
80.8\%/66.1\% at $\pm2q_f$); the released p90 setting is an interior
operating point on that frontier, not its permissive end. The
E-method near-price band is the honest exception: its multiplicative
near rule with $\delta{\approx}1.0$ drives the lower edge toward
zero, so it rejects upward $\pm2$-dispersion counterfeits at 38.5\%
but downward ones at 0\%. Near-price membership is therefore read
only as the partial-credit zone of Eq.~\eqref{eq:envstate}; the
strict price band, which alone earns full credit, rejects 65.4\% of
the same counterfeits against 53.8\% real-peer coverage. Full
per-facet counts, non-positive-counterfeit accounting, and the pooled
variant are in \texttt{reference/envelope\_selectivity\_gics.json}.

\begin{table}[t]
\caption{\textbf{Perturbation control: real-peer coverage vs.\ counterfeit
rejection.} Each held-out peer value is displaced by $\pm k\,q_f$ and
re-tested under the unchanged coverage rule. For assumptions,
$q_f$ is the leave-one-company-out median absolute same-company peer
difference; for price, the p90 directed cross-analyst relative dispersion
($\delta{=}0.995$), applied on the ratio scale. E-industry bands are
per-GICS-group, leave-one-company-out; $n$ counts real held-out values
($2n$ counterfeits per $k$).}
\label{tab:selectivity}
\centering\small
\setlength{\tabcolsep}{3.2pt}
\begin{tabular}{lrrrrr}
\toprule
& & \multicolumn{3}{c}{Counterfeit rejection (\%)} & \\
\cmidrule(lr){3-5}
Layer / facet & Cov.\,\% & $\pm1q_f$ & $\pm2q_f$ & $\pm3q_f$ & $n$ \\
\midrule
E-method price, near band  & 94.9 & 9.0  & 19.2 & 23.1 & 39 \\
\quad upward counterfeits only & --- & 17.9 & 38.5 & 46.2 & 39 \\
E-method price, strict band & 53.8 & 64.1 & 65.4 & 65.4 & 39 \\
\midrule
E-industry (6 facets, pooled) & 80.8 & 36.8 & 66.1 & 85.4 & 453 \\
\quad WACC           & 82.2 & 29.0 & 57.9 & 80.4 & 107 \\
\quad terminal $g$   & 90.0 & 36.7 & 60.0 & 85.0 & 30 \\
\quad beta           & 76.2 & 45.8 & 79.2 & 93.5 & 84 \\
\quad tax rate       & 79.6 & 33.5 & 63.6 & 85.8 & 88 \\
\quad ERP            & 79.7 & 41.9 & 69.6 & 86.5 & 74 \\
\quad risk-free rate & 82.9 & 36.4 & 65.0 & 82.1 & 70 \\
\midrule
\multicolumn{6}{@{}l}{\emph{E-industry band-percentile frontier}
  (coverage\,\% / rejection\,\% at $\pm2q_f$):} \\
\multicolumn{6}{@{}l}{\quad p75: 56.3\,/\,89.6 \quad
  p80: 65.6\,/\,85.0 \quad p85: 72.6\,/\,77.6 \quad
  p90: 80.8\,/\,66.1} \\
\bottomrule
\end{tabular}
\end{table}

\subsection{Failure-Aware Panel Uncertainty}
The primary score is
$\phi_0(m)=48^{-1}\sum_{t=1}^{48}\widetilde{\phi}_{m,t}$, where failed
or unscoreable cells contribute zero. The frozen matrix has 1{,}011
scoreable cells and 141 failures. In 50{,}000 paired task-cluster
bootstraps, the point leader remains first in 99.998\% of replicates,
the exact top-three set in 99.848\%, and the exact top-five set in only
22.446\%; mean Spearman correlation with the point ranking is 0.972.
The largest completed-only to failure-aware movement is Doubao 2.1-pro,
rank 8 to 22. These intervals describe the fixed operational core, not
the full 196-task bank, generation randomness, or judge-family
uncertainty.

\subsection{Grader-Family Dependence}
The taxonomy contains 29 deterministic (D), 4 direct-rule (R), and 23
judged (J) facets. We reconstruct the pre-gate failure-aware ranking and
remove one family at a time over 5{,}000 paired task bootstraps. Mean
Kendall $\tau_b$ against the full ranking is 0.888
$[0.833,0.935]$ without D, 0.965 $[0.928,0.993]$ without R, and
0.683 $[0.601,0.761]$ without J; top-1 preservation is
1.000/1.000/0.305. These are dependence ablations, not correctness
tests. R has only four facets and a median of one scored facet per
completed cell; operational D removal additionally disables all eight
validity gates.

\subsection{Gates vs.\ Additive Credit}
\label{app2:gateslinear}

\begin{figure}[t]
\centering
\includegraphics[width=0.55\textwidth]{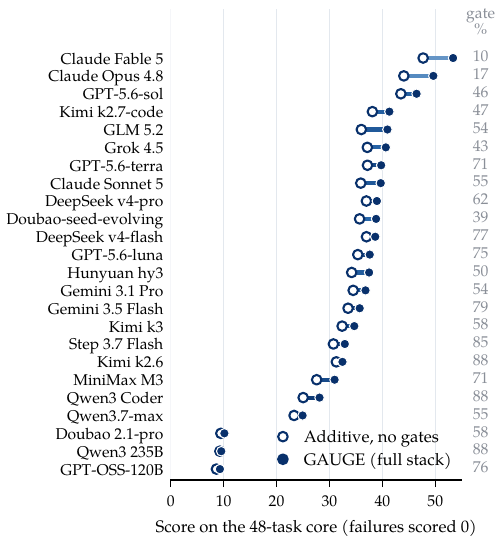}
\caption{\textbf{Full stack vs.\ additive credit on identical facet
outcomes.} Removing gates flips 11/276 model-pair orderings; the
additive-minus-GAUGE difference has $r=0.73$ with gate-trigger rate.}
\label{fig:gateslinear}
\end{figure}

On identical facet outcomes, replacing the gate caps of
Eq.~\eqref{eq:gates} with equal-weight additive credit reverses 11 of
276 model-pair orderings; in 10 of the 11 reversals the additive rule
prefers the model with the \emph{higher} gate-trigger rate, and the
additive-minus-GAUGE score difference correlates $r=0.73$ with gate
rate (Figure~\ref{fig:gateslinear}). Removing the caps therefore
systematically rewards structurally broken workbooks. Per-pair records
are released in \texttt{reference/gates\_vs\_linear.json}.

\subsection{Cross-Family Judge Replication}
\label{app2:xfam}

The frozen judge (Claude Opus 4.8) shares a provider with the top two
leaderboard agents, so we re-judge a stratified subset with a judge
from a different provider under the identical protocol: GPT-5.6-sol,
$k=5$ votes with majority reduction, the same prompts, evidence
packs, and overlays. The subset is 12 tasks (one per GICS sector
group, seeded draw from the 46 tasks where all eight target agents
have a frozen judged cell) $\times$ 8 agents spanning five providers
(Claude Fable 5, Opus 4.8, Sonnet 5; GPT-5.6-sol, -terra; Gemini 3.1
Pro; DeepSeek v4-pro; Kimi k2.7-code) --- 96 cells, all judged
successfully, 2{,}208 facet pairs.

Facet-level agreement with the frozen judge is 73.5\% exact and
92.2\% within one rung of the 0/1/2 ladder (quadratic-weighted
$\kappa=0.675$; 7.1\% of pairs disagree on N/A status). The
cross-family judge is uniformly stricter, and
Table~\ref{tab:xfamjudge} shows the shift is not concentrated on any
provider: per-agent judged-facet means move $-1.9$ to $-6.1\,\phi$,
with the Anthropic-generated cells moving $-5.2$ against $-4.8$ for
non-Anthropic cells (permutation $p=0.71$ on the gap over 20{,}000
label shuffles). Subset agent ordering is preserved at Kendall
$\tau=0.857$: the only movements are among the three near-tied
bottom ranks (Sonnet 5, Gemini 3.1 Pro, Kimi k2.7-code, spanning
2.3 points under the frozen judge). The top of the table is
unchanged --- the OpenAI judge also places Claude Fable 5 first,
above its own provider's GPT-5.6-sol. The five lowest-agreement
facets (3.3.3, 3.2.3, 1.3.2, 5.1.2, 5.2.3) are flagged as judge
calibration targets. Artifacts: \texttt{runs/campaign/xfam\_judge/}
(subset ledger, per-cell votes, analysis).

\begin{table}[t]
\caption{\textbf{Cross-family judge replication on the 96-cell
subset.} Judged-facet cell means on the $\phi$ scale under the frozen
judge (Claude Opus 4.8) and the cross-family judge (GPT-5.6-sol),
both $k=5$, identical protocol; $\Delta$ = frozen $-$ cross-family;
$n=12$ tasks per agent. Rk = rank within the subset under each
judge. The cross-family judge is uniformly stricter; ordering moves
only among the near-tied bottom three (Kendall $\tau=0.857$).}
\label{tab:xfamjudge}
\centering\small
\setlength{\tabcolsep}{3.4pt}
\begin{tabular}{lrrrcc}
\toprule
& \multicolumn{2}{c}{Judged-facet mean} & & \multicolumn{2}{c}{Rk}\\
\cmidrule(lr){2-3}\cmidrule(lr){5-6}
Agent & Opus 4.8 & 5.6-sol & $\Delta$ & O & S\\
\midrule
Claude Fable 5   & 47.8 & 41.7 & $+6.1$ & 1 & 1\\
GPT-5.6-sol      & 42.5 & 36.9 & $+5.5$ & 2 & 2\\
Claude Opus 4.8  & 37.5 & 31.8 & $+5.7$ & 3 & 3\\
GPT-5.6-terra    & 35.5 & 29.9 & $+5.6$ & 4 & 4\\
DeepSeek v4-pro  & 32.1 & 26.8 & $+5.4$ & 5 & 5\\
Kimi k2.7-code   & 31.4 & 25.8 & $+5.6$ & 6 & 8\\
Claude Sonnet 5  & 30.1 & 26.3 & $+3.8$ & 7 & 6\\
Gemini 3.1 Pro   & 27.9 & 26.0 & $+1.9$ & 8 & 7\\
\midrule
\multicolumn{6}{@{}l}{Anthropic cells $\Delta=+5.2$; non-Anthropic
$\Delta=+4.8$; gap 0.4 ($p=0.71$)}\\
\multicolumn{6}{@{}l}{Facet agreement: 73.5\% exact, 92.2\% within
one rung, $\kappa_w=0.675$}\\
\bottomrule
\end{tabular}
\end{table}

\section{Training-Signal Experiments}
\label{app2:training}

\begin{figure}[t]
\centering
\includegraphics[width=0.55\textwidth]{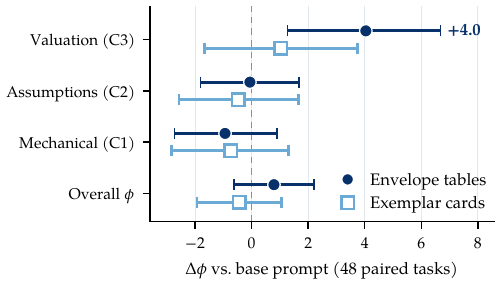}
\caption{\textbf{Corpus-as-context.} Paired score deltas with 95\%
intervals. E-industry assumption tables move valuation judgment by
$+4.0\,\phi$; the mechanical interval includes zero; worked-example
cards produce a non-significant $+1.0$ change.}
\label{fig:context}
\end{figure}

The training split contains 200 workbooks (80/70/40/10 across vendor
tiers), excludes evaluation and multi-coverage companies, and has
167/200 automatically certified harness-ready. The main text reports
the paired effect estimates and Figure~\ref{fig:context} shows the
complete context result.

For trajectory fine-tuning, two Claude Opus 4.8 trajectories were
generated for each of 150 training companies and rejection-sampled by
deterministic score, yielding 146 retained trajectories. On 15 paired
core tasks, the current checkpoint changes valuation by $+8.4\,\phi$
(95\% CI $[+1.8,+14.9]$), mechanics by $+0.6$ (CI spans zero),
assumptions by $-4.9$ $[-8.9,-0.6]$, and overall score by $-0.8$
$[-3.6,+2.0]$. Earlier tool-template and stopping failures are retained
in the release to make this exploratory result auditable.

\section{Cost, Latency, and Compute}
\label{app2:cost}

Every number in this section is mined from two frozen run ledgers: the
generation ledger (\texttt{ledger.jsonl}, 1{,}241 rows deduplicated to
the last attempt per agent$\times$task, i.e.\ the 1{,}152-cell matrix)
and the judge ledger (\texttt{judge\_k5.jsonl}, 1{,}204 rows covering
the 1{,}011 judged cells). Wall seconds, provider-reported token usage,
vote vectors, and cache counters are logged at the transport layer and
are independent of how the cells score.

\paragraph{Generation layer.} Table~\ref{app2:cost:gen} reports the
per-agent profile. Completing the $24\times48$ matrix consumed
24{,}428 agent API rounds, 366.0M input plus 55.5M output tokens as
reported, and 296 agent-hours of summed wall time over a 6.5-day
campaign (2026-07-16 19:27 to 07-23 07:50 UTC). The traffic includes
the retry budget of Appendix~\ref{app:harness}: 89 superseded ledger
rows --- all failed attempts (\texttt{gen\_error} ``no workbook
produced'') --- were re-driven under the infrastructure-failure
policy, 247 of the 1{,}152 cells consumed the harness's single second
attempt, and every one of the 141 recorded failures carries
\texttt{attempts}${=}2$, i.e.\ exhausted the full budget before being
recorded; none was ever re-driven to success (Section~\ref{sec:results}).
Median wall time per completed task spans $23\times$, from 89\,s for
GPT-5.6-luna to 2{,}086\,s for Doubao 2.1-pro. Three of the four
low-completion agents use $4$--$7\times$ the fleet median of 0.20M
tokens per valid workbook: Qwen3 235B uses 1.5M, Doubao 2.1-pro 1.3M,
and Qwen3.7-max 0.8M; GPT-OSS-120B is the exception at 0.21M. The 141
failed cells consume 32.9M tokens, or 7.8\% of the fleet total, and are
included in the Tot.\ column.

Two accounting disclosures. First, we sum each adapter's reported
per-round input-token field (provider names differ). Its semantics
therefore follow the provider: endpoints that serve the
harness through prompt caching report far less input per round than
endpoints that re-count the full context. The contrast is stark ---
GLM~5.2 reports 2.0k input tokens per round across 45.0 rounds while
Claude Sonnet~5 reports 48.4k per round across 49.6 --- so the In
column is honest per agent but not comparable across wire adapters;
output tokens, API rounds, and wall time are. We report the as-logged
numbers rather than impute a common accounting. Second, five failed
cells (four Qwen3.7-max, one Kimi k3; \texttt{gen\_error} ``no workbook
produced'') carry zero rounds and zero tokens because the harness found
no usage-bearing transcript records: their failure is counted, their
cost is invisible.

\begin{table*}[t]
\caption{\textbf{Per-agent generation cost and latency on the 48-task
core} (frozen \texttt{w2\_main} ledger; rows grouped for compact cost
comparison rather than leaderboard order). Med.\,s:
median wall seconds per completed task. Rounds/In/Out: mean API rounds
and provider-reported input/output tokens (thousands) per completed
task; In is as-reported and not comparable across wire adapters (see
text). Tot.: total tokens over all attempted cells, failed generations
included.}
\label{app2:cost:gen}
\footnotesize
\begin{tabular}{llrrrrrr}
\toprule
Agent & Wire & Compl. & Med.\,s & Rounds & In (k) & Out (k) & Tot.\ (M)\\
\midrule
Claude Fable 5       & anthropic & 48/48 &   468 &  9.4 &   316.6 &  39.2 &  17.08\\
Claude Opus 4.8      & anthropic & 48/48 &   422 & 20.2 &   519.7 &  37.3 &  26.74\\
GPT-5.6-sol          & openai    & 48/48 &   188 &  3.9 &    41.8 &  15.2 &   2.74\\
Grok 4.5             & openai    & 42/48 & 1{,}024 &  7.0 &   242.1 &  51.6 &  12.39\\
Doubao-seed-evolving & openai    & 44/48 & 1{,}324 & 26.1 & 1{,}058.5 &  72.4 &  50.35\\
Kimi k2.7-code       & anthropic & 47/48 & 1{,}200 & 37.1 &   102.8 &  98.4 &   9.51\\
GLM 5.2              & anthropic & 48/48 & 1{,}261 & 45.0 &    90.8 & 106.8 &   9.48\\
Claude Sonnet 5      & anthropic & 47/48 &   656 & 49.6 & 2{,}404.6 &  61.7 & 118.90\\
Doubao 2.1-pro       & openai    & 12/48 & 2{,}086 & 23.2 & 1{,}024.7 &  62.0 &  15.45\\
GPT-5.6-terra        & openai    & 48/48 &   122 &  6.3 &    71.6 &  13.7 &   4.09\\
MiniMax M3           & anthropic & 38/48 &   731 & 48.0 &   109.2 &  95.0 &   8.02\\
DeepSeek v4-pro      & anthropic & 48/48 & 1{,}078 & 36.4 &    97.9 &  67.6 &   7.94\\
DeepSeek v4-flash    & anthropic & 48/48 &   502 & 26.4 &   129.6 &  61.7 &   9.18\\
Qwen3.7-max          & openai    & 31/48 & 1{,}858 & 14.3 &   701.8 &  70.7 &  25.53\\
Kimi k3              & kimi\_official & 43/48 &   742 & 37.0 &    53.5 &  40.4 &   4.12\\
GPT-5.6-luna         & openai    & 48/48 &    89 &  7.2 &    77.2 &  10.9 &   4.23\\
Hunyuan hy3          & anthropic & 48/48 &   574 & 37.6 &    77.1 &  53.5 &   6.27\\
Gemini 3.1 Pro       & gemini    & 48/48 &   200 &  8.2 &   184.3 &  16.0 &   9.62\\
Gemini 3.5 Flash     & gemini    & 48/48 &   112 & 10.4 &   206.3 &  23.6 &  11.03\\
Step 3.7 Flash       & anthropic & 46/48 &   523 & 22.3 &    77.3 &  80.1 &   7.40\\
Kimi k2.6            & anthropic & 48/48 &   444 & 35.0 &   126.1 &  54.4 &   8.66\\
Qwen3 235B           & openai    & 16/48 &   408 & 11.8 &   150.3 &  17.2 &  23.66\\
Qwen3 Coder          & openai    & 48/48 &   235 &  8.8 &   191.8 &  21.7 &  10.25\\
GPT-OSS-120B         & openai    & 21/48 &   259 &  6.9 &    98.8 &  24.7 &   4.32\\
\midrule
Fleet & --- & 1{,}011/1{,}152 & 513 & 23.0 & 321.9 & 50.1 & 421.5\\
\bottomrule
\end{tabular}
\end{table*}

\paragraph{Judge layer.} Each judged cell costs 23 facets $\times$
$k{=}5$ votes $=$ 115 scoring draws, the first of which doubles as the
cache-priming call before the remaining draws fan out concurrently
(Appendix~\ref{app:judge}); Table~\ref{app2:cost:judge} aggregates the
fleet. Of the 116{,}265 nominal draws, 110{,}038 (94.6\%) returned a
parseable verdict after the retry discipline of
Appendix~\ref{app:judge}; the 257 facet-cells (1.1\%) that lost all
five draws record an explicit error and score N/A-dropped, never zero.
The shared evidence block incurs 99.5M cache-write tokens and 2{,}059M
cache-read tokens fleet-wide, a 20.7:1 read-to-write ratio; 95.4\% of
cache-accounted judge input is served as reads. The paired parity check
in Appendix~\ref{app:judge} finds no score change between cached and
uncached serving. Median cell latency is 111.9\,s at concurrency 6
(830 cells); 175 cells run at concurrency 4 and 2 at concurrency 3
during rate-limit windows, with cache TTL \texttt{5m}. Summed judge
compute is 33.5\,h, about 11\% of the fleet's 296 agent-hours.

\noindent\begin{minipage}{\columnwidth}\centering
\captionof{table}{\textbf{Judge-layer cost structure} (frozen
\texttt{judge\_k5.jsonl}, deduplicated; judge
\texttt{claude-opus-4-8}, $k{=}5$).}
\label{app2:cost:judge}
\footnotesize
\begin{tabular}{lr}
\toprule
Judged cells (23 facets each, all \texttt{judge\_ok}) & 1{,}011\\
Scoring draws ($1011\times23\times5$; first primes cache) & 116{,}265\\
Parseable verdicts returned & 110{,}038 (94.6\%)\\
Facet-cells losing all five draws & 257 (1.1\%)\\
Mean within-cell vote agreement & 0.978\\
Cache-write / cache-read tokens & 99.5M / 2{,}059M\\
Read-to-write ratio (share served as reads) & 20.7:1 (95.4\%)\\
Median / mean wall s per cell & 111.9 / 119.1\\
Summed judge compute & 33.5 h\\
\bottomrule
\end{tabular}
\end{minipage}

\paragraph{No dollar totals.} We deliberately do not convert to
dollars: the 24 agents were served through heterogeneous provider
endpoints and an internal proxy whose contract pricing is not public,
so any dollar figure would be an estimate layered on non-comparable
accounting. The token, call, and wall-time counts above are the
reproducible quantities, and both ledgers are released with the
harness.

\section{Worked Example: One Task End to End}
\label{app:example}

We walk one real evaluation cell --- EMR (Capital Goods) under the
best-scoring agent --- through the full pipeline; all numbers below
are from the released artifacts of that run.

\textbf{Input.} The pack shows revenue of 15{,}165\,/\,17{,}492\,/\,%
18{,}016 (USD mm) for FY2023--25, last close \$135.20,
554.6\,mm shares, and a verified balance-sheet identity
($41{,}964 = 21{,}666 + 20{,}298$).

\textbf{Output.} The agent produced the eleven-sheet workbook, memo,
and assumptions file in one pass. The deterministic checker records:
worst balance-sheet imbalance $2\times10^{-15}$\,\% across five
forecast periods (S-04), cash-flow-to-balance-sheet cash gap 0.0\,\%
(S-05), projection-region formula density 97.9\,\% (S-09), and full
blue-input color compliance (B-05); no validity gate fires. The
assumptions file resolves its numbers to cells --- e.g.\
\texttt{\{"name": "wacc", "value": 0.0885, "source":
"WACC!B19"\}}, whose source note documents its own composition
(85.4\% equity at $K_e$~9.75\%, 14.6\% debt at after-tax
$K_d$~3.6\%), alongside terminal growth 0.025 --- and the memo's
headline (DCF implied \$99.13, 26.7\,\% below
the close) matches the workbook.

\textbf{Where judgment thins out.} The workbook's sensitivity tab is a
WACC~$\times$~$g$ grid spanning $0.0785$--$0.0985$ by
$0.015$--$0.035$. Every cell is formula-driven, but the axes reproduce
the parenthetical example in the task prompt (``e.g., WACC $\times$
terminal growth''). Facet 4.4.1 asks, under the Capital Goods overlay,
for end-market volume, price realization, or incremental margin; the
judge scores the facet 0 with 5/5 agreement. The workbook also reports
that 74.2\,\% of enterprise value lies in the terminal period, but none
of the sensitivity axes tests a real operating driver. The cell scores
$\phi=52.9$ over 45 active facets, the 92nd percentile among 1{,}011
scored cells (fleet median 39.5): high mechanical execution coexists
with a zero on real-driver sensitivity.

The same cell, seen through two graded facets --- the judge's five
independent votes are unanimous in both directions:

\begin{quotebox}{Graded facets, cell EMR $\times$ best agent (from the
released vote records)}
\textbf{Facet 5.1.1} --- thesis with falsifiable claims \graderJ{}
\hfill votes \texttt{[2\,2\,2\,2\,2]} $\Rightarrow$ \textbf{2}\\
The memo states quantified, workbook-tied claims (implied \$99.13,
26.7\% below the close; margin path named per driver) --- clears the
``3--5 falsifiable claims tied to drivers'' bar.\\[3pt]
\textbf{Facet 4.4.1} --- sensitivity on real value drivers \graderJ{}
\hfill votes \texttt{[0\,0\,0\,0\,0]} $\Rightarrow$ \textbf{0}\\
Rubric rung 0 is literally ``only WACC $\times$ $g$ (cookbook)'';
the Capital Goods overlay emphasis names the drivers that move an
industrial --- end-market volume, price realization, incremental
margin --- and the workbook's single grid spans none of them.
\end{quotebox}

The four reference-envelope [R] facets are graded by the
\texttt{R\_envelope} checker
against per-ticker reference packs assembled from corpus artifacts
only: the WACC band is the per-industry p10--p90 distribution
intersected with the reference workbook's own WACC widened by the
374\,bp cross-analyst p90
of Section~\ref{sec:ava}; the multiple bands are the reference
workbook's implied
P/E and EV/EBITDA with $\pm20\%$ flex; and the hidden margin base is
the reference workbook's own forecast EBIT margin. Coverage over the
1{,}011 scored cells: WACC-in-band
801, implied multiples 364, margin-trajectory 270, segment-margin
differentiation 9; where a pack lacks a band the grader abstains to
N/A rather than fabricate one. Pack construction is released as
\texttt{tools/build\_core\_reference\_packs.py}.

\section{Worked Example II: A Gated Cell}
\label{app2:example2}

Appendix~\ref{app:example} reports a completed cell with no triggered
gate. Here we examine a completed VZ cell (Telecommunication Services)
from DeepSeek v4-flash, which triggers G1 on 56\% of its tasks
(Appendix~\ref{app:gates}). The cell passes most deterministic checks
but has an unbalanced balance sheet; all values below come from the
frozen ledger and stored scorecard.

\textbf{Input.} The pack (as-of 2026-04-30, FY2025 vintage) describes a
mature carrier: revenue 133{,}974\,/\,134{,}788\,/\,135{,}607 (mm) for
2023A--25A, cash 4{,}194 against total debt 167{,}005, and five
forecast years 2026E--2030E.

\textbf{Output, and what passed.} One attempt, 27 API calls,
159{,}846 input / 58{,}464 output tokens, 461.9\,s wall clock; the
full eleven-sheet workbook plus memo and a 29-entry assumptions file
--- superficially the same deliverable as Appendix~\ref{app:example}.
Most of the deterministic battery passes, much of it to machine
precision: zero Excel errors (S-03); cash-flow-to-balance-sheet cash
gap 0.0\% in all five periods (S-05); net income identical on the IS
bottom line and the CF top line (S-08); projection-region formula
density 100\% --- 314 of 314 projected cells are formulas (S-09); a
constant 24.0\% tax rate, in band every year (B-03); blue-input color
compliance 34/34 (B-05); D\&A consistency, the PP\&E roll, and the
retained-earnings roll all tie; and the EV$\to$equity$\to$per-share
bridge is complete and arithmetically consistent (G8 record), ending
in a clearly labelled implied share price of \$108.57
(\texttt{DCF!B29}, S-06). The only other deterministic check that
fails outright is S-10: 20 in-formula literals (e.g.\
\texttt{=Revenue\_Build!E2*0.08}), which holds the hardcode facet
2.5.1 at 1 rather than 2 but fires no gate.

\textbf{The gate.} S-04 asks the balance sheet to balance within
0.1\% in every period. This one is off by more than two orders of
magnitude beyond that tolerance, in every forecast column:

\begin{quotebox}{The gate evidence --- deterministic record S-04,
stored with the cell}
\textbf{S-04} --- BS balances every period within 0.1\% \graderD{}
\hfill \texttt{pass: false}\\
\texttt{worst\_imbalance\_pct = 0.2237}: total assets differ from
liabilities plus equity by 21.2\% (2026E), 21.5\%, 21.9\%, 22.1\%, and
22.4\% (2030E) of total assets. Detection facet 2.2.1 scores 0; gate
G1 (``Balance sheet does not balance'') fires with ceiling
$\kappa_{\mathrm{G1}}=40$.
\end{quotebox}

The agent knew. Its own \texttt{Checks} tab computes assets minus
liabilities-and-equity per year and prints \texttt{FAIL} five times
--- a hole that opens at 83{,}455 (mm) and compounds to 96{,}284 by
2030E --- and it shipped the workbook anyway:

\begin{promptbox}{Checks sheet of the shipped VZ workbook
(recalculated values; elisions ours)}
Check                           Value              Status
BS Balances 2026E               -83455.03138771    FAIL
BS Balances 2027E               -86762.4408437935  FAIL
BS Balances 2028E               -90004.5547394265  FAIL
BS Balances 2029E               -93175.7938387731  FAIL
BS Balances 2030E               -96284.2695395109  FAIL
CF Ending Cash = BS Cash 2026E  0                  PASS
  ... (2027E-2030E identical: 0, PASS)
NI flows to RE 2026E            0                  PASS
  ... (2027E-2030E identical: 0, PASS)
\end{promptbox}

The memo and workbook also disagree on valuation. The memo's fourth
falsifiable claim gives a DCF range of $\sim$\$35--42, while the
labelled workbook cell gives \$108.57, or $2.7\times$ the \$39.92
current price quoted in the memo.

\textbf{The additive counterfactual.} Of 25 scored facets, 18 score 1,
six score 2, and one scores 0: facet 2.2.1, the balance-sheet identity.
This yields 96.0\% passing facets. Under the linear
$0/1/2\mapsto0/50/100$ map used in
Appendix~\ref{app2:additional}, with the cap in
Eq.~\eqref{eq:gates} removed, the same stored outcomes average 60.0.
GAUGE's pre-gate aggregate is 69.6, and G1 sets
$\Phi=\min(69.6,\,40)=40$. The roughly twenty-point difference is
entirely due to the G1 ceiling on a workbook whose checks tab reports a
\$96bn balance-sheet gap.

\section{Three Failure Trajectories, Verbatim}
\label{app2:failures}

Section~\ref{sec:anatomy} splits the 141 never-completed cells into
``three different engineering failures --- planning collapse, silent
truncation, and structural incompleteness --- that a single `accuracy'
number would conflate.'' Here we walk three real trajectories from the
frozen \texttt{w2\_main} campaign: the first two instantiate the first
two never-completed mechanisms; the third is the boundary case --- a
workbook that \emph{ships} and is gate-capped, whose extreme form is
the third mechanism. Everything quoted is copy-pasted from the released
transcripts and build scripts; excerpts are condensed, and bracketed
turn annotations are ours.

\textbf{Trajectory 1: planning collapse --- the full-budget rebuild
loop (Doubao 2.1-pro $\times$ MA).} All 36 of Doubao 2.1-pro's failed
cells are full-budget non-convergences (every one consumed at least
2{,}652\,s across its two attempts), and the MA cell shows the
mechanism in miniature. The agent builds its workbook script by
appending heredoc chunks to one growing Python file; at turn~14 the
accreted script crashes at line~910. It correctly diagnoses the
architecture problem --- then the recovery itself collapses: two
consecutive turns emit 16{,}427 and 16{,}460 completion tokens, flush
against the 16{,}384-token output cap, so both \texttt{write\_file}
calls arrive with their arguments truncated away (one turn's entire
surviving prose is the fragment ``rate''). The agent then restarts the
whole script from scratch, is cut off again, and the attempt dies on an
API read timeout; the retry is killed at the 1{,}800\,s wall clock.
Total: 4{,}072\,s, 649{,}474 input and 103{,}482 output tokens across
19 API calls, no workbook.

\begin{promptbox}{Doubao 2.1-pro on MA: turns 14--18 of the released
transcript (condensed)}
[T14 run_bash result]
appended Sens/Checks/JSON/Memo + SAVE
Assumptions sheet done.
Saved interim.
Traceback (most recent call last):
  File ".../MA_FY2024/20260719T144440-d801ff/build_model.py",
  line 910, in <module>  [...]
[T15 | 16427 completion tokens] "The script appends in pieces
but the variables defined in later appends aren't available
when earlier chunks run in Python (since the whole file runs
once). Let me consolidate everything properly and "
   write_file {}   ->   <write_file: missing path>
[T16 | 16460 completion tokens] "rate"
   write_file {}   ->   <write_file: missing path>
[T17] "Let me write the entire model script at once, properly
organized:"  ->  wrote 37479 bytes to build_full.py
[T18] "I was cut off. Let me complete the script cleanly by
appending the rest:"
[attempt 1 ends] <api error: The read operation timed out>
[attempt 2 ends] <timeout: killed process group after 1800s>
\end{promptbox}

\noindent\emph{Deterministic consequence:} the generation harness
records \texttt{gen\_\allowbreak error:\allowbreak timeout} after
both attempts; the cell
enters the leaderboard as a capability failure
(Section~\ref{sec:anatomy}), not as a scored workbook.

\textbf{Trajectory 2: silent truncation --- the workbook that never
existed (GPT-OSS-120B $\times$ EMR).} Where Doubao burns its budget
fighting the same cap, GPT-OSS-120B dies in it instantly. On EMR the
first turn emits 8{,}708 completion tokens of one giant command whose
serialized tool call arrives with \emph{no arguments}; the harness's
recovery nudge --- issued uniformly to every model, like the wire
shims of Appendix~\ref{app:harness} --- tells it exactly how to
recover, and the very next request is rejected by the provider with
HTTP~400 (``unexpected character''). Both attempts die the same way,
77\,s total. This is not an EMR quirk: all 19 of GPT-OSS-120B's
absent-workbook cells show the HTTP~400 signature, 18 of them the
empty-arguments nudge first. The same wire-level truncation that
Doubao survives (and loops on) is, for this model, unrecoverable.

\begin{promptbox}{GPT-OSS-120B on EMR: the entire productive
transcript (stored attempt)}
[T0 | 8708 completion tokens; tool call arrives empty; the
harness recovery nudge replies:]
<run_bash: no command received. Your previous message was
likely cut off by the 16384-token output limit before the
command finished. Do NOT resend one giant command. Instead:
write long scripts to a file in small pieces with write_file
(append mode), or split the work across several short run_bash
[T1] <api error: HTTP 400: {"error":{"message":"unexpected
character: line 1 column 15435 (char 15434)
[trace_id=0adf5fc57ebc3a3bd142fe6dbd2271da]",
"type":"invalid_request_error","param":null,"code":null}}>
[attempt 1 had already ended in the same HTTP 400, at a
different offset (line 2 column 3556); output/ remains empty]
\end{promptbox}

\noindent\emph{Deterministic consequence:}
\texttt{gen\_error: no workbook produced}. The cell is counted as a
capability failure because the tested serving stack does not produce a
serializable workbook within the output limit.

\textbf{Trajectory 3: the confident hardcode --- a completed workbook
that does not flex (Step-3.7-flash $\times$ MO).} The third failure
ships. Step-3.7-flash finishes MO in 320.6\,s and 7 API calls: eleven
sheets, memo, assumptions file, checks tab --- and a balance sheet
whose ``inputs'' are invented. The build script stamps round constants
across \emph{all} columns, historical and forecast alike
(\texttt{range(2, 10)} spans 2022A--2029E): Goodwill \& Intangibles
15{,}000, Other Non-Current Assets 5{,}000, Other Current Assets 500
--- none of these values appears anywhere in the input pack --- while
the pack's genuine FY2024 cash figure (3{,}127) is copied backwards
into 2022A and 2023A as well. Each constant is styled
\texttt{blue\_font}, the analyst convention for a legitimate hardcoded
input (facet B-05): cosmetically compliant, substantively fabricated.

\begin{promptbox}{Step-3.7-flash on MO: \texttt{output/build\_model.py}
(Balance\_Sheet section, verbatim lines)}
ws_bs["A5"] = "Cash & Equivalents"
ws_bs["B5"] = 3127
ws_bs["B5"].font = blue_font
ws_bs["C5"] = 3127
[...]
ws_bs["A8"] = "Other Current Assets"
for col in range(2, 10):
    ws_bs.cell(row=8, column=col, value=500).font = blue_font
[...]
ws_bs["A11"] = "Goodwill & Intangibles"
for col in range(2, 10):
    ws_bs.cell(row=11, column=col, value=15000).font = blue_font

ws_bs["A12"] = "Other Non-Current Assets"
for col in range(2, 10):
    ws_bs.cell(row=12, column=col, value=5000).font = blue_font
\end{promptbox}

\noindent\emph{Deterministic consequence:} check S-09 measures
projection-region formula density 86.6\,\% (233 formulas over 269
numeric cells; Balance\_Sheet 75/105) --- below the 95\,\% facet bar
and the 90\,\% gate trigger --- so G4 (\emph{hardcoded forecast
cells}, Appendix~\ref{app:gatedefs}) fires and caps the cell's $\phi$
at 50 (pillar caps P2~40, P3~50). No partial-credit average would
surface this: line by line the sheet looks like a model, but nothing
downstream of those cells flexes, so sensitivity and reasoning are
unverifiable --- in the gate's words, ``a scenario snapshot, not a
model.'' At its extreme the same behavior never reaches scoring at
all: the artifact validator rejected 10 workbooks fleet-wide as
``too few forecast formulas \ldots{} --- looks like a paste-only /
dead model, not a live forecast'' (seven with 0 forecast formulas, two
with 5, one with 10, against a minimum of 12).

The three trajectories end in different recorded outcomes:
non-convergence, an absent artifact, and a completed artifact capped by
a validity gate. A single end-to-end accuracy value would map all three
to failure; the released ledger preserves the distinct failure class for
each cell.

\section{Gate Trigger Profiles}
\label{app:gates}

Figure~\ref{fig:gates} reports, per agent, the share of completed tasks
triggering each validity gate --- G1 balance-sheet identity, G2
cash-flow tie-out, G3 segment reconciliation, G4 hardcoded forecast
cells, G5 look-ahead leakage, G6 unresolvable source citation, G7
unresolved circular references, G8 EV$\to$equity bridge --- alongside
the any-gate rate from Table~\ref{tab:leaderboard}. Failure signatures
are model-specific rather than uniform: DeepSeek v4-flash concentrates
in G1 (56\% of tasks build a balance sheet that does not balance), the
Qwen3 235B and Qwen3 Coder in G7 circular references (69\% and 73\% ---
while the newer Qwen3.7-max shows no G7 at all, failing instead on G1
balance, 42\%), GPT-OSS-120B in G4
hardcoded forecasts (38\%), and GPT-5.6-terra and its sibling luna trip
the G2 cash-flow tie-out on 38\% and 27\% of tasks where sol almost
never does
(2\%). G5, the look-ahead-leakage gate, is the rarest and the most
skewed toward the weak tail: it fires on 17 of the 1{,}011 completed
cells (1.7\%), never for any of the top-eight agents, and its
detections are structural time inversions --- a forecast line item in
period $t$ whose formula reads a \emph{later} period of itself (a
balance-sheet roll run backwards from a future anchor, a discount
factor indexed off the wrong column). The pack clamps every dated
input to the as-of date by construction, so what G5 catches in
practice is not information leakage from the world but time-inverted
model mechanics --- and the cap is warranted: a model that computes
period $t$ from period $t{+}1$ is not forecasting.

\noindent\begin{minipage}{\columnwidth}
\centering
\includegraphics[width=0.8\textwidth]{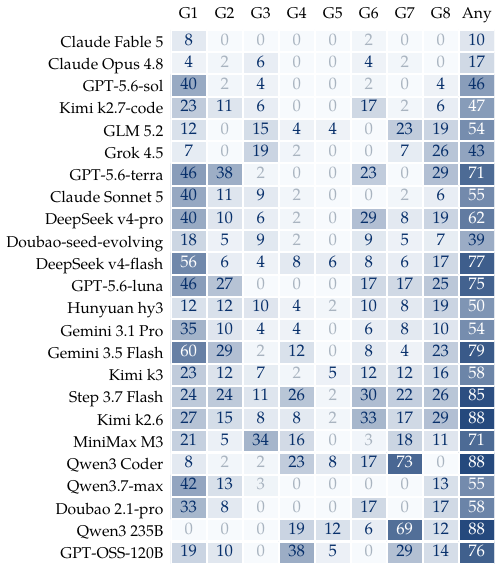}
\captionof{figure}{\textbf{Share of completed tasks (\%) triggering each validity
gate, per agent, in leaderboard order.} Shading encodes the printed value;
``Any'' is the any-gate rate of Table~\ref{tab:leaderboard}. Failure
signatures are model-specific; G5, the look-ahead-leakage gate, is
the rarest (1.7\% of cells) and never fires for a top-eight agent.}
\label{fig:gates}
\end{minipage}

\section{Judge Vote-Sampling Ablation: Detailed Setup}
\label{app:kablation}

\begin{table}[t]
\caption{\textbf{Judge vote-sampling ablation.} Six generation cells,
23 judged facets, 15-vote pools, and 800 bootstrap resamples. Subscore
std is on the 0--2 facet scale; $\tau$ compares two independent
re-judgings of the six cells.}
\label{tab:kablation}
\centering
\small
\begin{tabular}{crrr}
\toprule
$k$ & Subscore std & Ranking $\tau$ & Flip rate\\
\midrule
1  & 0.023 & 0.912 & 3.1\%\\
3  & 0.020 & 0.928 & 2.5\%\\
\textbf{5} & \textbf{0.019} & \textbf{0.944} & \textbf{2.2\%}\\
7  & 0.018 & 0.952 & 1.9\%\\
10 & 0.015 & 0.976 & 1.6\%\\
\bottomrule
\end{tabular}
\end{table}

Table~\ref{tab:kablation} reports the sweep over the vote count $k$.

\textbf{Judge configuration.} Full protocol --- verbatim prompts,
evidence-block budgets, vote reduction, caching parity --- in
Appendix~\ref{app:judge}. Relevant here: the serving path does not
accept a sampling temperature for this model, so vote sampling is the
\emph{only} variance-control mechanism --- which is exactly what this
ablation measures.

\textbf{Cells.} Six generation cells drawn from the main run, one from
each of six distinct generator models (two proprietary frontier
families, one open-weight, spanning three GICS sectors), chosen so their
deterministic scores span the observed range ($\phi_{\text{det}}$
26--50). Each cell activates all 23 judged facets (no
industry-conditional deactivations in this sample).

\textbf{Collection.} Each facet is judged once at a pool size of 15
votes (per-cell: $23\times15=345$ judge calls), so every $k$ condition
is analyzed from the \emph{same} underlying pool rather than from
separate paid runs. Raw votes are released with the harness.

\textbf{Bootstrap.} For each $k\in\{1,3,5,7,10\}$ and each metric we
draw $B=800$ resamples with replacement from the vote pools.
(i)~\emph{Subscore std}: a resampled judged subscore is the mean over
facets of the reduced numeric scores (N-A dropped); we report the mean
over cells of the per-cell standard deviation across resamples, on the
0--2 facet scale. (ii)~\emph{Ranking stability}: two independent
resampled re-judgings of all six cells are ranked and compared by
Kendall's $\tau$; we report the mean over $B$ pairs.
(iii)~\emph{Flip rate}: per facet, the probability that a $k$-vote
reduction differs from that facet's modal reduced outcome, averaged over
all facets and cells.

\textbf{Reading.} Because each vote pool is finite (15), the bootstrap
slightly \emph{understates} true run-to-run variance for $k$ close to
the pool size; conclusions about small $k$ (the operating regime) are
unaffected. The absence of an elbow means the choice of $k$ is a cost
knob, not a validity cliff; we fix $k{=}5$ before scoring any main-table
run.

\section{Release Protocol and Benchmark-Design Checklist}
\label{app2:release}

The Reproducibility Statement in the main text states \emph{what} is
released and under which access tier; this appendix
records the operational protocol: the artifact inventory, the
withholding rationale, the de-identification acceptance gates, the
refresh mechanics, and a mapping onto the axes of the
agentic-benchmark checklist~\citep{abc}.

\textbf{Artifact inventory.} Every scoring input is shipped as data,
not baked into harness code, so a third party can re-derive any
reported number --- or dispute any rubric --- without us. Artifacts
carrying workbook content ship in the gated data tier; rubrics,
scorer, run records, and reports are public:

\begin{promptbox}{Release inventory (annotations ours)}
workbooks/M{ID}-{TICKER}-{Tier}.xlsx  199 de-identified books
deid_report.{md,json}     TRANSFORM+VERIFY, masked findings
rescan_report.{md,json}   independent re-scan: 0 residual
inputs/                   visible packs, forecast cells blank
rubrics/taxonomy.json     5 pillars / 21 sub-caps / 56 facets
rubrics/gates.json        G1-G8 caps + detection wiring
rubrics/industry_overlays/matrix/  25 GICS overlay files
rubrics/facet_grader_map.json      check-id -> facet wiring
rubrics/{phi_aggregate,R1_*,R2v2_*,R3_*}  the full scorer
  + R2v2 judge system/user prompts, verbatim
envelopes_full/E_company.json      65 multi-covered companies
envelopes_full/E_industry_gics.json  industry distributions
envelopes_full/envelopes/M*.json   1,000 per-book extractions
envelopes_full/_coverage.json      per-method coverage counts
broker_extractions/       extraction scripts + reports
runs_analyst_vs_analyst/pairs.jsonl   158 directed pairs
                                      x 4 tolerance mults
runs_analyst_vs_analyst/summary.json
reference/gates_vs_linear.json     gate-cap vs linear ablation
runs/campaign/w2_main/ledger.jsonl    per-cell run ledger
runs/campaign/w2_main/judge_k5.jsonl  raw votes + agreement
runs/campaign/w2_main/manifest.json   frozen run config
train_split/train_split_manifest.json 200-ticker split
qc_excel200_summary.json  batch QC: 196/200 harness-ready
identity_triage_200.jsonl per-workbook identity triage
dupe_groups.json          cross-batch dedup, 1,201 files
\end{promptbox}

\noindent Three of these deserve a sentence. The envelope statistics
ship as raw value lists, not summaries: \texttt{E\_company.json} keys
each of the 65 multiply-covered companies to its contributing
workbooks and their extracted values, and \texttt{\_coverage.json}
states, per extraction method, how many of the 1{,}000 parsed
workbooks (of 1{,}001 delivered) yield each ingredient (WACC from
862, an in-book triangulation from 354, a sensitivity grid from 310,
terminal growth from 494) --- so the partial-coverage caveat of
Section~\ref{sec:envelope} is checkable, not just stated. The
analyst-vs-analyst audit (Section~\ref{sec:ava}) releases its 632
per-pair, per-tolerance records (158 directed same-ticker pairs at
four tolerance multipliers), so the paper's central negative result is
recomputable from JSONL. And every judged cell in the frozen run
retains all five raw votes with per-facet agreement in
\texttt{judge\_k5.jsonl}, so any judged score can be re-aggregated
without an API call; the run ledger records per-cell generation
outcome, wall-clock, token usage, attempts, and wire adapter.

\textbf{What is withheld, and why.} Three classes. (1)~The
$\sim$600 non-evaluation, non-training-split workbooks: published data
enters future training corpora, so the withheld pool is the
contamination-resistant refresh track (below). (2)~The internal
manual-review companions to the de-identification reports, which
contain raw pre-masking values (names, e-mails, paths); the public
reports carry only \texttt{sha256}-masked digests of removed content.
(3)~Per-task evaluation targets: input packs deliberately contain no
peer multiples, consensus, or broker data --- they are the evaluation
target. Separately, the one content-mislabeled workbook found by
identity triage is quarantined and ships
in no pool.

\textbf{De-identification verification.} The pipeline
(\texttt{deidentify.py} v1.0.0) treats each workbook at the XML level
across eleven numbered metadata surfaces plus an extras scan, in a
transform-then-verify pass followed by an \emph{independent} scan-only
re-scan of the treated output. On the 199-workbook release batch (201
candidates: one lockfile skipped, one quarantined), 180 workbooks
required at least one hard scrub --- analyst-identifying absolute
paths in 169, local external-link targets (836 findings across 38
books), personal-cloud link targets in 25, comment authors rewritten
to \texttt{analyst}. Acceptance is gated, per book, on: re-opening
under \texttt{openpyxl}, zip integrity, formula-count delta of
exactly zero verified two ways (parsed \texttt{<f>} elements and raw
byte count), and zero residual hard findings; the independent re-scan
then reports zero hard findings across all 199. A 2{,}123-item
manual-review queue (dominated by 1{,}928 defined-name tokens) is
surfaced for human triage in the internal file that does not ship.

\textbf{Refresh protocol.} The vendor has confirmed the evaluation
workbooks never entered any model training pipeline
(Section~\ref{sec:corpus}). The released training split
(\texttt{train\_split\_v1}; 200 tickers; seed 20260720) excludes, by
recorded policy, the evaluation batch, the 65 multiply-covered
calibration companies, and the quarantined workbook, stratified by
tier ($80/70/40/10$) with sector round-robin. Promoting a refresh
batch from the withheld pool means re-running the same staged
pipeline, each stage emitting the machine-readable report shown
above: corpus QC and archetype readiness (the current batch's report
records 196/200 harness-ready), identity triage, cross-batch content
de-duplication (\texttt{dupe\_groups.json} already spans 1{,}201
files; 11 duplicate groups), de-identification with the identical
acceptance gates, provenance-tracked pack extraction with the
per-pack $TA = TL + TE$ identity check, and envelope re-extraction. A
batch ships only when every gate that gated the current release
passes on the new batch; the scoring configuration (taxonomy, gates,
overlays, judge prompts) ships as explicit versioned artifacts rather
than harness internals, so refreshed scores are attributable to new
tasks, not silent rubric drift.

\textbf{Checklist compliance.} Mapping GAUGE onto the checklist
of \citet{abc}, each axis against shipped artifacts:

\emph{(1) Outcome validity.} The grading target is a calibrated
envelope, not one analyst's spreadsheet, and the case against the
single-golden alternative is itself released as data
(\texttt{pairs.jsonl}). Structural facets are graded by a
deterministic checker with zero rerun variance conditional on the
implemented detector; judged facets retain
$k{=}5$ raw votes so judge noise is inspectable rather than laundered
into a scalar.

\emph{(2) Task validity.} Input packs are generated by
provenance-tracked extractors --- every number carries its source
sheet, row, and label --- with a verified accounting identity per pack.
Reference data follows the same ``abstain, don't fabricate''
discipline imposed on agents: the four unresolvable reference cases
are documented abstentions, not imputed fields
(Section~\ref{sec:task}).

\emph{(3) Honest reporting.} Capability failures score as failures
and count against completion --- retrying them to success would bias
toward each model's best case (Section~\ref{sec:results}). Every
scorecard carries a coverage manifest; inactive facets are excluded
from aggregates, never zero-filled; per-gate trigger profiles are
disclosed in Appendix~\ref{app:gates}.

\emph{(4) Contamination.} Vendor confirmation for the evaluation set,
a withheld refresh pool as a stated design feature, and a training
split that is leakage-controlled by construction against both task
answers and envelope calibration data.

\emph{(5) Overfitting and gaming.} Gaming the letter of the gates
while failing judgment facets is bounded by envelope bands that are
distributional rather than point targets, and detectable over time by
the refresh track: an agent tuned to a published batch must survive a
batch that was never public.

\emph{(6) Grader-integrity self-audit.} Grading instruments are
audited artifacts: the G6 over-scoping incident
(Appendix~\ref{app:gatedefs}) and the unattached activation-overlay
incident (Appendix~\ref{app2:provenance}) were both caught by our own
audits, corrected, and reported with their measured impact rather
than suppressed.

We claim no blanket compliance. Where GAUGE deviates --- partial
envelope coverage, residual LLM-judge risk --- the deviation is
quantified and stated in Appendix~\ref{app:limitations}, and the
artifacts
above are sufficient for a reader to re-measure it.

\section{Related Work}
\label{app:related}
\label{sec:related}

\textbf{Benchmarks for professional finance agents.}
Until recently, financial evaluation of language models mostly meant
question answering over filings or market data. A newer generation of
benchmarks instead asks agents to do the professional job end to end,
and each makes a distinct measurement choice. BankerToolBench
evaluates complete junior-banker workflows, from data rooms and
market-data tools through Excel models, presentations, and written
reports, and grades them with stakeholder-oriented criteria authored
with practitioner input~\citep{btb,zhu2026knowing}. FrontierFinance
is the closest to GAUGE in task design: it contains 25 from-scratch
financial-modeling tasks across five model types, each with an
expert-built reference model, a detailed rubric, an initial validity
gate, and a human-expert baseline~\citep{ff}. BigFinanceBench
evaluates 928 open-ended, multi-source research questions by scoring
the visible steps of the workflow, such as source choice,
definitions, adjustments, and calculations, against point-weighted
rubrics~\citep{bfb}. BlueFin covers 131 spreadsheet synthesis,
manipulation, and comprehension tasks with 3{,}225 granular criteria
and validates its agentic judge against expert
labels~\citep{bluefin}.

These are strong designs for what they measure, and
Table~\ref{tab:compare} makes the comparison explicit along the
measurement dimensions this paper is about. The shared choice that
GAUGE departs from is the referent: when a task's target is a
forward-looking valuation, all four grade the agent's numbers by
proximity to one expert's numbers, and Section~\ref{sec:ava} measures
what that rule does to professionals themselves. The critique is
scoped accordingly. It concerns from-scratch valuations scored
against a single author's point estimates; it does not concern the
uniquely verifiable spreadsheet operations that several of these
benchmarks intentionally target, for which a single reference is the
right design. GAUGE's second departure is the treatment of structural
failure: categorical defects impose hard score caps instead of being
folded into a weighted average, with FrontierFinance's initial
validity gate as the closest published analog.

\begin{table}[t]
\caption{\textbf{Measurement-design positioning} against the four
single-reference financial-agent benchmarks, on judgment-bearing
modeling tasks.
\yes{} yes; \pmark{} partial; \nomark{} absent in the published
design.}
\label{tab:compare}
\centering
\footnotesize
\renewcommand{\arraystretch}{1.05}
\begin{tabular}{@{}lccccc@{}}
\toprule
Measurement design & BTB & FF & BFB & BlueFin & \textbf{GAUGE}\\
\midrule
From-scratch valuation            & \pmark & \yes & \nomark & \pmark & \yes\\
Multi-analyst referent            & \nomark & \nomark & \nomark & \nomark & \yes\\
Empirical ranges for judgment     & \nomark & \nomark & \nomark & \nomark & \yes\\
Human performance baseline        & \nomark & \yes & \nomark & \nomark & \yes\\
Known-groups validation           & \nomark & \nomark & \nomark & \nomark & \yes\\
Categorical gates / score caps    & \nomark & \yes & \nomark & \nomark & \yes\\
Industry-conditional rubric       & \nomark & \nomark & \nomark & \nomark & \yes\\
Private holdout / refresh supply  & \pmark & \nomark & \nomark & \pmark & \yes\\
\bottomrule
\end{tabular}
\end{table}

\textbf{Spreadsheet and artifact-level evaluation.}
SpreadsheetBench and related spreadsheet suites express correctness
as a deterministic cell, formula, or state match~\citep{ssbench}, and
repository-level code benchmarks such as SWE-bench grade a complete
artifact rather than an isolated answer~\mbox{\citep{swebench}}. A financial
model shares the artifact-level dependency structure of a repository,
since one wrong link propagates through everything downstream, but it
adds a property that code mostly lacks: non-identifiability. Two
linked, internally consistent models can legitimately differ in their
revenue paths, capital structures, and discount rates, so state
matching cannot serve as the whole grading rule. GAUGE therefore
keeps deterministic checking exactly where it is valid (accounting
identities, formula liveness, statement linkage) and switches to
empirical ranges only for the judgment-bearing quantities. LiveBench
and LiveCodeBench motivate continuously refreshed evaluation data as
a contamination control~\citep{livebench,livecodebench}; GAUGE
implements the same idea for licensed private data through its
withheld refresh pool, with overlap anchors so that scores remain
comparable across versions.

\textbf{Measurement validity and LLM judging.}
A growing measurement-science literature treats the benchmark itself
as the object under study. Construct-validity audits ask whether a
score supports its intended interpretation rather than merely being
repeatable~\citep{measuringwhatmatters}, and BetterBench codifies
benchmark-quality criteria at the level of documentation and
design~\citep{betterbench}. The agentic-benchmark checklist examines
task validity, contamination, honest failure accounting, and grader
integrity~\citep{abc}; work on rating indeterminacy separates judge
error from items that genuinely admit several defensible
answers~\citep{indeterminacy}; and item-sampling methods study how
much evaluation a finite budget buys~\citep{tinybench,fluid}. GAUGE
operationalizes this program for a professional domain. The
peer-workbook audit tests the single-reference assumption directly;
the human study supplies known-groups evidence; the perturbation
control measures selectivity rather than coverage alone; and the
released per-vote judge records make judge noise inspectable instead
of laundering it into a scalar. Appendix~\ref{app2:release} maps each
checklist axis to a shipped artifact.

\section{Limitations}
\label{app:limitations}
\label{sec:ethics}

The envelope is a reference distribution of observed practice, not
ground truth: an in-band value is defensible in the sense that
professionals demonstrably use it, which is weaker than being
correct, and only 17 undirected pairs support the implied-price
disagreement tail. The corpus, its seniority labels, and the human
participants all come from one commercial provider network with a US
listing skew, and we did not independently re-verify individual
credentials. Twenty-three of 56 facets depend on one frozen LLM
judge; the audits of Section~\ref{sec:validation} establish
stability, 460-case expert agreement, and rank-neutrality under a
rival judge, but not per-case correctness. The longitudinal panel covers
48 of 196 tasks at one generation per agent--task pair, so full-bank
generalization and generation variance are quantified by bootstrap
and sentinel-set estimates rather than measured exhaustively.
Finally, the peer-workbook audit of Section~\ref{sec:ava} applies a
shared tolerance rule rather than reproducing each competing
benchmark's complete rubric, and observed workbook differences mix
legitimate choice with date, horizon, purpose, and quality
heterogeneity; the audit's conclusion is correspondingly scoped to
the single-reference rule itself.

%% file: figs/tab_tooluse.tex
\begin{tabular}{@{}lrrrrrrrrc@{}}
\toprule
 & & \multicolumn{2}{c}{Mix} & \multicolumn{3}{c}{Errors fed back (\%)} & & & \\
\cmidrule(lr){3-4}\cmidrule(lr){5-7}
Agent & Calls & /task & \texttt{wf}\,\% & Py & Trunc. & All & Cap & Rec.\,\% & Open\\
\midrule
Claude Fable 5 & 446 & 9.3 & 36 & 1.8 & 2.5 & 4.5 & 21 & 98 & \nomark\\
Claude Opus 4.8 & 920 & 19.2 & 20 & 2.8 & 0.0 & 2.8 & 27 & 100 & \nomark\\
GPT-5.6-sol & 140 & 2.9 & 14 & 5.7 & 6.4 & 12.1 & 0 & 56 & \nomark\\
Kimi k2.7-code & 1,637 & 34.8 & 12 & 6.4 & 2.0 & 8.6 & 85 & 62 & \yes\\
GLM 5.2 & 2,159 & 45.0 & 19 & 5.0 & 4.4 & 9.4 & 91 & 88 & \yes\\
Grok 4.5 & 257 & 6.0 & 0 & 6.6 & 0.0 & 6.6 & 55 & 10 & \nomark\\
GPT-5.6-terra & 254 & 5.3 & 0 & 13.0 & 0.4 & 13.4 & 0 & 54 & \nomark\\
Claude Sonnet 5 & 2,381 & 49.6 & 19 & 2.3 & 0.6 & 3.1 & 9 & 100 & \nomark\\
DeepSeek v4-pro & 1,708 & 35.6 & 24 & 3.8 & 5.7 & 9.5 & 51 & 0 & \yes\\
Doubao-seed-evolving & 1,145 & 25.3 & 26 & 7.9 & 7.5 & 15.4 & 98 & 68 & \nomark\\
DeepSeek v4-flash & 1,217 & 25.4 & 22 & 8.1 & 6.1 & 14.2 & 57 & 0 & \yes\\
GPT-5.6-luna & 297 & 6.2 & 1 & 13.8 & 0.0 & 14.1 & 0 & 40 & \nomark\\
Hunyuan hy3 & 1,778 & 37.0 & 8 & 14.6 & 0.2 & 15.0 & 18 & 94 & \yes\\
Gemini 3.1 Pro & 358 & 7.5 & 11 & 4.5 & 0.3 & 4.8 & 0 & 2 & \nomark\\
Gemini 3.5 Flash & 450 & 9.4 & 32 & 3.1 & 1.8 & 4.9 & 0 & 0 & \nomark\\
Kimi k3 & 1,593 & 36.8 & 2 & 4.5 & 0.1 & 4.6 & 2 & 23 & \nomark\\
Step 3.7 Flash & 905 & 19.7 & 17 & 7.6 & 0.9 & 8.5 & 91 & 0 & \yes\\
Kimi k2.6 & 1,645 & 34.3 & 4 & 3.7 & 1.3 & 5.0 & 15 & 0 & \yes\\
MiniMax M3 & 1,798 & 46.6 & 19 & 6.1 & 0.2 & 6.6 & 57 & 97 & \yes\\
Qwen3 Coder & 375 & 7.8 & 23 & 6.7 & 0.0 & 6.9 & 0 & 0 & \yes\\
Qwen3.7-max & 479 & 13.9 & 40 & 1.7 & 17.1 & 18.8 & 72 & 0 & \nomark\\
Doubao 2.1-pro & 394 & 22.9 & 30 & 6.1 & 10.7 & 16.8 & 40 & 42 & \nomark\\
Qwen3 235B & 593 & 10.8 & 15 & 4.4 & 1.2 & 5.6 & 6 & 0 & \yes\\
GPT-OSS-120B & 257 & 6.0 & 13 & 4.3 & 12.1 & 20.2 & 29 & 0 & \yes\\
\bottomrule
\end{tabular}